\documentclass[lettersize,journal]{IEEEtran}
\usepackage{amsmath,amsfonts}
\usepackage{algorithmic}
\usepackage{algorithm}
\usepackage{array}
\usepackage{subfigure}
\usepackage{textcomp}
\usepackage{stfloats}
\usepackage{url}
\usepackage{verbatim}
\usepackage{graphicx}
\usepackage{cite}
\usepackage{times}
\usepackage{soul}
\usepackage{amsthm}
\usepackage{amssymb}
\usepackage{mathtools}
\usepackage{booktabs}
\usepackage[hidelinks]{hyperref}
\usepackage[utf8]{inputenc}
\usepackage[small]{caption}
\usepackage{multirow}
\usepackage{subcaption}
\usepackage{cuted}
\usepackage{lineno}
\usepackage{pifont}

\usepackage{tikz}  
\newcommand*\circled[1]{\tikz[baseline=(char.base)]{
            \node[shape=circle,draw,inner sep=1pt] (char) {#1};}}

\usepackage{xcolor}
\definecolor{lightred}{rgb}{1, 0.8, 0.8}
\definecolor{lightorange}{rgb}{1, 0.9, 0.8}
\definecolor{lightblue}{rgb}{0.85, 0.85, 1}
\definecolor{lightgreen}{rgb}{0.85, 1, 0.85}
\definecolor{lightpurple}{rgb}{0.9, 0.8, 1}

\definecolor{darkgreen}{RGB}{30, 160, 30}

\theoremstyle{plain}
\newtheorem{theorem}{Theorem}
\newtheorem{proposition}{Proposition}
\newtheorem{lemma}{Lemma}
\newtheorem{corollary}{Corollary}
\theoremstyle{definition}
\newtheorem{definition}{Definition}
\newtheorem{assumption}{Assumption}
\theoremstyle{remark}
\newtheorem{remark}{Remark}
\usepackage[textsize=tiny]{todonotes}
\begin{document}

\title{FedCoSR: Personalized Federated Learning with Contrastive Shareable Representations for \\Label Heterogeneity in Non-IID Data}

\author{Chenghao Huang,~\IEEEmembership{Student Member,~IEEE}, 
Xiaolu Chen,
Yanru Zhang,~\IEEEmembership{Senior Member,~IEEE},
and \\Hao Wang,~\IEEEmembership{Member,~IEEE}
\thanks{This work was supported in part by the Australian Research Council (ARC) Discovery Early Career Researcher Award (DECRA) under Grant DE230100046, the Major Project of Shenzhen Science and Technology Research and Development Fund under Grant No. JCYJ20241206180207010, Project of Sichuan Science and Technology Program under Grant No. 2024ZYD0274 and No. 2024YFG0006, and Project of Guangdong Science and Technology Program under Grant No. 2025A1515010246. (Corresponding author: Hao Wang.)}
\thanks{C. Huang and H. Wang are with the Department of Data Science and AI, Faculty of IT and Monash Energy Institute, Monash University, Melbourne, VIC 3800, Australia (e-mails: \{chenghao.huang, hao.wang2\}@monash.edu).}
\thanks{X. Chen and Y. Zhang are with the School of Computer Science and Technology, University of Electronic Science and Technology of China (UESTC), Chengdu, and Shenzhen Institute for Advanced Study of UESTC, Shenzhen, China (e-mails: jzzcqebd@gmail.com, yanruzhang@uestc.edu.cn).}
}

\markboth{IEEE Transactions on Cybernetics, 2025}
{Huang \MakeLowercase{\textit{et al.}}: FedCoSR: PFL for Contrastive Shareable Representations for Label Heterogeneity in Non-IID Data}


\maketitle

\begin{abstract}
Heterogeneity arising from label distribution skew and data scarcity can cause inaccuracy and unfairness in intelligent communication applications that heavily rely on distributed computing. To deal with it, this paper proposes a novel personalized federated learning algorithm, named Federated Contrastive Shareable Representations (FedCoSR), to facilitate knowledge sharing among clients while maintaining data privacy. 
Specifically, the parameters of local models' shallow layers and typical local representations are both considered as shareable information for the server and are aggregated globally. 
To address performance degradation caused by label distribution skew among clients, contrastive learning is adopted between local and global representations to enrich local knowledge.
Additionally, to ensure fairness for clients with scarce data, FedCoSR introduces adaptive local aggregation to coordinate the global model involvement in each client. Our simulations demonstrate FedCoSR's effectiveness in mitigating label heterogeneity by achieving accuracy and fairness improvements over existing methods on datasets with varying degrees of label heterogeneity. 
\end{abstract}


\section{Introduction}\label{Sec:introduction}

\subsection{Background and Motivation}

In today's connected world, Intelligent Communication (IC) plays a pivotal role in enabling data-driven applications. From personalized recommendations on smartphones to real-time health monitoring via wearable devices, these systems rely heavily on large volumes of data collected from distributed sources, such as mobile devices or sensors, which can form complex and diverse systems through Internet-of-Things (IoT), data centers, and cloud servers ~\cite{fortino2020internet,lim2020federated,li2023fedlga,zhang2023federated,kaheni2024selective}. Since data-driven solutions based on IC have empowered industries like healthcare, smart homes, and finance to provide tailored services, they also raise privacy concerns, as personal and sensitive data are frequently involved~\cite{zhang2021privacy,liang2022secure}.

To address privacy issues, Federated Learning (FL) has emerged as a promising paradigm. Rather than centralizing sensitive data in one location, FL allows distributed clients, such as smartphones, IoT devices, or edge servers, to collaboratively train machine learning models while keeping their data locally stored~\cite{mcmahan2017communication}. This distributed approach helps preserve user privacy while leveraging the computational capabilities of these distributed devices.
Although FedAvg~\cite{mcmahan2017communication} performs well with IID data, real-world client data are often non-IID due to diverse user behaviors, preferences, and environments~\cite{kairouz2021advances}, limiting the generalization of a single global model.

In this paper, we focus on two major forms of statistical heterogeneity below.
\begin{itemize}
    \item \textbf{Label distribution skew} refers to the FL situation where the distribution of labels varies significantly across different clients~\cite{ye2023heterogeneous}. Due to the imbalance in label distribution, it is challenging to train one global model that generalizes well across all clients.
    For instance, in smart home devices, geographic location, user interests, and lifestyle differences can result in vastly different user behavior patterns captured by devices.
    \item \textbf{Data scarcity} in a distributed system, also known as data quantity skew among clients, poses an additional challenge on fairness~\cite{imteaj2021survey,tan2022towards}. On one hand, the labels of scarce data are sometimes unique and important, such as rare anomalies in industrial applications or survey results of minority groups. Unfairness may arise from the failure to integrate valuable knowledge contained in scarce data into the global model. On the other hand, scarce data are highly susceptible to causing overfitting, impeding personalization. This usually occurs in scenarios with monopolized clients and minority clients, or newly participating clients with few historical data, resulting in challenges in integrating these data globally.
\end{itemize}

\begin{figure}[t]
    \centering
    \includegraphics[width=0.48\textwidth]{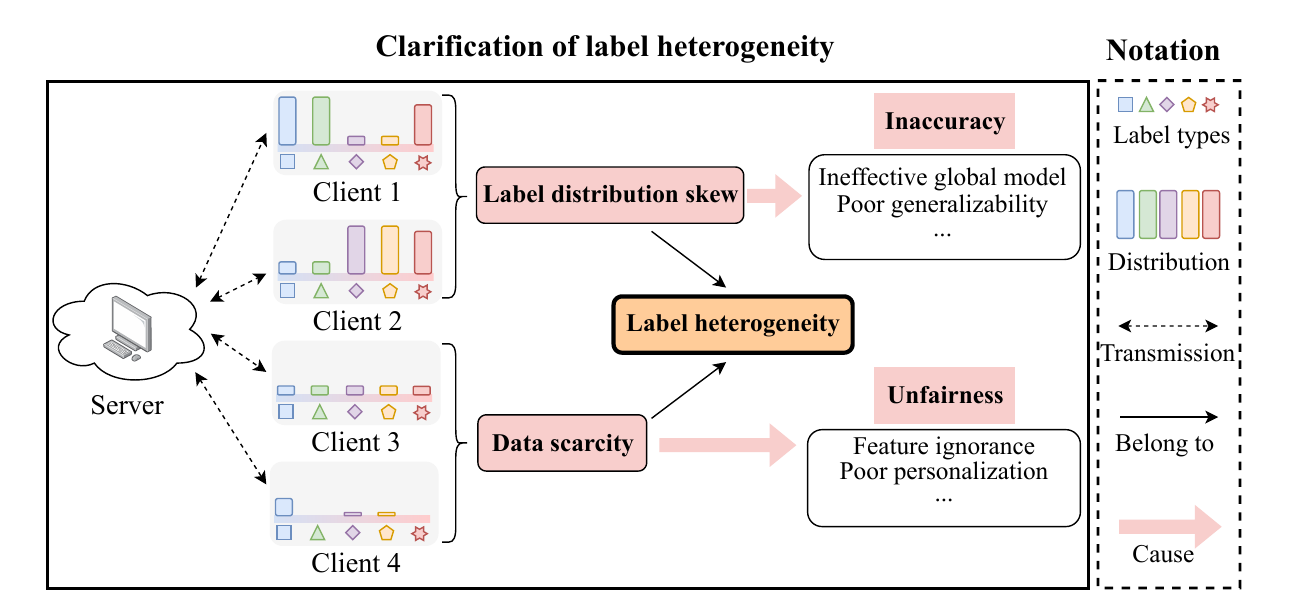}
    \caption{Clarification of label heterogeneity, including label distribution skew and data scarcity, which are the primary challenges focused in this work through contrastive representation learning-based local adaptive aggregation and local personalized training.}
    \label{Fig:label heterogeneity}
\end{figure}

In real-world applications, both types of statistical heterogeneity often coexist. For instance, in industrial monitoring, varying production processes and equipment configurations can lead to imbalanced distributions of common fault labels while leaving rare, critical anomaly data scarce~\cite{10909241}. 
Similarly, mobile healthcare systems frequently have abundant data for common conditions but limited samples for rare diseases.
Consequently, such statistical heterogeneity present substantial challenges to FL, demanding solutions that balance generalizability, personalization, and fairness. Note that, since we mainly study the labels of scarce data, data scarcity is also regarded as a kind of label skew in this paper. Thus, we collectively refer to the above two types of statistical heterogeneity as label heterogeneity, clarified in Fig.~\ref{Fig:label heterogeneity}.

To deal with label heterogeneity, Personalized Federated Learning (PFL) has emerged as an extension of traditional FL and received attention~\cite{kairouz2021advances,tan2022towards}. PFL tailors personalized models for each client within a collaborative training paradigm to achieve robust performance on heterogeneous datasets. Notably, it has been proven that it is effective to coordinate the distance between the global model and local models through regularizing the local training objective~\cite{li2020federated,t2020personalized,li2021ditto}. Furthermore, a substantial body of research has focused on splitting the model structure into representation layers, serving as a common feature extractor, and projection layers which address specific tasks~\cite{arivazhagan2019federated,liang2020think,li2021model,collins2021exploiting,11075729}. Despite these advancements, the above methods only improve clients' performance through generic model parameters but ignore more fine-grained characteristics inherent in data, especially label distribution skew. On the other hand, recent studies have explored data quantity skew problems~\cite{wang2023federated}, but they have not adequately addressed co-existence of label distribution skew and data scarcity, leaving a research gap.

Thanks to the close correlation between data and representations~\cite{bengio2013representation,karg2020efficient}, shareable representations are introduced in FL to improve personalization while preserving privacy~\cite{tan2022fedproto}, and further integrated with the global model to facilitate knowledge integration~\cite{FedGH,xu2022personalized}. However, these methods primarily focus on same-label representation alignment, limiting generalizability, especially with skewed label distributions. 
In light of this, Contrastive Representation Learning (CRL) which emphasizes deriving knowledge from label-agnostic representations~\cite{chen2020simple}, offers a promising perspective. We believe that utilizing shared representations among clients can further contribute to mitigating label heterogeneity.

\subsection{Main Work and Contributions of FedCoSR}
\begin{figure}[t]
    \centering
    \includegraphics[width=0.48\textwidth]{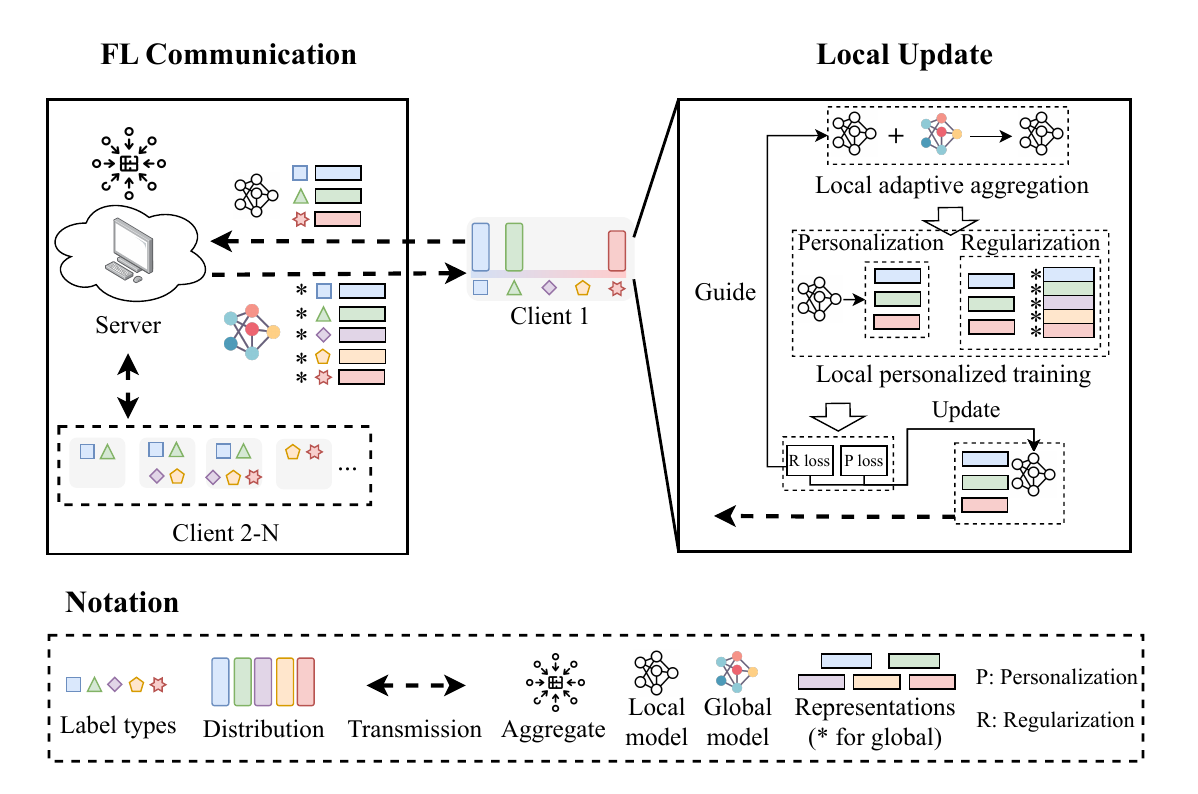}
    \caption{An illustrative diagram of FedCoSR--the proposed PFL framework aiming to address label heterogeneity, where the local personalized training combines loss functions of both personalization and regularization.}
    \label{Fig:brief diagram}
\end{figure}

This paper introduces a novel approach leveraging CRL~\cite{chen2020simple} on shareable representations among clients, aiming to address the label heterogeneity caused by label distribution skew and data scarcity in distributed ML scenarios. The brief process of the proposed PFL framework is illustrated in Fig.~\ref{Fig:brief diagram}.

Globally, the server receives and separately aggregates local model parameters and typical representations of each client. Then, the server sends the global model parameters and the global representations, which provide additional knowledge for performance improvement, to all clients for personalization. 

During the local update, each client conducts personalization primarily through local aggregation and local training.
For local training, CRL is adopted to foster similarity among representations with the same label and dissimilarity among those with different labels. To mitigate label distribution skew and simultaneously enhance knowledge of data-scarce clients, the global representations are used to construct positive and negative sample pairs for the local representations.
Moreover, for local aggregation, a loss-wise weighting mechanism is introduced to coordinate personalization among clients with data quantity skew, especially for data-scarce clients. This mechanism ensures that, when the local model's contrastive loss is high, indicating limited capability in distinguishing samples with different labels, the global model contributes more to facilitate improvement. Conversely, when the local model's contrastive loss is low, the global model participates less to avoid compromising personalization.

The contributions of this work are as follow.
\begin{itemize}
    \item We address a challenging IC scenario that concurrently involves both label distribution skew and data scarcity, in contrast to most FL studies that tackle these issues separately. By effectively tackling these two challenges, we aim to enhance the practicality of our FL algorithm, enabling more robust and equitable performance across diverse and data-limited IC applications in real world.
    \item We propose Federated Contrastive Shareable Representation (FedCoSR) to jointly integrate representation-level and model-level personalization for clients with heterogeneous labels and data quantity skew, thereby improving overall performance and fairness. Specifically, through contrastive learning on shareable representations, FedCoSR distills complementary knowledge from clients with different label distributions to improve representation quality. In parallel, an adaptive local aggregation principle driven by the contrastive loss dynamically adjusts the degree of global model participation, ensuring fairness by preventing local models with larger datasets from undermining the personalized knowledge of those with smaller datasets.
    \item We provide a theoretical analysis of the designed local loss function and prove the communication convergence of FedCoSR under the dual challenges of label distribution skew and data scarcity. Extensive image classification experiments across varying levels of heterogeneity demonstrate that FedCoSR consistently outperforms state-of-the-art PFL methods in both accuracy and fairness, particularly in highly challenging scenarios.
\end{itemize}

There are six sections. Section~\ref{Sec:related work} reviews the related work of FL and CRL. Section~\ref{Sec:method} presents
the formulation of PFL, representation sharing, and the developed FedCoSR algorithm. Section~\ref{Sec:theoretical} analyzes the effectiveness of the designed loss function and the non-convex convergence of the developed algorithm. The experimental results and discussions are provided in Section~\ref{Sec:simulation}. Section~\ref{Sec:conclusion} concludes the paper.

\section{Literature Review}\label{Sec:related work}

\subsection{Classical FL Methods}
FedAvg~\cite{mcmahan2017communication}, a classical FL method, aggregates locally-trained updates from distributed clients by averaging their weights, updating the global model, and redistributing the updated model. This simple yet effective approach works well under IID data but struggles with the non-IID data prevalent in real-world applications.
To address this, PFL methods have been proposed to mitigate statistical heterogeneity.
FedProx~\cite{li2020federated} and Ditto~\cite{li2021ditto} both introduce an L2 regularization between local and global models; FedProx focuses on stable global convergence, while Ditto enhances local personalization.
pFedMe~\cite{t2020personalized} improves personalization via a bi-level optimization framework that separates local adaptation from global aggregation.
PerFedAvg~\cite{fallah2020personalized} combines meta-learning with FedAvg to learn a global model that quickly adapts to local data with few gradient updates, addressing device heterogeneity.

\subsection{PFL Methods}
Some PFL research focuses on structural modifications to the model itself, such as model splitting and client-specific updates, to further enhance personalization.

Model splitting typically splits the local model parameters into two parts: the shallow layers, known as the base model, which is sent to the server for global aggregation; the deep layers,  known as the head model, which remains locally for local personalized training.
Representative models include FedPer~\cite{arivazhagan2019federated} and FedMPS~\cite{10856888} (which upload base), LG-FedAvg~\cite{liang2020think} (which uploads head), and FedRep~\cite{collins2021exploiting} (which uploads base plus extra local training).

Another approach explores client-tailored updates, which adapt local models in each round.
Specifically, FedAMP~\cite{huang2021personalized} exchanges local model information via the global server based on each local model's similarity to others, enabling personalized model updates. 
FedALA~\cite{zhang2023fedala} trains an additional parameter matrix for each client to adjust the local aggregation of global model parameters, but leading to the cost of extra computational overhead.
FedPDN~\cite{10835748} uses inter-class similarity constraints and parameter decoupling for medical image personalization.
FedAS~\cite{10655727} mitigates intra- and inter-client inconsistency via Fisher Information Matrix-based weighting.
PeFLL~\cite{scott2024pefll} jointly trains embedding and hyper-networks for ready-to-use models under data scarcity.
Despite the effectiveness in addressing statistical heterogeneity, these PFL methods only utilize model parameters for personalization but overlook more direct information related to raw data.

To further address finer-grained statistical heterogeneity, representation learning~\cite{bengio2013representation} has been incorporated into PFL, as it can capture more transferable knowledge from local data.
A representation is the feature embedding learned from raw data by deep models, encoding essential patterns and structures for downstream tasks.
FedProto~\cite{tan2022fedproto} shares local representations via the server and uses their L2 distance to regularize local training.
FedPC~\cite{10988632} applies prototype-based clustering for medical imaging to enhance personalization in highly heterogeneous settings.
FedGH~\cite{FedGH} extracts local representations via each client's base model, uploads them to the server for training the global head, and sends it back.
FedPAC~\cite{xu2022personalized} aggregates each client's head model using all clients' heads, weighted by the similarity of uploaded local representations.
pFedFDA~\cite{mclaughlin2025personalized} estimates a global feature distribution and adapts it to each client's local distribution via interpolation.
FedDBL~\cite{10572001} reduces communication and mitigates data scarcity by uploading low-dimensional pretrained features for server-side broad learning.
FedStream~\cite{10198520} maintains local prototypes to handle concept-drifting data streams under label shifts.
FedLFP~\cite{10950126} partitions representations to cut uplink costs in mobile edge computing while preserving personalization.
pFedKD~\cite{10855800} uses knowledge distillation with pseudo data to improve personalization under label scarcity.
While these methods address label distribution skew via representation learning, they often overlook degradation from poor-quality embeddings, common when scarce data cause overfitting, highlighting the need to maintain embedding quality under data scarcity.

\subsection{Contrastive Learning based FL Methods}
Early contrastive learning (CL) in FL, such as MOON~\cite{li2021model} and its variants~\cite{shi2023ffedcl,zhang2023doubly}, contrasted each local model with the global model, overlooking representation knowledge and inter-client interactions.
To improve CL efficiency in PFL, contrastive representation learning (CRL), which builds CL on data representations and captures knowledge from unlabeled data~\cite{chen2020simple,khosla2020supervised}, has been adopted. It works by minimizing distances between representations with identical labels (positive pairs) and maximizing those with different labels (negative pairs).

Some studies perform CL among local representations. For example, FedRCL~\cite{10658073} uses a relaxed supervised CL loss to mitigate inconsistent updates and stabilize aggregation for heterogeneous label distributions.
Alternatively, sharing representations among clients can yield more generalized local models. FedPCL and FedProc~\cite{tan2022federated,mu2023fedproc} employ prototypical CL, sharing local prototypes, averaged local representations, via the server to align local and global knowledge.
FedSeg~\cite{miao2023fedseg} applies pixel-level CRL for semantic segmentation under class heterogeneity, requiring high computation.
CreamFL~\cite{yu2023multimodal} extends CRL to multi-modal tasks using inter-modal and intra-modal contrasts to bridge modality and task gaps.

However, these works overlook model-level customization in PFL, which can be naturally integrated with CRL to enable more tailored local adaptations while enhancing representation quality.

\subsection{Distinguish of Our Work}

Owing to its self-supervised nature~\cite{liu2021self}, CRL can alleviate data scarcity in FL by extracting features from small-dataset clients.
However, existing studies have yet to fully account for client-level data distinctiveness in FL communications.
We propose FedCoSR, which shares label-wise knowledge among clients via the global server and integrates it, along with global model-level information, into each local model using CRL.
First, CRL is applied between local and global representations, where the latter are aggregated from label-centroid representations of all clients, bridging the representation gap.
This enables local models to learn distinctive, fine-grained features from small datasets or minority labels while maintaining coherence with the global model.
Second, the CRL loss, indicating each local model's ability to distinguish label classes, is used to adjust the aggregation ratio between global and local models, thus capturing the most relevant global information.

\section{Our Proposed Federated Contrastive Shareable Representation}\label{Sec:method}

\subsection{Problem Statement}\label{Sec:problem statement}
We consider $N$ clients, and for the $i$-th client, a local model $f_{\theta^i}$ is deployed to conduct training on a dataset $\mathcal{D}^i$, where $\theta^i$ denotes the parameter set of the $i$-th local model. 
For each sample-label pair $(\mathbf{x}^i, \mathbf{y}^i) \sim \mathcal{D}^i$, the local model $f_{\theta^i}$ maps $\mathbf{x}^i \in \mathbb{R}^d$ to a prediction $\hat{\mathbf{y}}^i = f_{\theta^i}(\mathbf{x}^i) \in \mathcal{Y}$, where $\mathcal{Y}$ denotes the target label space, and $\mathbf{y}^i$ is an one-hot vector. 
All clients share the objective of improving performance by minimizing the empirical risk over their respective local datasets:
\begin{align}
    \mathcal{F}:=\mathbb{E}_{(\mathbf{x}^i, \mathbf{y}^i)\sim \mathcal{D}^i}\mathcal{L}(\mathbf{x}^i, \mathbf{y}^i;\theta^i), \label{Eq1}
\end{align}
where $\mathcal{L}:\mathcal{Y}\times\mathcal{Y}\to\mathbb{R}$ is the loss function of the specific task to penalize the distance between $\mathbf{y}^i$ and $\hat{\mathbf{y}}^i$. The primary goal of the server is to personalize $\{\theta^i\}_{i=1}^N$ for each client to minimize $\mathcal{F}$. Thus, the global objective is to find a set of local model parameters $\Theta^*=\{\theta^{i*}\}_{i=1}^N$ that satisfy
\begin{align}
    \Theta^* = \arg \min_{\Theta^*} \frac{1}{N} \sum_{i=1}^N \mathcal{F}^i, \label{Eq3}
\end{align}
where $\mathcal{F}^i := \mathcal{F}(\theta^{i*}, \mathcal{D}^i)$ is the personalized objective of the $i$-th client.

\subsection{Representation Sharing in FL}
Heterogeneous data distributed across tasks may share a common representation despite having different labels~\cite{bengio2013representation}. Inspired by insights from~\cite{collins2021exploiting,tan2022fedproto} that representations shared among clients, e.g., shared features across many types of images or across word-prediction tasks, may provide auxiliary information without privacy intrusion, we consider utilizing shared representations to assist local adaptive aggregation and local personalized training.

Briefly, we let $f_{\theta^i}=[f_{\phi^i};f_{\pi^i}]$, where $f_{\phi^i}(\cdot)\in\mathbb{R}^{d\times k}$ is the representation layers of the $i$-th local model used to generate representations, and $f_{\pi^i}(\cdot)\in\mathbb{R}^{k}$ is the projection layers for the output, where $d$ is the data input dimension, and $k$ is the dimension of the output from $f_{\phi^i}$, known as a \textbf{representation}. 
Next, the definition of label-centroid representations is provided as follows, which is critical to the overall framework:

\begin{definition}[Label-Centroid Representations]\label{def:local_rep}

For the $i$-th client, a local label-centroid representation $\bar{\boldsymbol{\omega}}^i_c\in\mathbb{R}^k$ is the mean value of the representations with the label class $c \in \mathcal{C}^i$, where $\mathcal{C}^i$ encompasses all label classes existing in $\mathcal{D}^i$. Then, we denote $\mathcal{D}^i_c$ as the subset of $\mathcal{D}^i$ which only contains samples with the label class $c$, and $\boldsymbol{\omega}^i = f_{\phi^i}(\mathbf{x}^i) \in \mathbb{R}^k$ as the representation of $\mathbf{x}^i$. Formally, the label-centroid representation of the label class $c$ can be calculated as:
\begin{align}
    \bar{\boldsymbol{\omega}}^i_c&= \frac{1}{|\mathcal{D}^i_c|} \sum_{\mathbf{x}^i \in \mathcal{D}^i_c} \boldsymbol{\omega}^i, \quad c \in \mathcal{C}^i. \label{Eq4}    
\end{align}
Thereby, the local label-centroid representation set of the $i$-th client can be denoted as:
\begin{align}
    \bar{\boldsymbol{\Omega}}^i&=\{\bar{\boldsymbol{\omega}}^i_c|c\in\mathcal{C}^i\}. \label{Eq5}
\end{align}
\end{definition}
To obtain the final prediction, $f_{\pi^i}(\cdot)$ maps $\boldsymbol{\omega}^i$ to the label space. In other words, $f_{\theta^i}(\mathbf{x}^i)$ equals to $f_{\pi^i}(\boldsymbol{\omega}^i)$.

\begin{figure*}[htbp]
    \centering
    \includegraphics[width=0.99\textwidth]{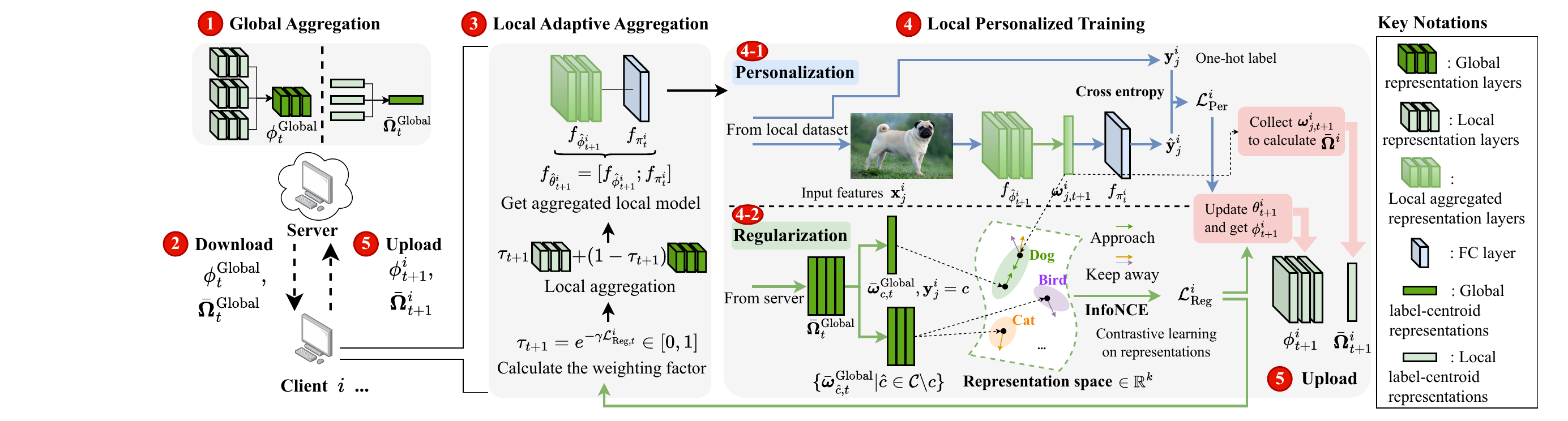}
    \caption{The overall process of the proposed FedCoSR.}
    \label{Fig:FedCoSR_process}
\end{figure*}

\subsection{Our Developed FedCoSR}\label{Sec: alg}
At each iteration, the server conducts global aggregation on local models and local representations to generate the global parameters and global representations. Then all clients download the global information for local updates. The main processes of the local update include two parts: local adaptive aggregation by loss-wise weighting and local personalized training by CRL. The overall architecture of FedCoSR is shown in Fig.~\ref{Fig:FedCoSR_process}.

\subsubsection{Global Aggregation}
Shallow layers can learn more generic information and they are more suitable for sharing~\cite{yosinski2014transferable}.
In FedCoSR, clients only send the parameters of representation layers $\phi$ to the server for aggregation and keep the parameters of projection layers $\pi$ local for maintaining personalization.
This design not only effectively integrates generic knowledge but also effectively reduces the risk of reverse engineering by dropping the last one or more layers~\cite{ghimire2022recent}. The exact number can be determined via hyperparameter tuning experiments to balance the trade-off between privacy preservation and model performance.
Unlike traditional global model aggregation~\cite{mcmahan2017communication}, the server in FedCoSR aggregates $\{\phi^i_t\}^N_{i=1}$ at iteration $t$ as follows:
\begin{align}
    \phi^{\text{Global}}_t = \sum_{i=1}^N \frac{|\mathcal{D}^i|}{|\mathcal{D}|} \phi^i_t, \label{Eq6}
\end{align}
where $\mathcal{D}=\{\mathcal{D}^i\}_{i=1}^N$ contains all clients' datasets.

Beside the parameter aggregation, all clients upload local label-centroid representations to the server for aggregation as:
\begin{align}
    \bar{\boldsymbol{\Omega}}^{\text{Global}}_t = \big\{\sum_{i \in \mathcal{N}_c} \frac{|\mathcal{D}^i_c|}{\sum_i|\mathcal{D}^i_c|} \bar{\boldsymbol{\omega}}^i_{c,t} \big| c \in \mathcal{C}\big\} \in \mathbb{R}^{C \times k},\label{Eq7}
\end{align}
where $\mathcal{N}_c$ denotes the set of clients owning samples with the label class $c$, and $\mathcal{C}= \bigcup_{i=1}^N \mathcal{C}^i$ encompasses all label classes existing in all clients. Rather than averaging, the label volume-wise weighting is adopted for the global label-centroid representation aggregation by considering the size of $\mathcal{D}^i_c$, mitigating the degradation of global knowledge aggregation caused by local label-centroid representations of scarce samples.

\subsubsection{Local Adaptive Aggregation}
When completing the global aggregation at iteration $t$, we start the local update at iteration $t+1$, which begins with the initial step of local adaptive aggregation. 
To enhance the local model's ability to learn discriminative representations, that is, to better differentiate samples with different label classes, each local model's CRL loss from iteration $t$, calculated following InfoNCE~\cite{chen2020simple} and elaborated in Section~\ref{Sec: regularization}, is applied at iteration $t+1$ to determine the proportion of aggregating the global model parameters into each local model.
Specifically, the aggregating proportion $\tau^i_{t+1}$ is determined by exponentially scaling the CRL loss value $\mathcal{L}^i_{\text{Reg},t}$ to be within $[0,1]$.
Then, the $i$-th client conducts local aggregation and obtains the aggregated local model $f_{\hat{\theta^i}_{t+1}}$ for further local training, formulated as follows:
\begin{align}
    \tau^i_{t+1} &= e^{-\gamma \mathcal{L}^i_{\text{Reg},t}} \in [0,1], \gamma>0,
    \label{Eq8} \\
    \hat{\phi}^i_{t+1} &= \tau^i_{t+1} \phi^i_t + (1-\tau^i_{t+1})\phi_t^{\text{Global}}, \label{Eq9}\\
    f_{\hat{\theta}^i_{t+1}} &= [f_{\hat{\phi}^i_{t+1}};f_{\pi^i_t}], \label{Eq9-1}
\end{align}
where $\gamma$ is a hyperparameter to control the sensitivity of scaling.
When $\mathcal{L}^i_{\text{Reg},t}$ is higher, $\tau^i_{t+1}$ becomes smaller to incorporate more parameters from the global model for knowledge enhancement. On the other hand, when $\mathcal{L}^i_{\text{Reg},t}$ is lower, the local model becomes less dependent on acquiring knowledge from the global model.
The exponential mapping in Eq. (\!\!~\ref{Eq8}) provides a smooth, sensitive, and tunable mechanism that sharply reduces the weight of a local model as its regularization loss increases, thereby amplifying the enhancement from the global model.

The reason for choosing the CRL loss as the basis for weighting aggregation, instead of the supervised loss, is that the CRL loss reflects the local model's ability to distinguish among multiple samples with different label classes.
On the other hand, although the supervised loss, such as cross entropy, can reflect the model's recognition capability for individual samples, it tends to cause severe overfitting in clients with scarce data, resulting in low training loss but poor predictive performance.

\subsubsection{Local Personalized Training}\label{Sec: regularization}

The data distributions are significantly different among clients due to heterogeneity. So it is necessary to consider both personalization for adapting patterns of local datasets and regularization for properly acquiring external information. 

\textbf{Personalization:} 
To update the local model parameters according to the local dataset, personalization is conducted following a supervised learning paradigm.
Here, cross entropy~\cite{de2005tutorial}, the most widely used loss function for classification tasks, is employed as the personalization loss function.
For the $i$-th client, we randomly divide $\mathcal{D}^i$ into batches $\mathcal{D}^i_{b,t+1}$ for training efficiency, each of which contains $b$ samples.
Formally, the expectation of cross entropy $H(\cdot)$ is calculated as the personalization loss function $\mathcal{L}^i_{\text{Per}}$ at iteration $t+1$:
\begin{subequations}
\label{Eq: local loss}
\begin{align}
    &\mathcal{L}^i_{\text{Per},t+1} := \mathbb{E}_{\mathcal{D}^i_{b,t+1}\sim\mathcal{D}^i} \Big[\mathbb{E}_{(\mathbf{x}^i_j,\mathbf{y}^i_j) \sim \mathcal{D}^i_{b,t+1}} \big[H(\mathbf{y}^i_j,\hat{\mathbf{y}}^i_j)\big]\Big], \\
    &H(\mathbf{y}^i_j,\hat{\mathbf{y}}^i_j) =-\sum_{c=1}^C y^i_{j,c} \log (\hat{y}^i_{j,c}),  \\
    &\mathbf{y}^i_j=(y^i_{j,c})_{c=1}^C,\quad \hat{\mathbf{y}}^i_j= f_{\hat{\theta}^i_{t+1}}(\mathbf{x}^i_j) =(\hat{y}^i_{j,c})_{c=1}^C,
\end{align}
\end{subequations}
where $y^i_{j,c}$ denotes the $c$-th entry of the one-hot label for the $j$-th sample from batches on the $i$-th client.
Through personalization, each client can focus on local patterns and adjust the distance between $\theta^i$ and $\theta^{\text{Global}}$.

\textbf{Regularization:} To utilize useful external information while maintaining the local personalization, we adopt CRL on representations for regularization.

Inspired by~\cite{chen2020simple}, CRL, an ML technique aiming to learn representations by contrasting positive pairs (similar samples) against negative pairs (dissimilar samples), can be utilized between the local representations and the global label-centroid representations to assist personalization. Specifically, at iteration $t+1$, the $i$-th client receives the global label-centroid representations $\bar{\boldsymbol{\Omega}}^{\text{Global}}_t$ from the server. We assume that the labels unseen to $\mathcal{D}^i$ can provide external knowledge for $f_{\theta^i}$. We let $b$ be the batch size, and representations at iteration $t+1$ can be obtained by forwarding $\{\mathbf{x}^i_j\}_{j=1}^b$ towards $f_{\hat{\phi}^i_{t+1}}$:
\begin{align}
    \boldsymbol{\Omega}^i_{t+1}=\{\boldsymbol{\omega}^i_{j,t+1}\}_{j=1}^b.
\end{align}

Then, we construct positive pair and negative pairs for the $\boldsymbol{\omega}^i_{j,t+1}$ by $\bar{\boldsymbol{\omega}}^{\text{Global}}_{c,t}$ and \{$\bar{\boldsymbol{\omega}}^{\text{Global}}_{\hat{c},t}|\hat{c} \in \mathcal{C} \backslash c\}$, respectively, where the label class of $\boldsymbol{\omega}^i_{j,t+1}$ is $c$.

\begin{definition}[Positive and Negative Pairs of Representations]\label{def:rep_CL}
Intuitively, for $\boldsymbol{\omega}^i_{j,t+1}$ whose label class is $c$, we consider the global label-centroid representation $\bar{\boldsymbol{\omega}}^{\text{Global}}_{c,t}$ as the positive sample, which has the same label class. On the other hand, the remaining global label-centroid representations with different label classes are considered as negative samples. Thereby, one positive pair $p^+_{j,t+1}$ and $|\mathcal{C}|-1$ negative pairs $\{p^-_{j,t+1}\}$ are constructed as:
\begin{subequations}
\label{Eq10}
\begin{align}
    p^+_{j,t+1} &= (\boldsymbol{\omega}^i_{j,t+1}, \bar{\boldsymbol{\omega}}^{\text{Global}}_{c,t}), \\
    \{p^-_{j,t+1}\} &= \{(\boldsymbol{\omega}^i_{j,t+1}, \bar{\boldsymbol{\omega}}^{\text{Global}}_{\hat{c},t})|\hat{c} \sim \mathcal{C} \backslash c\}, \\
    y^i_j=c, & \quad i\in\{1,...,N\}, \quad j \in \{1,...,b\}. 
\end{align}
\end{subequations}
\end{definition}


We adopt InfoNCE~\cite{chen2020simple} as the form of regularization loss function, aiming to maximize the similarity of positive pairs and minimize the similarity of negative pairs, 
InfoNCE offers good gradient stability and enables efficient utilization of negative samples.
Let $D(\cdot)$ denote cosine similarity. For the $i$-th client, according to Definition.~\ref{def:rep_CL}, the regularization loss can be expressed as:
\begin{subequations}
\label{Eq11}
\begin{align}
    &\mathcal{L}^i_{\text{Reg},t+1} := \mathbb{E}_{\mathcal{D}^i_{b,t+1}\sim\mathcal{D}^i}\mathbb{E}_{(\mathbf{x}^i_j,\mathbf{y}^i_j) \sim \mathcal{D}^i_{b,t+1}} \notag\\
    & \Bigg[-\log \frac{e^{\big[D(p^+_{j,t+1})/\tau_{\text{CL}}\big]}}{e^{\big[D(p^+_{j,t+1})/\tau_{\text{CL}}\big]} + \sum_{\hat{c} \in \mathcal{C}\backslash c}e^{\big[D(p^-_{j,t+1})/\tau_{\text{CL}}\big]}} \Bigg], \\
    &D(\boldsymbol{\omega}^i_{j,t+1}, \bar{\boldsymbol{\omega}}^{\text{Global}}_{c,t}) = \frac{\boldsymbol{\omega}^i_{j,t+1} \cdot \bar{\boldsymbol{\omega}}^{\text{Global}}_{c,t}}{||\boldsymbol{\omega}^i_{j,t+1}||_2||\bar{\boldsymbol{\omega}}^{\text{Global}}_{c,t}||_2} \in [-1,1], 
\end{align}
\end{subequations}
where $\tau_{\text{CL}}$ is the temperature of CRL which controls the attention to positive or negative samples. 
Then, the $i$-th client calculates its local label-centroid representations $\bar{\boldsymbol{\Omega}}^i_{t+1}$ referring to Eq. (\!\!~\ref{Eq5}).

\textbf{Final Objective:}

As a result, we obtain the final objective function for locally updating $\theta^i$ at iteration $t$ according to Eq. (\!\!~\ref{Eq: local loss}) and Eq. (\!\!~\ref{Eq11}):
\begin{align}
    \hspace{-0.5em} \mathcal{L}^i_{t+1} := \mathcal{L}^i_{\text{Per},t+1}(\mathcal{D}^i;\hat{\theta}^i_{t+1}) + 
    \alpha \mathcal{L}^i_{\text{Reg},t+1}(\boldsymbol{\Omega}^i_{t+1}, \bar{\boldsymbol{\Omega}}^{\text{Global}}_t). \label{Eq12}
\end{align}
The process of local updates can be expressed as follows:
\begin{equation}
\begin{aligned}
    \theta^i_{t+1} \gets \hat{\theta}^i_{t+1} - \eta \nabla_{\hat{\theta}^i_{t+1}} \mathcal{L}^i_{t+1}(\mathcal{D}^i;\hat{\theta}^i_{t+1};\boldsymbol{\Omega}^i_{t+1},\bar{\boldsymbol{\Omega}}^{\text{Global}}_t),\label{Eq13}
\end{aligned}
\end{equation}
where $\alpha$ is the hyperparameter for trading off personalization and regularization, and $\eta$ is the learning rate.

\renewcommand{\baselinestretch}{1}
\begin{algorithm}[t]
\scriptsize
\caption{FedCoSR Framework}\label{alg1}
\begin{algorithmic}[1]
\STATE {\bfseries Input:} The server and $N$ clients; $\{\mathcal{D}^i\}_{i=1}^N$: datasets of $N$ clients; $\theta^{\text{Global}}_0$: the initial global model; $\eta$: the learning rate; $\alpha$: the trade-off factor between personalization and regularization.
\STATE {\bfseries Output:} $\theta^{1*}, ..., \theta^{N*}$: Optimal local model parameters.
\STATE {\bfseries Initialization:}
\STATE The server sends $\theta^{\text{Global}}_0$ to all clients to initialize $\{\hat{\theta}^i_0\}_{i=1}^N$ by overwriting, rather than local aggregation in Eq. (\!\!~\ref{Eq9}).
\STATE Clients train $\{\hat{\theta}^i_0\}_{i=1}^N$ by Eq. (\!\!~\ref{Eq13}) in parallel, where $\alpha=0$. Then clients get $\{\theta^i_1\}_{i=1}^N$.
\STATE Clients collect $\{\bar{\boldsymbol{\Omega}}^i_1\}_{i=1}^N$ by Eq. (\!\!~\ref{Eq4}) and Eq. (\!\!~\ref{Eq5}).
\STATE Clients upload $\{\phi^i_1\}_{i=1}^N$ and $\{\bar{\boldsymbol{\Omega}}^i_1\}_{i=1}^N$ to the server.
\STATE {\bfseries FL Communication:}
\STATE \textbf{for} iteration $t=1,...,T$ \textbf{do}
\STATE \hspace{0.35cm} {\bfseries Server:} \hfill $\blacktriangleright$ \circled{1} \textbf{Global aggregation}
\STATE \hspace{0.35cm} The server aggregates $\{\phi^i_t\}_{i=1}^N$ to get $\phi^{\text{Global}}_t$ by Eq. (\!\!~\ref{Eq6}).
\STATE \hspace{0.35cm} The server aggregates $\{\bar{\boldsymbol{\Omega}}^i_t\}_{i=1}^N$ to get $\bar{\boldsymbol{\Omega}}^{\text{Global}}_t$ by Eq. (\!\!~\ref{Eq7}).
\STATE \hspace{0.35cm} The server sends $\phi^{\text{Global}}_t$ and $\bar{\boldsymbol{\Omega}}^{\text{Global}}_t$ to all clients. \hfill $\blacktriangleright$ \circled{2}
\STATE \hspace{0.35cm} \textbf{Clients:}
\STATE \hspace{0.35cm} \textbf{for} the $i$-th client in parallel \textbf{do} \hfill $\blacktriangleright$ \textbf{Local Update}
\STATE \hspace{0.7cm} {\bfseries Local Aggregation:} \hfill $\blacktriangleright$ \circled{3}
\STATE \hspace{0.7cm} Calculate $\tau^i_{t+1}$ based on $\mathcal{L}^i_{\text{Reg},t}$ by Eq. (\!\!~\ref{Eq8}).
\STATE \hspace{0.7cm} Obtain local aggregated model $f_{\hat{\theta}^i_{t+1}}$ by Eq. (\!\!~\ref{Eq9-1}).

\STATE \hspace{0.7cm} {\bfseries Local Personalized Training:} \hfill $\blacktriangleright$ \circled{4}
\STATE \hspace{0.7cm} Construct positive pairs and negative pairs by Eq. (\!\!~\ref{Eq10}).
\STATE \hspace{0.7cm} Calculate $\mathcal{L}^i_{t+1}$ by Eq. (\!\!~\ref{Eq12}) for both \textbf{personalization} and \textbf{regularization}.
\STATE \hspace{0.7cm} Train $\hat{\theta}^i_{t+1}$ by Eq. (\!\!~\ref{Eq13}) and clip it into $[0,1]$ for normalization to get $f_{\theta^i_{t+1}}=[f_{\phi^i_{t+1}};f_{\pi^i_{t+1}}]$.
\STATE \hspace{0.7cm} {\bfseries Upload:} \hfill $\blacktriangleright$ \circled{5}
\STATE \hspace{0.7cm} Collect $\bar{\boldsymbol{\Omega}}^i_{t+1}$ by Eq. (\!\!~\ref{Eq4}) and Eq. (\!\!~\ref{Eq5}).
\STATE \hspace{0.7cm} Upload $\phi^i_{t+1}$ and $\bar{\boldsymbol{\Omega}}^i_{t+1}$ to the server.
\STATE \hspace{0.35cm} \textbf{end for}
\STATE \textbf{end for}
\STATE \textbf{return} $\theta^{1*}, ..., \theta^{N*}$.
\end{algorithmic}
\end{algorithm}
\renewcommand{\baselinestretch}{1.0}

Algorithm~\ref{alg1} presents the entire training process of FedCoSR, including 1) global aggregation for both model parameters and label-centroid representations (line 11-13); 2) local aggregation between each model and the global model (line 17-18); and 3) local training on the aggregated local model (line 20-22). 4) All local information is uploaded to the server (line 24-25). This loop is ended until all the optimal local parameters are found.

\section{Theoretical Analysis}\label{Sec:theoretical}
Before conducting experiments, we provide a theoretical analysis of the developed FedCoSR algorithm. Firstly, we focus on the effectiveness of the local loss function incorporating cross entropy and InfoNCE, as the local training is the core part of the personalization of FedCoSR. We explain that FedCoSR minimizes InfoNCE to maximize the mutual information between the local representations and global label-centroid representations, thus enhancing the distinguishing capability of each local model. Secondly, both global model aggregation and local model aggregation linearly change the model parameters, which may cause deviations in the loss expectation during each iteration. If the deviation remains unbounded, the convergence of FedCoSR may not be guaranteed. Therefore, we characterize the overall communication between the server and the clients to establish an upper bound on the loss expectation deviation, thereby ensuring the convergence of FedCoSR. Note that due to the non-convexity of the local loss induced by InfoNCE, we focus on non-convex setting. Moreover, we also derive the relationship between the convergence rate and key hyperparameters.

\subsection{Effectiveness of Local Loss Function}
Referring to Eq. (\ref{Eq12}), the local loss function is a linear combination of $\mathcal{L}_{\text{Per}}$ and $\mathcal{L}_{\text{Reg}}$. Since $\mathcal{L}_{\text{Per}}$ is in the form of cross entropy, its convexity and good convergence are known~\cite{de2005tutorial}. But the non-convexity of InfoNCE $\mathcal{L}_{\text{Reg}}$ may lead to sub-optimum of the training objective for each local model. Thus, we focus on analyzing $\mathcal{L}_{\text{Reg}}$ to demonstrate effectiveness of the local loss function.

To quantify the enhancement brought by the InfoNCE loss $\mathcal{L}_{\text{Reg}}$, we introduce the manifestation of mutual information in contrastive learning as follows.

\begin{definition}[Mutual Information in Contrastive Learning]\label{eq:loss-def-1}
To construct contrastive learning task, mutual information $I(\cdot)$ is introduced between anchor samples $X$ and the similar samples $X^+$ which is also known as positive samples:
\begin{align}
    I(X^+; X) = \sum_{x^+\in X^+,x \in X} p(x^+,x) \log \bigg[\frac{p(x^+|x)}{p(x^+)}\bigg], \label{eq:loss-def-eq2}
\end{align}
where $p(\cdot)$ is the notation of probability.
\end{definition}

Based on Definition~\ref{eq:loss-def-1}, we present the following theorem to illustrate the effectiveness of our local loss function design.

\begin{theorem}[InfoNCE Minimization in FedCoSR]\label{eq:theo-loss-1}
For each data batch of the $i$-th client, minimizing InfoNCE equals to maximizing the mutual information between each anchor representation in this batch and its positive representation, and meanwhile minimizing the mutual information between it and its negative representations, formulated as follows, where $b$ is the batch size:
\end{theorem}
\begin{equation}
\begin{aligned}
    \mathcal{L}^i_{\text{Reg},t+1} \geq - \frac{1}{b} \sum_{j=1}^b I(\bar{\boldsymbol{\omega}}^{\text{Global}}_{c,t}, \boldsymbol{\omega}^i_j) + \log (C).
\end{aligned}
\end{equation}

Intuitively, we have $\bar{I}(\cdot) \geq \log (C) - \mathcal{L}_{\text{Reg}}$, where $\bar{I}(\cdot)$ is the mean value of the mutual information. When $C$ becomes larger, the lower bound of similar representations increases, improving the performance of the $i$-th local model.
Theorem~\ref{eq:theo-loss-1} indicates that optimizing the regularization term $\mathcal{L}^i_{\text{Reg}}$ can enlarge the information acquisition for the $i$-th client, thereby confirming the effectiveness of combining $\mathcal{L}_{\text{Per}}$ and $\mathcal{L}_{\text{Reg}}$.

\subsection{Convergence of FedCoSR}
The convergence conditions for the $i$-th client is explained in this part. For ease of discussion, we denote the total number of local training epochs as $R$, which is set to 1 in this paper. Specifically, we define $tR+r$ as the $r$-th epoch at iteration $t$, $tR$ as the end of iteration $t$ (end of the local training), $tR+0$ as the beginning of iteration $t$ (local aggregation), and $tR+\frac{1}{2}$ as the utilization of global label-centroid representations at iteration $t$ (after local aggregation). 

Before showing the convergence of FL communication, we make three assumptions which are widely used in literature: Assumption~\ref{ass: 1}: Lipschitz Smoothness ensuring consistency in gradient changes~\cite{fallah2020personalized,li2020federated,huang2021personalized,tan2022fedproto}, Assumption~\ref{ass: 3}: Unbiased Gradient and Bounded Variance providing stable gradient estimates~\cite{fallah2020personalized,li2020federated,tan2022fedproto}, and Assumption~\ref{ass: 4}: Bounded Variance of Representation Layers guaranteeing an acceptable variance between the global model and each local model.

\begin{assumption}[Lipschitz Smoothness]\label{ass: 1}
The $i$-th local loss function is $L_1$-Lipschitz smooth, leading to that the gradient of the local loss function is $L_1$-Lipschitz continuous, where $L_1>0$, $\forall r_1, r_2 \in \{0,\frac{1}{2},1,...,R\}$, and $(\mathbf{x}^i,\mathbf{y}^i)\in\mathcal{D}^i$:
\begin{align}
        ||\nabla \mathcal{L}^i_{tR+r_1}(\mathbf{x}^i,\mathbf{y}^i;\hat{\theta}^i_{tR+r_2}) - \nabla \mathcal{L}^i_{tR+r_2}(\mathbf{x}^i,\mathbf{y}^i;\hat{\theta}^i_{tR+r_2})||_2  \notag\\
        \leq L_1||\hat{\theta}^i_{tR+r_1} - \hat{\theta}^i_{tR+r_2}||_2.
\end{align}
\end{assumption}

\begin{assumption}[Unbiased Gradient and Bounded Variance]\label{ass: 3}
The stochastic gradient $\nabla\mathcal{L}^i_{tR}(\hat{\theta}^i_{tR};\mathcal{D}^i_{b,tR})$ is an unbiased estimator of the local gradient,
and the variance of $\nabla\mathcal{L}^i_{tR}(\hat{\theta}^i_{tR};\mathcal{D}^i_{b,tR})$ is bounded by $\sigma$:
\begin{align}
    \text{Var}[\nabla \mathcal{L}^i_{tR}(\hat{\theta}^i_{tR};\mathcal{D}^i_{b,tR})] \leq \sigma^2.
\end{align}
\end{assumption}

\begin{assumption}[Bounded Variance of Representation Layers]\label{ass: 4}
The variance between $f^i_{tR}$ and $f^{\text{Global}}_{tR}$ is bounded, whose parameter bound is:
\begin{align}
    \mathbb{E}[||f^i_{tR}-f^{\text{Global}}_{tR}||^2_2] \leq \varepsilon^2.
\end{align}
\end{assumption}

Based on the above assumptions, the deviation in loss expectation for each iteration is bounded, as shown in Theorem~\ref{FL-theo1}, serving as the foundation for proving our algorithm's convergence.

\begin{theorem}[One-Iteration Deviation]\label{FL-theo1}
For the $i$-th client, between the iteration $t$ and the iteration $t+1$, we have:
\begin{align}
    \mathbb{E}[\mathcal{L}^i_{(t+1)R+\frac{1}{2}}] & \leq \mathbb{E}[\mathcal{L}^i_{tR+\frac{1}{2}}] - \big(\eta-\frac{L_1\eta^2}{2}\big)\sum_{r=\frac{1}{2}}^R||\nabla\mathcal{L}^i_{tR+r}||_2^2 \notag \\
    &+ \frac{R L_1\eta^2 \sigma^2}{2} + \frac{L_1(\varepsilon^2 + \varepsilon)}{2}+\frac{2\alpha}{\tau_{\text{CL}}}.
\end{align}    
\end{theorem}

Theorem~\ref{FL-theo1} indicates that, the deviation in the loss expectation for the $i$-th client is bounded from the iteration $t$ to the iteration $t+1$, which leads to the following corollary, establishing the convergence of FedCoSR in non-convex settings.

\begin{corollary}[Non-Convex FedCoSR Convergence]\label{app:FL-coro1}
The loss function of the $i$-th client monotonously decreases between the iteration $t$ and the iteration $t+1$, when the learning rate at iteration $r'$ satisfies:
\begin{align}
     \eta_{r'} < \frac{\mathbb{S} + \sqrt{\big[\mathbb{S}\big]^2 - \frac{(L_1 \mathbb{S} + RL_1\sigma^2)(L_1\varepsilon^2\tau_{\text{CL}} + L_1 \varepsilon\tau_{\text{CL}}+4\alpha)}{\tau_{\text{CL}}}} }{L_1 \mathbb{S} + RL_1\sigma^2},
\end{align}
where $r'=\frac{1}{2}, 1, ..., R$, and briefly, $\mathbb{S}=\sum^{r'}_{r=\frac{1}{2}} ||\nabla \mathcal{L}^i_{tR+r}||^2_2$.
\end{corollary}

Corollary~\ref{app:FL-coro1} indicates that as long as the learning rate is small enough at iteration $r'$, FedCoSR will converge. Furthermore, the convergence rate of FedCoSR can be also obtained as follows.

\begin{theorem}[Non-Convex Convergence Rate of FedCoSR]\label{FL-theo2}
Given any $\epsilon > 0$, after $T$ iterations, the $i$-th client converges with the rate:
\begin{align}
    \frac{1}{TR} \sum^T_{t=1}\sum^R_{r=\frac{1}{2}} \mathbb{E}[||\nabla \mathcal{L}^i_{tR+r}||^2_2]< \epsilon,
\end{align}
when
{\scriptsize
\begin{align}
    T &> \frac{2\tau_{\text{CL}} (\mathcal{L}^i_\frac{1}{2} - \mathcal{L}^{i,*}) }{(2\eta-L_1\eta^2)\epsilon\tau_{\text{CL}}R - L_1\eta^2\sigma^2\tau_{\text{CL}}R - \tau_{\text{CL}}L_1(\varepsilon^2-\varepsilon) - 4\alpha}, \\
    \eta &< \frac{2\epsilon\tau_{\text{CL}}R + \sqrt{4\epsilon^2\tau_{\text{CL}}^2R^2 - 4 L_1\tau_{\text{CL}}R(\epsilon+\sigma^2)(\tau_{\text{CL}}L_1(\varepsilon^2+\varepsilon)+4\alpha)}}{2L_1\tau_{\text{CL}}R(\epsilon+\sigma^2)}, \\
    \alpha &< \frac{\epsilon^2}{4L_1(\epsilon+\sigma^2)} - \frac{\tau_{\text{CL}}L_1}{4}(\varepsilon^2+\varepsilon).
\end{align}
}
\end{theorem}

Theorem~\ref{FL-theo2} outlines the specific conditions for convergence. To ensure the algorithm converging with a rate, the minimum number of training iterations $T$ should be determined. Correspondingly, the upper bounds for two hyperparameters, $\eta$ and $\alpha$, are also presented. 

\section{Experiments and Discussion}\label{Sec:simulation}

\subsection{Experiment Setup}

\subsubsection{Dataset Description}
In our experimental setup, we simulate a realistic image-based FL scenario, where, clients represent distributed devices, e.g., smartphones or surveillance cameras, collecting images from diverse, real-world environments. This setup leads to two types of statistical heterogeneity: 1) Label distribution skew due to differences in local image contents, such as urban versus rural scenes; and 2) Data scarcity, resulting from varied image capture frequencies.
The server functions as a central aggregator that collects local model updates, while each client trains its model using its own data and hyperparameters.

We consider three popular datasets of image classification for evaluation: CIFAR-10 consists of 10 categories of items, each containing 6,000 images; EMNIST consists of 47 categories of handwritten characters, each containing 2,400 images; and CIFAR-100 consists of 100 categories of items, each containing 600 images. 
Based on these datasets, we evaluate the performance of our method in tasks with different scales of label classes.

\subsubsection{Heterogeneity Setting on Datasets}
We simulate label distribution skew and data scarcity with two widely adopted settings. 
The first setting is informed by~\cite{zhang2023fedala}, called practical setting, using the Dirichlet distribution $Dir(\beta)$, where $\beta \in (0,1]$. We set $\beta=0.1$ as the default value, since smaller $\beta$ results in more heterogeneous simulations.
The second setting is the pathological setting~\cite{mcmahan2017communication,shamsian2021personalized}, which samples 2, 10, and 20 label categories from CIFAR-10, EMNIST, and CIFAR-100, respectively.
While both settings can lead to differences in label distributions among clients, the difference is: in the practical setting, the Dirichlet distribution controls the proportion of labels assigned to each client, with varying levels of skewness based on the value of $\beta$, allowing all clients to potentially receive all labels but in different proportions. In contrast, the pathological setting enforces a hard limit on the number of label categories each client can receive, leading to more extreme label distribution heterogeneity. 
Thus, the practical setting results in more gradual label distribution shifts, while the pathological setting imposes rigid category constraints.

\begin{figure}[hbpt]
    \centering   
    \subfigure{\label{Fig:dist_prac_cifar10}
    \includegraphics[width=.22\textwidth]{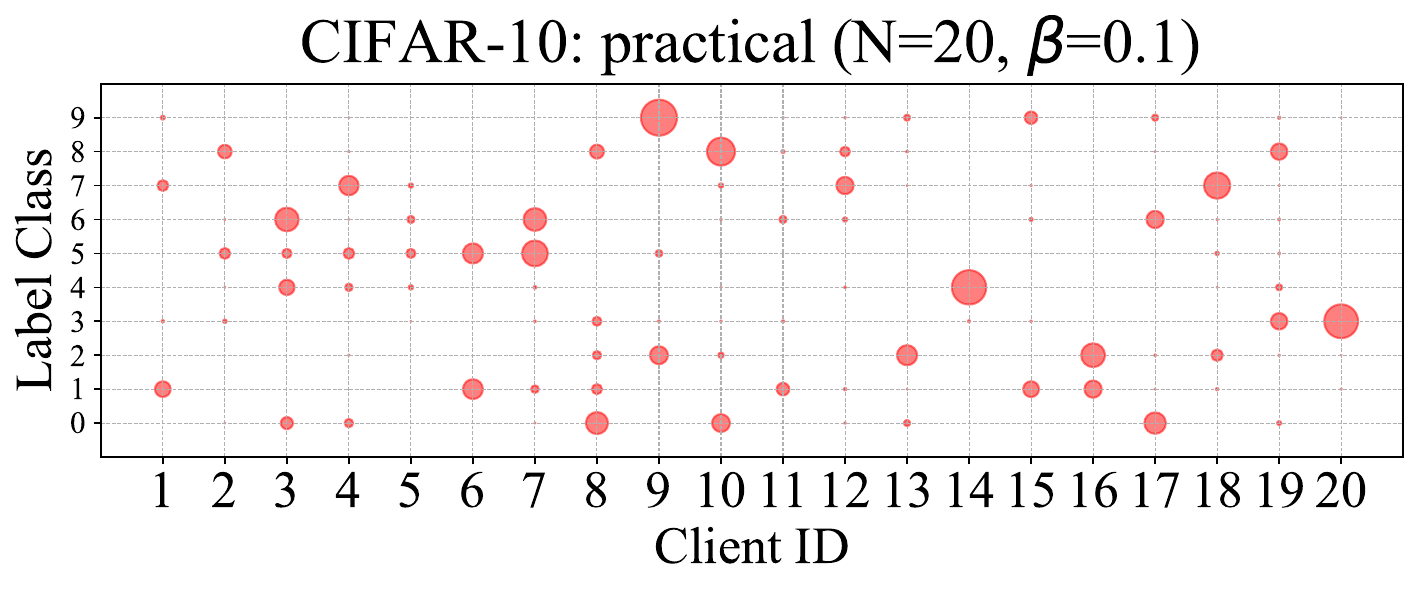}  }
    \subfigure{\label{Fig:dist_path_cifar10}
    \includegraphics[width=.22\textwidth]{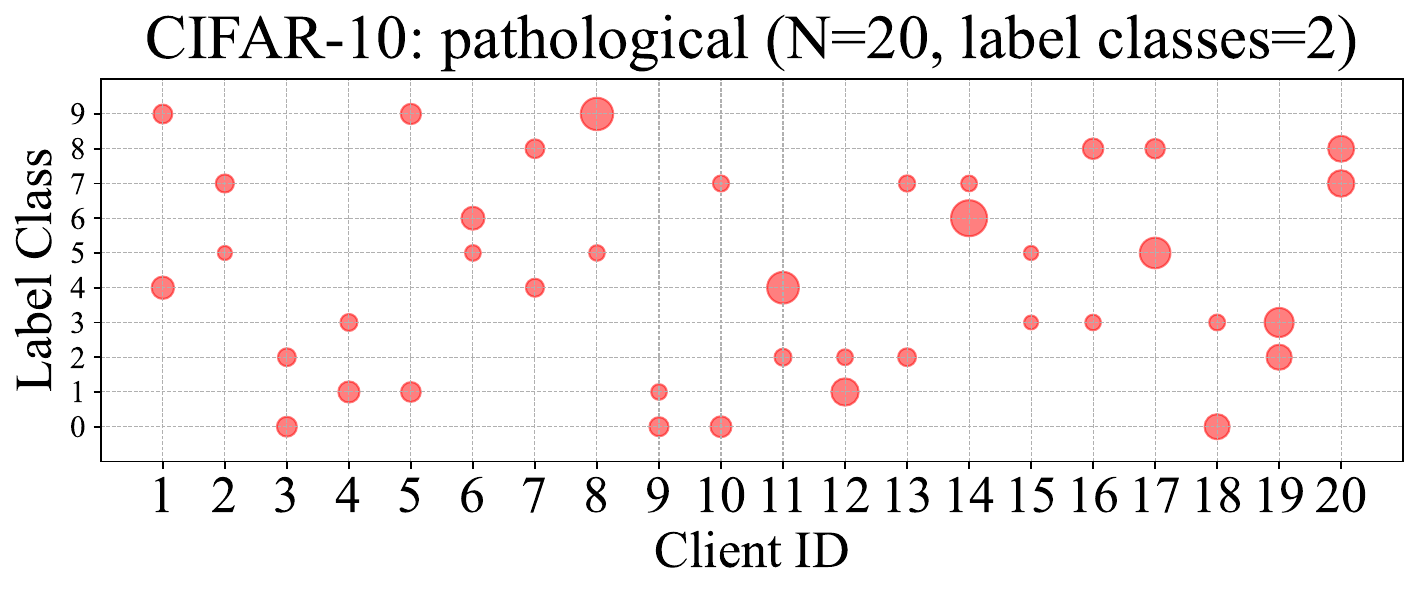}  }
    \subfigure{\label{Fig:dist_prac_emnist}
    \includegraphics[width=.22\textwidth]{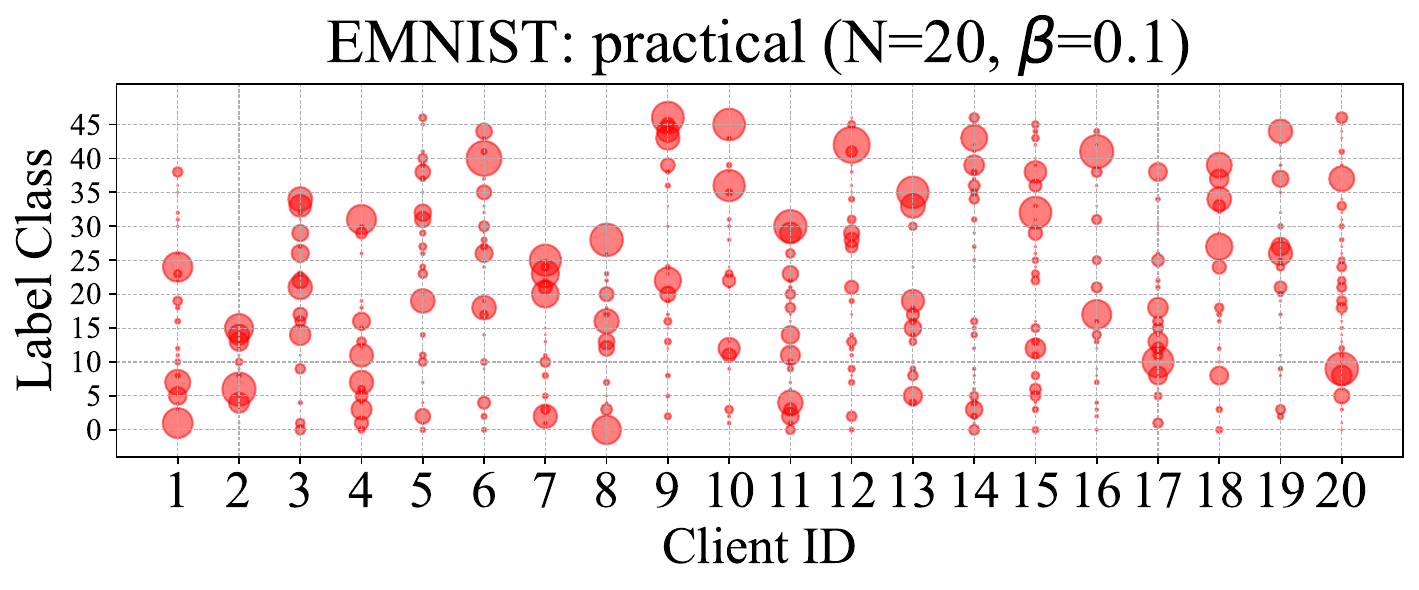}  }
    \subfigure{\label{Fig:dist_path_emnist}
    \includegraphics[width=.22\textwidth]{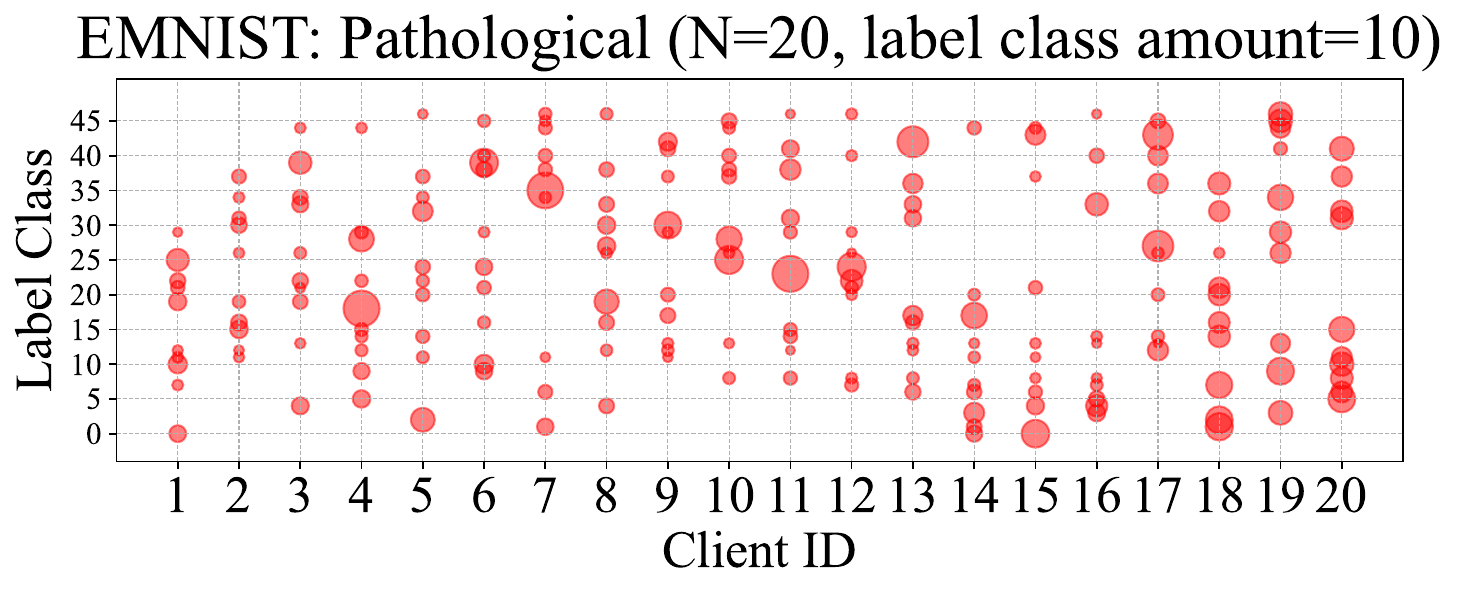}  }
    \subfigure{\label{Fig:dist_prac_cifar100}
    \includegraphics[width=.22\textwidth]{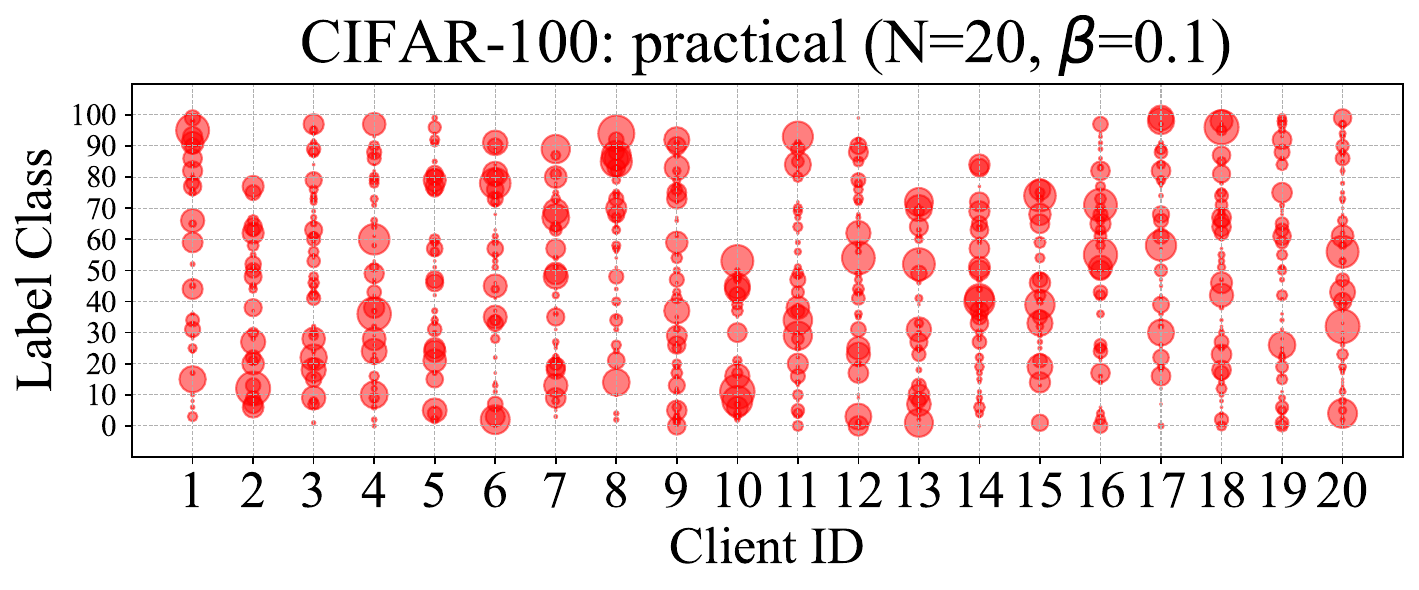}  }
    \subfigure{\label{Fig:dist_path_cifar100}
    \includegraphics[width=.22\textwidth]{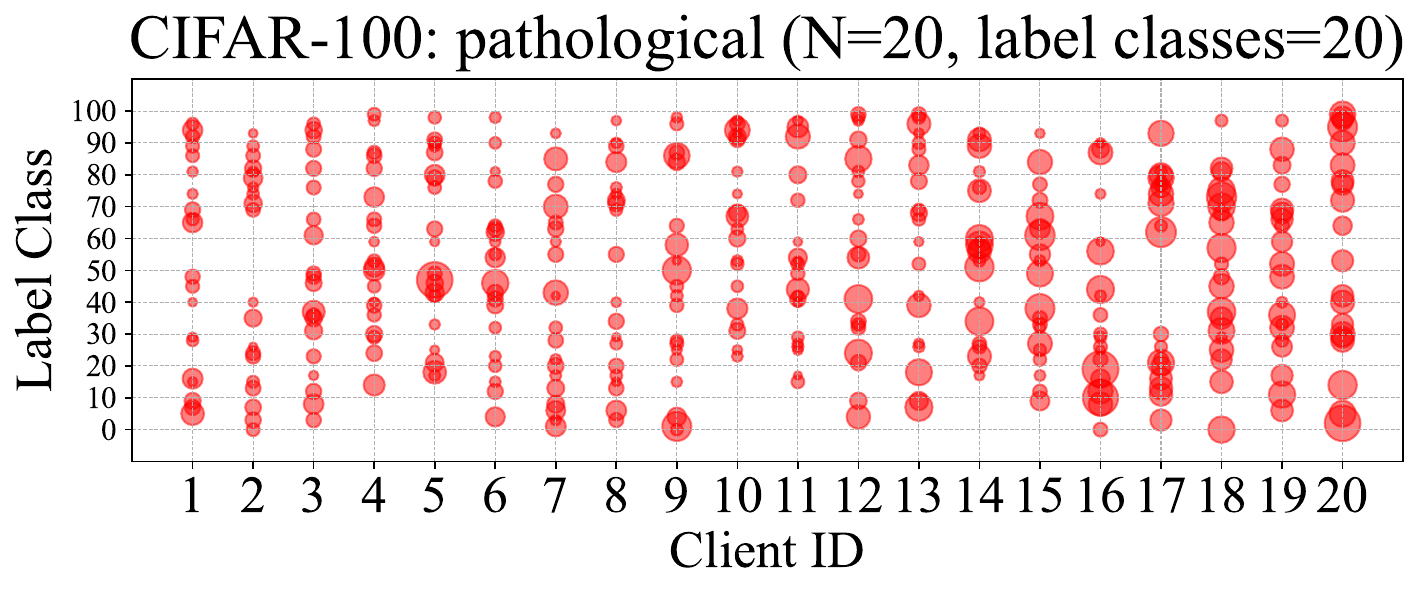}  }
    \captionsetup{width=.99\linewidth}
    \caption{The data distributions of CIFAR-10, EMNIST, and CIFAR-100 in the default settings.}
    \label{Fig:dist_cifar100_default}
\end{figure}

For evaluation, we ensure that all clients receive datasets of approximately similar sizes, which are about from 8,000 to 10,000 samples. The dataset is split into 75 \% for training and 25 \% for testing.
Fig.~\ref{Fig:dist_cifar100_default} shows the data distribution visualization of the default settings for the three datasets. The corresponding results are presented and analyzed in Section.~\ref{sec: robustness skew 1} and Section.~\ref{sec: robustness skew}.

\subsubsection{Scarcity Setting on Datasets}\label{sec: scarcity setting}

To evaluate the performance of the proposed method under data scarcity, we design the following two experiments based on the default heterogeneity settings:
\begin{itemize}
    \item \textbf{Fairness for New Participation:} Under the default CIFAR-10 settings where 20 clients are set, we reduce the data size of the last five clients to 10\% of their original amounts while keeping the data distribution unchanged.
    \item \textbf{Robustness against Model Degradation:} Additionally, to explore the robustness of the FL algorithms in a scenario where all clients' data are scarce, we reduce the data size for each client to between 5\% and 25\%, while maintaining the same data distribution.
\end{itemize}

The corresponding results are presented and analyzed in Section.~\ref{sec: fairness} and Section.~\ref{sec: robustness scarce}.

\subsubsection{Baselines for Comparison}

We compare FedCoSR with individual local training and 20 popular FL methods, as categorized in Table~\ref{tab:taxonomy}, where \ding{115} means fairness among clients is discussed but data scarcity is not considered. 

\begin{table}[!t]
\centering
\scriptsize
\caption{Key features of comparison baselines, where P1 and P2 represent label distribution skew problem and data scarcity problem, respectively. \ding{51} and \ding{55} indicate that the corresponding problem is studied or not. \ding{115} indicates that fairness among clients is discussed but data scarcity is not considered.}
\label{tab:taxonomy}
\begin{tabular}{ccccc}
\hline\hline
\textbf{Algorithm} & \textbf{Type}      & \textbf{Main technique}     & \textbf{P1} & \textbf{P2}                                        \\ \hline
FedAvg~\cite{mcmahan2017communication}               & FL                 &        Standard       & \ding{55} &  \ding{55}                        \\
FedProx~\cite{li2020federated}                     & FL                   & Regularization    & \ding{55} & \ding{55}         \\
MOON~\cite{li2021model}    &   FL    &   Contrastive learning  & \ding{51} & \ding{55} \\
FedRCL~\cite{10658073}                    & FL                   &  Contrastive learning   &  \ding{51}  &  \ding{55}    \\

pFedMe~\cite{t2020personalized}                   & PFL                   &  Regularization  & \ding{51} & \ding{55}          \\
Ditto~\cite{li2021ditto}                   & PFL                   &  Multi-task learning    & \ding{55} & \ding{115}         \\

PerFedAvg~\cite{fallah2020personalized}                   & PFL                   &  Meta-learning   &    \ding{51} & \ding{55}         \\
PeFLL~\cite{scott2024pefll}                    & PFL                   &  Meta-learning   &  \ding{55}  &  \ding{51}  \\
FedRep~\cite{collins2021exploiting}                    & PFL                   &  Model splitting      & \ding{51} & \ding{55}       \\
LG-FedAvg~\cite{liang2020think}                     & PFL                   &  Model splitting   & \ding{51} &   \ding{55}         \\
FedPer~\cite{arivazhagan2019federated}                     & PFL                   &  Model splitting    & \ding{51} & \ding{115}        \\

FedAMP~\cite{huang2021personalized}                    & PFL                   &  Model collaboration    &  \ding{51} & \ding{55}         \\
FedALA~\cite{zhang2023fedala}                    & PFL                   &  Local aggregation    &  \ding{51}  &  \ding{55}         \\
FedAS~\cite{10655727}                    & PFL                   &  Local aggregation   &  \ding{55}  &  \ding{115}          \\
FedProto~\cite{tan2022fedproto}                    & PFL                   &  Representation sharing     & \ding{51}  &  \ding{115}        \\
FedPAC~\cite{xu2022personalized}                    & PFL                   &  Representation sharing    & \ding{51} & \ding{55}         \\
FedGH~\cite{FedGH}                    & PFL                   &  Representation sharing   &  \ding{51}  &  \ding{55}          \\   pFedFDA~\cite{mclaughlin2025personalized}                    & PFL                  &  Representation sharing   &  \ding{55}  &  \ding{51}          \\
FedDBL~\cite{10572001}                    & PFL                   &  Representation sharing   &  \ding{55}  &  \ding{115}          \\
FedStream~\cite{10198520}                    & PFL                   &  Representation sharing   &  \ding{51}  &  \ding{55}          \\\hline
\textbf{Proposed FedCoSR}                 & \textbf{PFL}                   &  \textbf{Representation sharing}   &  \ding{51}  &  \ding{51}          \\

\hline\hline
\end{tabular}
\end{table}

\subsubsection{Other Settings}
We implement all experiments using PyTorch-1.12 on an Ubuntu 18.04 server with two Intel Xeon Gold 6142M CPUs with 16 cores, 24G memory, and one NVIDIA 3090 GPU. 
For simplicity, we construct a two-layer Convolutional Neural Network (CNN) followed by two Fully-Connected (FC) layers for all datasets. 
In FedCoSR, the representation layers $\phi$ consists of the two-layer CNN and the first FC layer, and the second FC layer is the projection layer $\pi$. Additionally, for reliability, five-time experiments are conducted to calculate the mean and standard deviation, where the mean reflects the overall accuracy, and the standard deviation quantifies the fairness.

\subsection{Result Analysis and Discussion on Heterogeneity}

\subsubsection{Label Distribution Skew-Effectiveness}\label{sec: robustness skew 1}

\begin{table*}[!t]
\renewcommand{\arraystretch}{1.1}
\centering
\setlength{\tabcolsep}{5pt}
\scriptsize
\caption{The accuracy and standard deviations of the three datasets in the practical and pathological heterogeneous setting. We use superscripts $*$, $\dagger$, and $\ddagger$, to emphasize the 1st, 2nd, and 3rd best values in each column, respectively. \textcolor{darkgreen}{Green} means better and \textcolor{red}{red} means worse than the averaged result. The names of the newly added methods are in \textcolor{red}{red}, too.}
\label{tab:main_tab}
\begin{tabular}{cp{1cm}p{1cm}|p{1cm}p{1cm}|p{1cm}p{1cm}|p{1cm}p{1cm}|p{1cm}p{1cm}|p{1cm}p{1cm}}
\hline\hline
\multicolumn{1}{c}{\multirow{3}{*}{\textbf{Method}}} & \multicolumn{6}{c}{\textbf{Practical heterogeneous ($\beta=0.1, N=20$)}} & \multicolumn{6}{c}{\textbf{Pathological heterogeneous ($N=20$)}} \\ \cline{2-13} 
 \multicolumn{1}{l}{}    &  \multicolumn{2}{c}{\textbf{CIFAR-10}} & \multicolumn{2}{c}{\textbf{EMNIST}} & \multicolumn{2}{c}{\textbf{CIFAR-100}} &  \multicolumn{2}{c}{\textbf{CIFAR-10}} & \multicolumn{2}{c}{\textbf{EMNIST}} & \multicolumn{2}{c}{\textbf{CIFAR-100}} \\  \cline{2-13} 
       \multicolumn{1}{l}{}         &  \textbf{Acc. (\%)} & \textbf{Std. (\%)}  &  \textbf{Acc. (\%)}  & \textbf{Std. (\%)}     & \textbf{Acc. (\%)} & \textbf{Std. (\%)}   & \textbf{Acc. (\%)}  & \textbf{Std. (\%)}       & \textbf{Acc. (\%)}    & \textbf{Std. (\%)}     & \textbf{Acc. (\%)} & \textbf{Std. (\%)}   \\ \hline
       
Local           & 69.90\textcolor{red}{$\downarrow$}  & 9.44\textcolor{red}{$\downarrow$} & 91.80\textcolor{red}{$\downarrow$} & 4.50\textcolor{red}{$\downarrow$} & 35.12\textcolor{red}{$\downarrow$} & 6.25\textcolor{red}{$\downarrow$} & 61.94\textcolor{red}{$\downarrow$} & 11.56\textcolor{red}{$\downarrow$} & 81.21\textcolor{red}{$\downarrow$} & 7.25\textcolor{red}{$\downarrow$} & 30.66\textcolor{red}{$\downarrow$} & 6.33\textcolor{red}{$\downarrow$}   \\\hline

FedAvg          & 61.51\textcolor{red}{$\downarrow$}  & 11.23\textcolor{red}{$\downarrow$} & 83.59\textcolor{red}{$\downarrow$} & 10.98\textcolor{red}{$\downarrow$} &  30.88\textcolor{red}{$\downarrow$} & \underline{3.46}$^*$\textcolor{darkgreen}{$\uparrow$} & 51.72\textcolor{red}{$\downarrow$} & 17.72\textcolor{red}{$\downarrow$} & 74.19\textcolor{red}{$\downarrow$} & 13.36\textcolor{red}{$\downarrow$} & 25.84\textcolor{red}{$\downarrow$} & 6.32\textcolor{red}{$\downarrow$}   \\

FedProx         & 62.91\textcolor{red}{$\downarrow$} & 9.75\textcolor{red}{$\downarrow$} & 85.33\textcolor{red}{$\downarrow$}  & 7.54\textcolor{red}{$\downarrow$} & 32.45\textcolor{red}{$\downarrow$} & \underline{3.69}$^{\dagger}$\textcolor{darkgreen}{$\uparrow$} & 62.09\textcolor{red}{$\downarrow$} & 7.01\textcolor{darkgreen}{$\uparrow$} & 84.32\textcolor{red}{$\downarrow$} & 6.09\textcolor{red}{$\downarrow$} & 31.23\textcolor{red}{$\downarrow$} & \underline{4.28}$^{\ddagger}$\textcolor{darkgreen}{$\uparrow$}  \\

MOON            & 82.96\textcolor{red}{$\downarrow$} & 10.19\textcolor{red}{$\downarrow$} & 93.61\textcolor{red}{$\downarrow$}& 4.77\textcolor{red}{$\downarrow$} & 48.04\textcolor{red}{$\downarrow$} & 5.11\textcolor{darkgreen}{$\uparrow$} & 86.88\textcolor{darkgreen}{$\uparrow$} & 13.49\textcolor{red}{$\downarrow$} & 95.24\textcolor{darkgreen}{$\uparrow$} & 7.40\textcolor{red}{$\downarrow$} & 51.31\textcolor{darkgreen}{$\uparrow$} & 5.91\textcolor{red}{$\downarrow$}  \\ \hline

FedRCL        & 87.26\textcolor{darkgreen}{$\uparrow$} & 8.54\textcolor{red}{$\downarrow$} & 93.77\textcolor{red}{$\downarrow$} & 3.22\textcolor{red}{$\downarrow$} & 50.55\textcolor{darkgreen}{$\uparrow$} & 7.89\textcolor{red}{$\downarrow$} & 88.56\textcolor{darkgreen}{$\uparrow$} & 7.83\textcolor{darkgreen}{$\uparrow$} & 93.16\textcolor{red}{$\downarrow$} & 3.58\textcolor{darkgreen}{$\uparrow$} & 49.31\textcolor{darkgreen}{$\uparrow$} & 7.14\textcolor{red}{$\downarrow$} \\ \hline

pFedMe      &    84.75   \textcolor{red}{$\downarrow$}   & 9.55\textcolor{red}{$\downarrow$} &  95.55\textcolor{darkgreen}{$\uparrow$}  & 1.47\textcolor{darkgreen}{$\uparrow$} &   46.20   \textcolor{red}{$\downarrow$}  & 6.89\textcolor{red}{$\downarrow$} &    85.72\textcolor{darkgreen}{$\uparrow$}  &  9.63\textcolor{red}{$\downarrow$} &  95.78\textcolor{darkgreen}{$\uparrow$}  & 3.03\textcolor{darkgreen}{$\uparrow$} & 46.88   \textcolor{red}{$\downarrow$} & 4.76\textcolor{darkgreen}{$\uparrow$}  \\

Ditto     &    87.53\textcolor{darkgreen}{$\uparrow$}    & 8.92\textcolor{red}{$\downarrow$} &   96.53   \textcolor{darkgreen}{$\uparrow$}   & \underline{1.17}$^{\ddagger}$\textcolor{darkgreen}{$\uparrow$} &  46.40\textcolor{red}{$\downarrow$}  & 4.81\textcolor{darkgreen}{$\uparrow$} &  89.08   \textcolor{darkgreen}{$\uparrow$}   & 7.37\textcolor{darkgreen}{$\uparrow$} &  96.48\textcolor{darkgreen}{$\uparrow$}  & 2.24\textcolor{darkgreen}{$\uparrow$} &  48.81\textcolor{darkgreen}{$\uparrow$} & 4.51\textcolor{darkgreen}{$\uparrow$}   \\\hline

PerFedAvg       & 88.95\textcolor{darkgreen}{$\uparrow$} & 6.98\textcolor{darkgreen}{$\uparrow$} & 94.63\textcolor{darkgreen}{$\uparrow$} & 1.72\textcolor{darkgreen}{$\uparrow$} & 48.80\textcolor{darkgreen}{$\uparrow$} & 5.32\textcolor{darkgreen}{$\uparrow$} & 89.45\textcolor{darkgreen}{$\uparrow$}  & 7.69\textcolor{darkgreen}{$\uparrow$} & 93.95\textcolor{darkgreen}{$\uparrow$} & 2.47\textcolor{darkgreen}{$\uparrow$} & 48.03\textcolor{darkgreen}{$\uparrow$} & \underline{4.23}$^{\dagger}$\textcolor{darkgreen}{$\uparrow$}  \\

PeFLL        & 90.53\textcolor{darkgreen}{$\uparrow$} & 7.21\textcolor{darkgreen}{$\uparrow$} & 96.45\textcolor{darkgreen}{$\uparrow$} & 2.96\textcolor{red}{$\downarrow$} & \underline{53.89}$^{\dagger}$\textcolor{darkgreen}{$\uparrow$} & 7.39\textcolor{red}{$\downarrow$} & 90.94\textcolor{darkgreen}{$\uparrow$} & 8.27\textcolor{darkgreen}{$\uparrow$} & 95.12\textcolor{darkgreen}{$\uparrow$} & 3.28\textcolor{darkgreen}{$\uparrow$} & 52.80\textcolor{darkgreen}{$\uparrow$} & 6.10\textcolor{red}{$\downarrow$} \\ 
\hline

FedRep          & \underline{90.66}$^{\ddagger}$\textcolor{darkgreen}{$\uparrow$} & \underline{6.28}$^{\dagger}$\textcolor{darkgreen}{$\uparrow$} & \underline{97.00}$^{\dagger}$\textcolor{darkgreen}{$\uparrow$} & 1.21\textcolor{darkgreen}{$\uparrow$} & 52.06\textcolor{darkgreen}{$\uparrow$} & 5.15\textcolor{darkgreen}{$\uparrow$} & 91.47\textcolor{darkgreen}{$\uparrow$} & 7.43\textcolor{darkgreen}{$\uparrow$} & \underline{96.59}$^{\dagger}$\textcolor{darkgreen}{$\uparrow$} & 2.58\textcolor{darkgreen}{$\uparrow$} & 53.12\textcolor{darkgreen}{$\uparrow$} & 4.90\textcolor{darkgreen}{$\uparrow$} \\

LG-FedAvg       & 88.72\textcolor{darkgreen}{$\uparrow$} & 8.05\textcolor{darkgreen}{$\uparrow$} & 95.90\textcolor{darkgreen}{$\uparrow$} & 1.50\textcolor{darkgreen}{$\uparrow$} & 48.19\textcolor{darkgreen}{$\uparrow$} & 6.25\textcolor{red}{$\downarrow$} & 91.35\textcolor{darkgreen}{$\uparrow$} & 7.15\textcolor{darkgreen}{$\uparrow$} & 94.21\textcolor{darkgreen}{$\uparrow$} & \underline{1.95}$^*$\textcolor{darkgreen}{$\uparrow$} & 46.67\textcolor{red}{$\downarrow$} & 5.84\textcolor{red}{$\downarrow$}  \\

FedPer       & 89.94\textcolor{darkgreen}{$\uparrow$} & \underline{6.51}$^{\ddagger}$\textcolor{darkgreen}{$\uparrow$} & 96.18\textcolor{darkgreen}{$\uparrow$} & 1.61\textcolor{darkgreen}{$\uparrow$} & 53.42\textcolor{darkgreen}{$\uparrow$} & 5.23\textcolor{darkgreen}{$\uparrow$} & 91.23\textcolor{darkgreen}{$\uparrow$} & 6.83\textcolor{darkgreen}{$\uparrow$} & 94.82\textcolor{darkgreen}{$\uparrow$} & 2.56\textcolor{darkgreen}{$\uparrow$} & 53.34\textcolor{darkgreen}{$\uparrow$} & 5.25\textcolor{darkgreen}{$\uparrow$} \\
 \hline

FedAMP           & 89.27\textcolor{darkgreen}{$\uparrow$} & 7.38\textcolor{darkgreen}{$\uparrow$} & 96.32\textcolor{darkgreen}{$\uparrow$} & 1.30\textcolor{darkgreen}{$\uparrow$} & 51.62\textcolor{darkgreen}{$\uparrow$} & 5.88\textcolor{red}{$\downarrow$} & 91.42\textcolor{darkgreen}{$\uparrow$} & \underline{6.78}$^{\ddagger}$\textcolor{darkgreen}{$\uparrow$} & 95.99\textcolor{darkgreen}{$\uparrow$} & 2.42\textcolor{darkgreen}{$\uparrow$} & 51.27\textcolor{darkgreen}{$\uparrow$} & 5.83\textcolor{red}{$\downarrow$}  \\

FedALA            & 83.33\textcolor{red}{$\downarrow$} & 9.81\textcolor{red}{$\downarrow$} & 92.37\textcolor{red}{$\downarrow$}& 3.21\textcolor{red}{$\downarrow$} & 40.41\textcolor{red}{$\downarrow$} & 3.98\textcolor{darkgreen}{$\uparrow$} & 83.88\textcolor{red}{$\downarrow$} & 13.47\textcolor{red}{$\downarrow$} & 92.24\textcolor{red}{$\downarrow$} & 3.87\textcolor{red}{$\downarrow$} & 45.31\textcolor{red}{$\downarrow$} & 5.20\textcolor{darkgreen}{$\uparrow$}  \\ 

FedAS        & 90.15\textcolor{darkgreen}{$\uparrow$} & 6.77\textcolor{darkgreen}{$\uparrow$} & 96.77\textcolor{darkgreen}{$\uparrow$} & 1.41\textcolor{darkgreen}{$\uparrow$} & 52.85\textcolor{darkgreen}{$\uparrow$} & 4.84\textcolor{darkgreen}{$\uparrow$} & 90.66\textcolor{darkgreen}{$\uparrow$} & 6.88\textcolor{darkgreen}{$\uparrow$} & 94.84\textcolor{darkgreen}{$\uparrow$} & 2.92\textcolor{darkgreen}{$\uparrow$} & 52.13\textcolor{darkgreen}{$\uparrow$} & 5.05\textcolor{darkgreen}{$\uparrow$} \\

\hline

FedProto        & 89.82\textcolor{darkgreen}{$\uparrow$} & 7.18\textcolor{darkgreen}{$\uparrow$} & \underline{96.82}$^{\ddagger}$\textcolor{darkgreen}{$\uparrow$} & 1.21\textcolor{darkgreen}{$\uparrow$} & \underline{53.47}$^{\ddagger}$\textcolor{darkgreen}{$\uparrow$} & 6.00\textcolor{red}{$\downarrow$} & \underline{91.58}$^{\dagger}$\textcolor{darkgreen}{$\uparrow$} & \underline{6.52}$^{\dagger}$\textcolor{darkgreen}{$\uparrow$} & \underline{96.49}$^{\ddagger}$\textcolor{darkgreen}{$\uparrow$} & 2.25\textcolor{darkgreen}{$\uparrow$} & \underline{53.52}$^{\ddagger}$\textcolor{darkgreen}{$\uparrow$} & 5.09\textcolor{darkgreen}{$\uparrow$}  \\

FedPAC          & \underline{90.79}$^{\dagger}$\textcolor{darkgreen}{$\uparrow$} & 6.72\textcolor{darkgreen}{$\uparrow$} & 96.66\textcolor{darkgreen}{$\uparrow$}& \underline{1.02}$^*$\textcolor{darkgreen}{$\uparrow$} & 53.14\textcolor{darkgreen}{$\uparrow$}  & 4.27\textcolor{darkgreen}{$\uparrow$} & \underline{91.55}$^{\ddagger}$\textcolor{darkgreen}{$\uparrow$} & 7.52\textcolor{darkgreen}{$\uparrow$} & 95.95\textcolor{darkgreen}{$\uparrow$} & \underline{2.15}$^{\ddagger}$\textcolor{darkgreen}{$\uparrow$} & \underline{53.59}$^{\dagger}$\textcolor{darkgreen}{$\uparrow$} & 4.44\textcolor{darkgreen}{$\uparrow$}   \\ 

FedGH        & 88.95\textcolor{darkgreen}{$\uparrow$} & 7.54\textcolor{darkgreen}{$\uparrow$} & 96.01\textcolor{darkgreen}{$\uparrow$} & 1.46\textcolor{darkgreen}{$\uparrow$} & 46.85\textcolor{red}{$\downarrow$} & 6.27\textcolor{red}{$\downarrow$} & 91.39\textcolor{darkgreen}{$\uparrow$} & 7.04\textcolor{darkgreen}{$\uparrow$} & 95.66\textcolor{darkgreen}{$\uparrow$} & 2.58\textcolor{darkgreen}{$\uparrow$} & 48.65\textcolor{darkgreen}{$\uparrow$} & 5.44\textcolor{red}{$\downarrow$} \\

pFedFDA        & 89.25\textcolor{darkgreen}{$\uparrow$} & 9.01\textcolor{red}{$\downarrow$} & 95.47\textcolor{darkgreen}{$\uparrow$} & 4.64\textcolor{red}{$\downarrow$} & 52.30\textcolor{darkgreen}{$\uparrow$} & 8.59\textcolor{red}{$\downarrow$} & 89.56\textcolor{darkgreen}{$\uparrow$} & 8.55\textcolor{darkgreen}{$\uparrow$} & 94.80\textcolor{darkgreen}{$\uparrow$} & 4.99\textcolor{red}{$\downarrow$} & 51.85\textcolor{darkgreen}{$\uparrow$} & 7.10\textcolor{red}{$\downarrow$} \\

FedDBL        & 89.91\textcolor{darkgreen}{$\uparrow$} & 6.55\textcolor{darkgreen}{$\uparrow$} & 96.28\textcolor{darkgreen}{$\uparrow$} & 1.57\textcolor{darkgreen}{$\uparrow$} & 52.56\textcolor{darkgreen}{$\uparrow$} & 4.71\textcolor{darkgreen}{$\uparrow$} & 90.38\textcolor{darkgreen}{$\uparrow$} & 6.80\textcolor{darkgreen}{$\uparrow$} & 95.50\textcolor{darkgreen}{$\uparrow$} & 2.36\textcolor{darkgreen}{$\uparrow$} & 50.44\textcolor{darkgreen}{$\uparrow$} & 4.73\textcolor{darkgreen}{$\uparrow$} \\

FedStream        & 90.51\textcolor{darkgreen}{$\uparrow$} & 8.35\textcolor{red}{$\downarrow$} & 96.79\textcolor{darkgreen}{$\uparrow$} & 2.12\textcolor{darkgreen}{$\uparrow$} & 52.99\textcolor{darkgreen}{$\uparrow$} & 5.85\textcolor{red}{$\downarrow$} & 90.69\textcolor{darkgreen}{$\uparrow$} & 7.20\textcolor{darkgreen}{$\uparrow$} & 96.11\textcolor{darkgreen}{$\uparrow$} & 2.57\textcolor{darkgreen}{$\uparrow$} & 53.00\textcolor{darkgreen}{$\uparrow$} & 4.88\textcolor{darkgreen}{$\uparrow$} \\

\hline \hline

\textbf{FedCoSR}         & \textbf{92.15}$^*$\textcolor{darkgreen}{$\uparrow$} & \textbf{5.57}$^*$\textcolor{darkgreen}{$\uparrow$} & \textbf{97.98}$^*$\textcolor{darkgreen}{$\uparrow$} & \textbf{1.15}$^{\dagger}$\textcolor{darkgreen}{$\uparrow$} & \textbf{57.01}$^*$\textcolor{darkgreen}{$\uparrow$} & \textbf{3.76}$^{\ddagger}$\textcolor{darkgreen}{$\uparrow$} & \textbf{93.31}$^*$\textcolor{darkgreen}{$\uparrow$} & \textbf{6.05}$^*$\textcolor{darkgreen}{$\uparrow$} & \textbf{97.85}$^*$\textcolor{darkgreen}{$\uparrow$} & \textbf{2.01}$^{\dagger}$\textcolor{darkgreen}{$\uparrow$} & \textbf{57.60}$^*$\textcolor{darkgreen}{$\uparrow$} & \textbf{4.07}$^*$\textcolor{darkgreen}{$\uparrow$} \\ \hline

\textbf{Averaged}     & 85.51 & 8.07 & 94.59 & 2.80 & 48.17 & 5.52 & 85.74 & 8.56 & 93.17  & 3.81 & 47.95  & 5.33  \\
\hline\hline
\end{tabular}
\end{table*}

As shown in Table~\ref{tab:main_tab}, FedCoSR achieves the highest test accuracy on all three datasets in both practical and pathological settings, confirming its effectiveness.
Notably, FedCoSR's advantage grows with the number of label classes. In Table~\ref{tab:main_tab}, its margin over the runner-up, i.e., FedProto, widens in heterogeneous settings: CIFAR-100 (3.54\%/4.01\%) and CIFAR-10 (1.36\%/1.73\%).
This aligns with~\cite{chen2020simple}, which states that with fully reliable labels, the InfoNCE lower bound on mutual information for positive pairs tightens as the number of negative samples increases, often boosting performance.
Hence, contrastive learning on cross-client representation centroids is particularly effective for heterogeneity when label classes are relatively numerous.

\begin{figure}[t]
    \centering   
    \subfigure{
    \includegraphics[width=.5\textwidth]{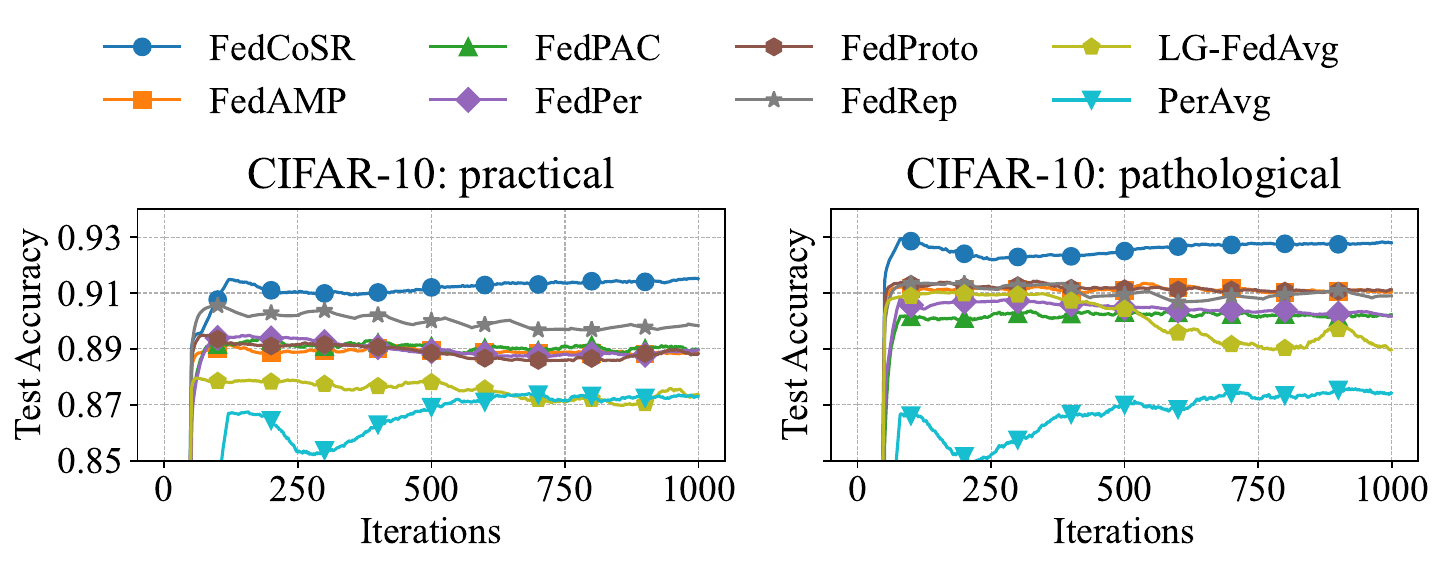}  }
    \subfigure{
    \includegraphics[width=.5\textwidth]{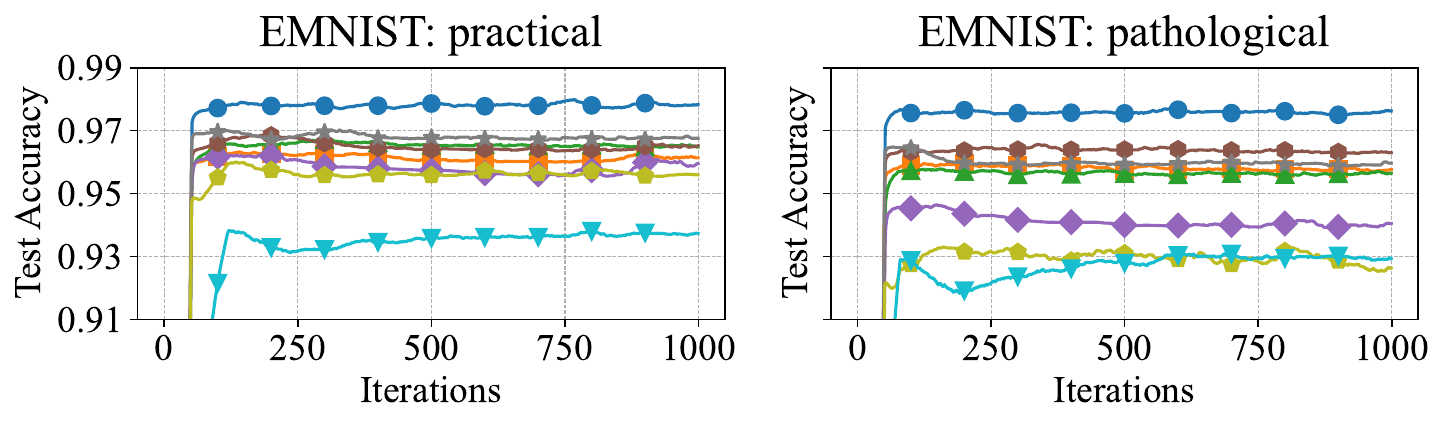} }
    \subfigure{
    \includegraphics[width=.5\textwidth]{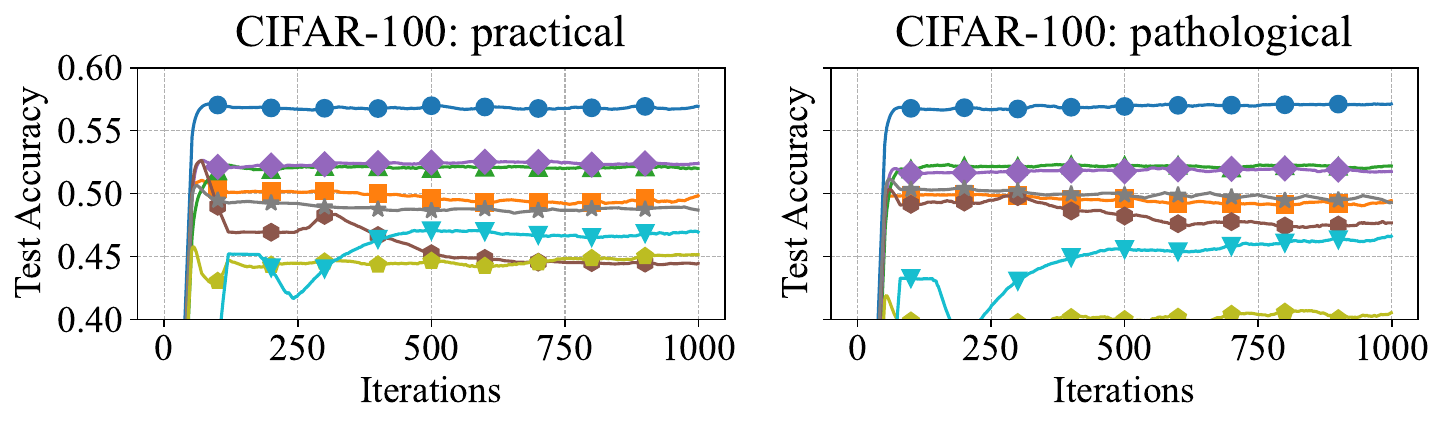}  }
    \captionsetup{width=.99\linewidth}
    \caption{The smoothed learning curves of well-performing methods in the default settings.}
    \vspace{-5pt}
    \label{Fig:curves}
\end{figure}

We evaluate learning efficiency using training curves of FedCoSR and other top-performing methods in Fig.~\ref{Fig:curves}, with each curve smoothed by a moving average of length 50.
FedCoSR attains the highest accuracy convergence in both practical and pathological settings, benefiting from CRL-enhanced local personalization.
While convergence speeds are comparable, FedCoSR shows more stable accuracy and loss trends, aided by stability from local adaptive aggregation.

\subsubsection{Label Distribution Skew-Robustness to Varying Heterogeneity}\label{sec: robustness skew}
To evaluate FedCoSR's ability to handle different heterogeneity levels, we vary the Dirichlet parameter $\beta$ on CIFAR-10 for practical heterogeneity and adjust per-client label classes on CIFAR-100 for pathological heterogeneity, as shown in Table~\ref{tab:scala_and_hete}.
Most PFL methods perform better in more heterogeneous settings, with FedCoSR consistently ranking in the top three.
Interestingly, accuracy generally drops but stability improves when heterogeneity becomes moderate, likely due to a smaller performance gap between data-rich and data-scarce clients.
Notably, FedPAC fails on practical CIFAR-10 with $\beta=0.01$ because extreme label scarcity can make its optimization problem infeasible.
In contrast, FedCoSR remains robust under both extreme and moderate label heterogeneity.

\subsubsection{Data Scarcity-Fairness Maintenance}\label{sec: fairness}

\begin{figure}[t]
    \centering
    \includegraphics[width=0.48\textwidth]{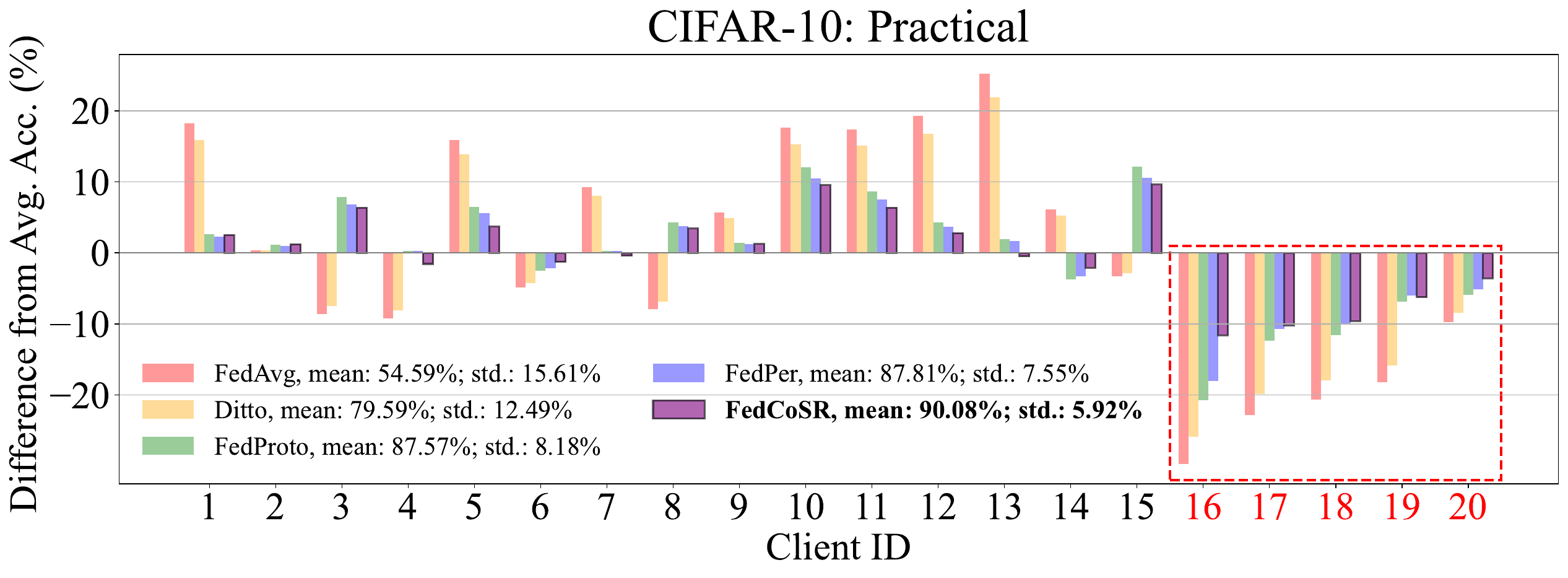}
    \caption{Performance of each client on CIFAR-10 with practical heterogeneity using different FL methods. The client number in red means its dataset size is much smaller than the others.}
    \vspace{-10pt}
    \label{Fig:fairness}
\end{figure}

In Fig.~\ref{Fig:fairness}, we assess FedAvg and fairness-focused algorithms under an FL setting where the last five clients (16-20) have only 10\% of their original data.
The aim is to evaluate fairness in model performance for severely data-scarce clients.
FedCoSR achieves the best results, with a mean accuracy of 90.08\% and a standard deviation of 5.92\%, owing to its ability to enhance shared representations via contrastive learning and leverage local adaptive aggregation, compensating for reduced data and improving personalization.
FedPer and FedProto also perform well by mitigating heterogeneity through model splitting and personalized aggregation, but still trail FedCoSR.
Ditto performs moderately, indicating its regularization strategy is less effective in data-scarce settings, while FedAvg performs worst, reflecting the limitations of pure model aggregation without personalization.

As shown in Tables~\ref{tab:main_tab} and \ref{tab:scala_and_hete}, FedCoSR consistently ranks in the top three for fairness, with smaller inter-group performance gaps.
While generalization-focused FL methods like FedProx and Ditto often score high on fairness, and PFL methods like FedRep and PerFedAvg achieve good fairness by balancing global and local updates, FedCoSR strikes a better balance between generalization and personalization, delivering both high accuracy and fairness.
Overall, FedCoSR offers the most balanced client performance through privacy-preserving knowledge sharing and local adaptation, proving effective for clients with scarce data and diverse labels.

\begin{table*}[!t]
\renewcommand{\arraystretch}{1.1}
\centering
\setlength{\tabcolsep}{2.5pt}
\scriptsize
\caption{The accuracy of changing $N$ of CIFAR-10 for scalability evaluation, $\beta$ of CIFAR-10 for practical heterogeneity evaluation, and label class number $C$ of each client of CIFAR-100 for pathological heterogeneity evaluation. We use superscripts $*$, $\dagger$, and $\ddagger$, to emphasize the 1st, 2nd, and 3rd best values in each column, respectively. \textcolor{purple}{Purple $+$} means improvement on previous settings, e.g., $N=100$ v.s. $N=20$, and $\beta=0.01$ v.s. $\beta=0.1$ in Table~\ref{tab:main_tab}, while \textcolor{blue}{blue $-$} means degradation. \textcolor{darkgreen}{Green $\uparrow$} means better and \textcolor{red}{red $\downarrow$} means worse than the averaged result. The data distributions of these settings are given in Appendix E of \cite{huang2024fedcosrapp}.}
\label{tab:scala_and_hete}
\begin{tabular}{cp{1.2cm}p{1.2cm}|p{1.2cm}p{1.2cm}|p{1.2cm}p{1.2cm}|p{1.2cm}p{1.2cm}|p{1.2cm}p{1.2cm}|p{1.2cm}p{1.2cm}}
\hline\hline
\multicolumn{1}{c}{\multirow{3}{*}{\textbf{Method}}} & \multicolumn{4}{c}{\textbf{Scalability (CIFAR-10)}} &  \multicolumn{4}{c}{\textbf{Heterogeneity (CIFAR-10 prac.)}} & \multicolumn{4}{c}{\textbf{Heterogeneity (CIFAR-100 path.)}} \\ \cline{2-13}
       \multicolumn{1}{l}{}         & \multicolumn{2}{c}{\textbf{Prac.} $N=100$}        & \multicolumn{2}{c}{\textbf{Path.} $N=100$}        & \multicolumn{2}{c}{$\beta=0.01$}  & 
 \multicolumn{2}{c}{$\beta=1$}     & \multicolumn{2}{c}{\textbf{$C$/Client}$=10$}   &   \multicolumn{2}{c}{\textbf{$C$/Client}$=50$}       \\ \cline{2-13}
 \multicolumn{1}{l}{} &  \textbf{Acc. (\%)} & \textbf{Std. (\%)} &  \textbf{Acc. (\%)} & \textbf{Std. (\%)} &  \textbf{Acc. (\%)} & \textbf{Std. (\%)} &  \textbf{Acc. (\%)} & \textbf{Std. (\%)} &  \textbf{Acc. (\%)} & \textbf{Std. (\%)} &  \textbf{Acc. (\%)} & \textbf{Std. (\%)} \\
 \hline
 
Local  & 65.72\textcolor{blue}{-}\textcolor{red}{$\downarrow$}  & 18.47\textcolor{blue}{-}\textcolor{red}{$\downarrow$} 
& 67.44\textcolor{purple}{+}\textcolor{red}{$\downarrow$} & 15.90\textcolor{blue}{-}\textcolor{red}{$\downarrow$}
& 44.25\textcolor{blue}{-}\textcolor{red}{$\downarrow$} & 20.77\textcolor{blue}{-}\textcolor{red}{$\downarrow$}
& \underline{77.57}$^*$\textcolor{purple}{+}\textcolor{darkgreen}{$\uparrow$} & 6.82\textcolor{purple}{+}\textcolor{red}{$\downarrow$}
& 30.13\textcolor{blue}{-}\textcolor{red}{$\downarrow$} & 7.12\textcolor{blue}{-}\textcolor{red}{$\downarrow$}
& 34.89\textcolor{purple}{+}\textcolor{darkgreen}{$\uparrow$} & 6.80\textcolor{blue}{-}\textcolor{red}{$\downarrow$} \\ \hline

FedAvg          & 57.86\textcolor{blue}{-}\textcolor{red}{$\downarrow$}  & \underline{10.34}$^{\ddagger}$\textcolor{purple}{+}\textcolor{darkgreen}{$\uparrow$} 
& 59.80\textcolor{purple}{+}\textcolor{red}{$\downarrow$} & 13.80\textcolor{blue}{-}\textcolor{red}{$\downarrow$}
& 30.45\textcolor{blue}{-}\textcolor{red}{$\downarrow$} & 22.34\textcolor{blue}{-}\textcolor{red}{$\downarrow$}
&  71.23\textcolor{purple}{+}\textcolor{darkgreen}{$\uparrow$}    & 4.26\textcolor{purple}{+}\textcolor{darkgreen}{$\uparrow$}
&  20.94\textcolor{blue}{-}\textcolor{red}{$\downarrow$} & 6.24\textcolor{blue}{-}\textcolor{red}{$\downarrow$}
& 31.19\textcolor{purple}{+}\textcolor{red}{$\downarrow$} & 2.59\textcolor{purple}{+}\textcolor{darkgreen}{$\uparrow$} \\

FedProx         & 60.82\textcolor{blue}{-}\textcolor{red}{$\downarrow$}  & \underline{8.81}$^*$\textcolor{blue}{-}\textcolor{darkgreen}{$\uparrow$} 
& 65.26\textcolor{blue}{-}\textcolor{red}{$\downarrow$} & \underline{7.69}$^{\ddagger}$\textcolor{blue}{-}\textcolor{darkgreen}{$\uparrow$} 
& 46.13\textcolor{blue}{-}\textcolor{red}{$\downarrow$} & 13.93\textcolor{blue}{-}\textcolor{red}{$\downarrow$}
&  70.31\textcolor{purple}{+}\textcolor{darkgreen}{$\uparrow$} & \underline{4.11}$^{\ddagger}$\textcolor{purple}{+}\textcolor{darkgreen}{$\uparrow$}
& 28.84\textcolor{blue}{-}\textcolor{red}{$\downarrow$} & \underline{4.08}$^{\ddagger}$\textcolor{blue}{-}\textcolor{darkgreen}{$\uparrow$}
& 33.37\textcolor{purple}{+}\textcolor{darkgreen}{$\uparrow$} & \underline{1.85}$^*$\textcolor{purple}{+}\textcolor{darkgreen}{$\uparrow$} \\

MOON            & 80.05\textcolor{blue}{-}\textcolor{darkgreen}{$\uparrow$}  & 16.80\textcolor{blue}{-}\textcolor{red}{$\downarrow$}
& 74.15\textcolor{blue}{-}\textcolor{red}{$\downarrow$} & 15.33\textcolor{blue}{-}\textcolor{red}{$\downarrow$} 
& 95.92\textcolor{purple}{+}\textcolor{red}{$\downarrow$} & 12.55\textcolor{blue}{-}\textcolor{darkgreen}{$\uparrow$}
&  66.32\textcolor{blue}{-}\textcolor{red}{$\downarrow$}   & 4.97\textcolor{blue}{-}\textcolor{darkgreen}{$\uparrow$}
&  57.62\textcolor{purple}{+}\textcolor{red}{$\downarrow$} & 4.99\textcolor{purple}{+}\textcolor{red}{$\downarrow$}
&  23.47\textcolor{blue}{-}\textcolor{red}{$\downarrow$} & 4.58\textcolor{purple}{+}\textcolor{red}{$\downarrow$} \\ \hline

FedRCL        & 83.62\textcolor{blue}{-}\textcolor{darkgreen}{$\uparrow$} & 12.54\textcolor{blue}{-}\textcolor{darkgreen}{$\uparrow$} & 78.39\textcolor{blue}{-}\textcolor{darkgreen}{$\uparrow$} & 9.74\textcolor{blue}{-}\textcolor{darkgreen}{$\uparrow$} & 95.14\textcolor{purple}{+}\textcolor{darkgreen}{$\uparrow$} & 11.84\textcolor{blue}{-}\textcolor{darkgreen}{$\uparrow$} & 70.78\textcolor{blue}{-}\textcolor{darkgreen}{$\uparrow$} & 5.21\textcolor{purple}{+}\textcolor{red}{$\downarrow$} & 62.85\textcolor{purple}{+}\textcolor{darkgreen}{$\uparrow$} & 4.69\textcolor{purple}{+}\textcolor{darkgreen}{$\uparrow$} & 33.73\textcolor{blue}{-}\textcolor{darkgreen}{$\uparrow$} & 4.54\textcolor{purple}{+}\textcolor{red}{$\downarrow$} \\ \hline

pFedMe          & 80.22\textcolor{blue}{-}\textcolor{darkgreen}{$\uparrow$}  & 13.99\textcolor{blue}{-}\textcolor{darkgreen}{$\uparrow$} 
& 76.09\textcolor{blue}{-}\textcolor{red}{$\downarrow$} & 8.65\textcolor{blue}{-}\textcolor{darkgreen}{$\uparrow$} 
& 98.82\textcolor{purple}{+}\textcolor{darkgreen}{$\uparrow$} & \underline{8.98}$^{\ddagger}$\textcolor{purple}{+}\textcolor{darkgreen}{$\uparrow$}
&  69.01\textcolor{blue}{-}\textcolor{red}{$\downarrow$} & 4.39\textcolor{purple}{+}\textcolor{darkgreen}{$\uparrow$}
&  59.22\textcolor{purple}{+}\textcolor{darkgreen}{$\uparrow$} & 4.66\textcolor{purple}{+}\textcolor{darkgreen}{$\uparrow$}
& 28.15\textcolor{blue}{-}\textcolor{red}{$\downarrow$} & 2.84\textcolor{purple}{+}\textcolor{darkgreen}{$\uparrow$} \\

Ditto           & 82.68\textcolor{blue}{-}\textcolor{darkgreen}{$\uparrow$}  & 15.78\textcolor{blue}{-}\textcolor{red}{$\downarrow$} 
& 74.85\textcolor{blue}{-}\textcolor{red}{$\downarrow$} & 8.48\textcolor{purple}{+}\textcolor{darkgreen}{$\uparrow$} 
& 99.05\textcolor{purple}{+}\textcolor{darkgreen}{$\uparrow$} & \underline{8.60}$^{\dagger}$\textcolor{purple}{+}\textcolor{darkgreen}{$\uparrow$}
&  65.03\textcolor{blue}{-}\textcolor{red}{$\downarrow$} & 4.86\textcolor{purple}{+}\textcolor{darkgreen}{$\uparrow$}
& 57.21\textcolor{purple}{+}\textcolor{red}{$\downarrow$} & 4.59\textcolor{blue}{-}\textcolor{darkgreen}{$\uparrow$}
&  27.70\textcolor{blue}{-}\textcolor{red}{$\downarrow$} &  2.90\textcolor{purple}{+}\textcolor{darkgreen}{$\uparrow$}    \\ \hline

PerFedAvg       & 84.03\textcolor{blue}{-}\textcolor{darkgreen}{$\uparrow$}  & 12.95\textcolor{blue}{-}\textcolor{darkgreen}{$\uparrow$} 
& 79.30\textcolor{blue}{-}\textcolor{darkgreen}{$\uparrow$} & 8.55\textcolor{purple}{+}\textcolor{darkgreen}{$\uparrow$} 
& 99.02\textcolor{purple}{+}\textcolor{darkgreen}{$\uparrow$} & \underline{8.28}$^*$\textcolor{blue}{-}\textcolor{darkgreen}{$\uparrow$}
&  73.93\textcolor{blue}{-}\textcolor{darkgreen}{$\uparrow$}     & \underline{4.07}$^{\dagger}$\textcolor{purple}{+}\textcolor{darkgreen}{$\uparrow$}
& 63.46\textcolor{purple}{+}\textcolor{darkgreen}{$\uparrow$} & 4.93\textcolor{purple}{+}\textcolor{red}{$\downarrow$}
& \underline{36.17}$^{\dagger}$\textcolor{blue}{-}\textcolor{darkgreen}{$\uparrow$} & \underline{2.50}$^{\ddagger}$\textcolor{purple}{+}\textcolor{darkgreen}{$\uparrow$} \\

PeFLL        & 84.22\textcolor{blue}{-}\textcolor{darkgreen}{$\uparrow$} & 14.21\textcolor{blue}{-}\textcolor{darkgreen}{$\uparrow$} & 80.25\textcolor{blue}{-}\textcolor{darkgreen}{$\uparrow$} & 8.16\textcolor{purple}{+}\textcolor{darkgreen}{$\uparrow$} & 97.89\textcolor{purple}{+}\textcolor{darkgreen}{$\uparrow$} & 10.73\textcolor{blue}{-}\textcolor{darkgreen}{$\uparrow$} & 72.55\textcolor{blue}{-}\textcolor{darkgreen}{$\uparrow$} & 5.27\textcolor{purple}{+}\textcolor{red}{$\downarrow$} & 65.29\textcolor{purple}{+}\textcolor{darkgreen}{$\uparrow$} & 5.11\textcolor{purple}{+}\textcolor{red}{$\downarrow$} & 34.69\textcolor{blue}{-}\textcolor{darkgreen}{$\uparrow$} & 4.84\textcolor{purple}{+}\textcolor{red}{$\downarrow$} \\
\hline

FedRep          & \underline{86.08}$^{\dagger}$\textcolor{blue}{-}\textcolor{darkgreen}{$\uparrow$} & 13.23\textcolor{blue}{-}\textcolor{darkgreen}{$\uparrow$}
& \underline{82.31}$^{\ddagger}$\textcolor{blue}{-}\textcolor{darkgreen}{$\uparrow$} & \underline{7.44}$^{\dagger}$\textcolor{blue}{-}\textcolor{darkgreen}{$\uparrow$} 
& 99.18\textcolor{purple}{+}\textcolor{darkgreen}{$\uparrow$} & 9.06\textcolor{blue}{-}\textcolor{darkgreen}{$\uparrow$}
&  71.84\textcolor{blue}{-}\textcolor{darkgreen}{$\uparrow$}  &   4.59\textcolor{purple}{+}\textcolor{darkgreen}{$\uparrow$}
& \underline{68.85}$^{\dagger}$\textcolor{purple}{+}\textcolor{darkgreen}{$\uparrow$} & 4.16\textcolor{purple}{+}\textcolor{darkgreen}{$\uparrow$}
& 33.71\textcolor{blue}{-}\textcolor{darkgreen}{$\uparrow$} & 3.83\textcolor{purple}{+}\textcolor{red}{$\downarrow$} \\

LG-FedAvg          & 82.87\textcolor{blue}{-}\textcolor{darkgreen}{$\uparrow$} & 16.44\textcolor{blue}{-}\textcolor{red}{$\downarrow$}
& 76.72\textcolor{blue}{-}\textcolor{darkgreen}{$\uparrow$} & 8.80\textcolor{purple}{+}\textcolor{darkgreen}{$\uparrow$} 
& 99.17\textcolor{purple}{+}\textcolor{darkgreen}{$\uparrow$} & 22.72\textcolor{blue}{-}\textcolor{red}{$\downarrow$}
&  61.38\textcolor{blue}{-}\textcolor{red}{$\downarrow$}  &   6.04\textcolor{purple}{+}\textcolor{red}{$\downarrow$}
& 62.45\textcolor{purple}{+}\textcolor{darkgreen}{$\uparrow$} & 7.53\textcolor{purple}{+}\textcolor{red}{$\downarrow$}
& 24.52\textcolor{blue}{-}\textcolor{red}{$\downarrow$} & 5.17\textcolor{purple}{+}\textcolor{red}{$\downarrow$}\\

FedPer          & \underline{84.43}$^{\ddagger}$\textcolor{blue}{-}\textcolor{darkgreen}{$\uparrow$} & 13.60\textcolor{blue}{-}\textcolor{darkgreen}{$\uparrow$}
& 82.08\textcolor{blue}{-}\textcolor{darkgreen}{$\uparrow$} & 8.04\textcolor{blue}{-}\textcolor{darkgreen}{$\uparrow$} 
& 98.70\textcolor{purple}{+}\textcolor{darkgreen}{$\uparrow$} & 12.58\textcolor{blue}{-}\textcolor{darkgreen}{$\uparrow$}
&  73.00\textcolor{blue}{-}\textcolor{darkgreen}{$\uparrow$} &    4.16\textcolor{purple}{+}\textcolor{darkgreen}{$\uparrow$}
&  68.35\textcolor{purple}{+}\textcolor{darkgreen}{$\uparrow$} & 4.28\textcolor{purple}{+}\textcolor{darkgreen}{$\uparrow$}
& 35.84\textcolor{blue}{-}\textcolor{darkgreen}{$\uparrow$} & 4.04\textcolor{purple}{+}\textcolor{red}{$\downarrow$}\\ \hline

FedAMP          & 82.26\textcolor{blue}{-}\textcolor{darkgreen}{$\uparrow$} & 13.25\textcolor{blue}{-}\textcolor{darkgreen}{$\uparrow$}
& 75.03\textcolor{blue}{-}\textcolor{red}{$\downarrow$} & 13.04\textcolor{blue}{-}\textcolor{red}{$\downarrow$} 
& \underline{99.25}$^{\ddagger}$\textcolor{purple}{+}\textcolor{darkgreen}{$\uparrow$} & 15.82\textcolor{blue}{-}\textcolor{red}{$\downarrow$}
&  61.58\textcolor{blue}{-}\textcolor{red}{$\downarrow$}  & 9.65\textcolor{blue}{-}\textcolor{red}{$\downarrow$}
& 67.51\textcolor{purple}{+}\textcolor{darkgreen}{$\uparrow$}  & 4.80\textcolor{purple}{+}\textcolor{darkgreen}{$\uparrow$}
& 28.89\textcolor{blue}{-}\textcolor{red}{$\downarrow$} & 4.34\textcolor{purple}{+}\textcolor{red}{$\downarrow$} \\

FedALA          & 79.27\textcolor{blue}{-}\textcolor{red}{$\downarrow$} & 15.71\textcolor{blue}{-}\textcolor{red}{$\downarrow$}
& 70.44\textcolor{blue}{-}\textcolor{red}{$\downarrow$} & 15.28\textcolor{blue}{-}\textcolor{red}{$\downarrow$} 
& 97.09\textcolor{purple}{+}\textcolor{darkgreen}{$\uparrow$} & 10.26\textcolor{blue}{-}\textcolor{darkgreen}{$\uparrow$}
&  72.99\textcolor{blue}{-}\textcolor{darkgreen}{$\uparrow$} & 4.78\textcolor{purple}{+}\textcolor{darkgreen}{$\uparrow$}
&  30.84\textcolor{blue}{-}\textcolor{red}{$\downarrow$} & 4.67\textcolor{purple}{+}\textcolor{darkgreen}{$\uparrow$}
& 30.82\textcolor{blue}{-}\textcolor{red}{$\downarrow$} & 4.29\textcolor{purple}{+}\textcolor{red}{$\downarrow$} \\ 

FedAS        & 81.63\textcolor{blue}{-}\textcolor{darkgreen}{$\uparrow$} & 14.15\textcolor{blue}{-}\textcolor{darkgreen}{$\uparrow$} & 80.09\textcolor{blue}{-}\textcolor{darkgreen}{$\uparrow$} & 9.22\textcolor{blue}{-}\textcolor{darkgreen}{$\uparrow$} & 98.40\textcolor{purple}{+}\textcolor{darkgreen}{$\uparrow$} & 10.34\textcolor{blue}{-}\textcolor{darkgreen}{$\uparrow$} & 71.90\textcolor{blue}{-}\textcolor{darkgreen}{$\uparrow$} & 4.11\textcolor{purple}{+}\textcolor{darkgreen}{$\uparrow$} & 66.74\textcolor{purple}{+}\textcolor{darkgreen}{$\uparrow$} & 3.84\textcolor{purple}{+}\textcolor{darkgreen}{$\uparrow$} & 34.58\textcolor{blue}{-}\textcolor{darkgreen}{$\uparrow$} & 3.86\textcolor{purple}{+}\textcolor{red}{$\downarrow$} \\ \hline

FedProto        & 83.61\textcolor{blue}{-}\textcolor{darkgreen}{$\uparrow$} & 15.96\textcolor{blue}{-}\textcolor{red}{$\downarrow$}
& 78.01\textcolor{blue}{-}\textcolor{darkgreen}{$\uparrow$} & 8.20\textcolor{blue}{-}\textcolor{darkgreen}{$\uparrow$} 
& \underline{99.32}$^{\dagger}$\textcolor{purple}{+}\textcolor{darkgreen}{$\uparrow$} & 11.21\textcolor{blue}{-}\textcolor{darkgreen}{$\uparrow$}
&  62.92\textcolor{blue}{-}\textcolor{red}{$\downarrow$} & 5.17\textcolor{purple}{+}\textcolor{red}{$\downarrow$}
& \underline{68.81}$^{\ddagger}$\textcolor{purple}{+}\textcolor{darkgreen}{$\uparrow$} & 4.56\textcolor{purple}{+}\textcolor{darkgreen}{$\uparrow$}
& 31.48\textcolor{blue}{-}\textcolor{red}{$\downarrow$} & 4.64\textcolor{purple}{+}\textcolor{red}{$\downarrow$}\\

FedPAC          & 70.80\textcolor{blue}{-}\textcolor{red}{$\downarrow$} & 20.04\textcolor{blue}{-}\textcolor{red}{$\downarrow$}
& \underline{82.33}$^{\dagger}$\textcolor{blue}{-}\textcolor{darkgreen}{$\uparrow$}  & 9.59\textcolor{blue}{-}\textcolor{darkgreen}{$\uparrow$} 
& -  & -
&  \underline{74.84}$^{\ddagger}$\textcolor{blue}{-}\textcolor{darkgreen}{$\uparrow$} & \underline{3.44}$^*$\textcolor{purple}{+}\textcolor{darkgreen}{$\uparrow$}
& 63.30\textcolor{purple}{+}\textcolor{darkgreen}{$\uparrow$} & \underline{2.93}$^*$\textcolor{purple}{+}\textcolor{darkgreen}{$\uparrow$}
& \underline{36.12}$^{\ddagger}$\textcolor{blue}{-}\textcolor{darkgreen}{$\uparrow$} & 3.60\textcolor{purple}{+}\textcolor{darkgreen}{$\uparrow$} \\

FedGH          & 82.64\textcolor{blue}{-}\textcolor{darkgreen}{$\uparrow$} & 15.66\textcolor{blue}{-}\textcolor{red}{$\downarrow$}
& 76.98\textcolor{blue}{-}\textcolor{darkgreen}{$\uparrow$}  & 8.52\textcolor{blue}{-}\textcolor{darkgreen}{$\uparrow$} 
& 99.23\textcolor{purple}{+}\textcolor{darkgreen}{$\uparrow$} & 22.45\textcolor{blue}{-}\textcolor{red}{$\downarrow$}
&  61.16\textcolor{blue}{-}\textcolor{red}{$\downarrow$} & 5.25\textcolor{purple}{+}\textcolor{red}{$\downarrow$}
& 65.50\textcolor{purple}{+}\textcolor{darkgreen}{$\uparrow$} & 4.74\textcolor{blue}{-}\textcolor{darkgreen}{$\uparrow$}
& 27.55\textcolor{blue}{-}\textcolor{red}{$\downarrow$} & 3.98\textcolor{purple}{+}\textcolor{red}{$\downarrow$} \\

pFedFDA        & 84.06\textcolor{blue}{-}\textcolor{darkgreen}{$\uparrow$} & 13.88\textcolor{blue}{-}\textcolor{darkgreen}{$\uparrow$} & 81.50\textcolor{blue}{-}\textcolor{darkgreen}{$\uparrow$} & 8.10\textcolor{purple}{+}\textcolor{darkgreen}{$\uparrow$} & 98.94\textcolor{purple}{+}\textcolor{darkgreen}{$\uparrow$} & 14.71\textcolor{blue}{-}\textcolor{red}{$\downarrow$} & 73.23\textcolor{blue}{-}\textcolor{darkgreen}{$\uparrow$} & 4.46\textcolor{purple}{+}\textcolor{darkgreen}{$\uparrow$} & 67.58\textcolor{purple}{+}\textcolor{darkgreen}{$\uparrow$} & 4.25\textcolor{purple}{+}\textcolor{darkgreen}{$\uparrow$} & 34.88\textcolor{blue}{-}\textcolor{darkgreen}{$\uparrow$} & 3.10\textcolor{purple}{+}\textcolor{darkgreen}{$\uparrow$} \\

FedDBL        & 81.54\textcolor{blue}{-}\textcolor{darkgreen}{$\uparrow$} & 17.61\textcolor{blue}{-}\textcolor{red}{$\downarrow$} & 77.37\textcolor{blue}{-}\textcolor{darkgreen}{$\uparrow$} & 9.25\textcolor{blue}{-}\textcolor{darkgreen}{$\uparrow$} & 98.13\textcolor{purple}{+}\textcolor{darkgreen}{$\uparrow$} & 16.11\textcolor{blue}{-}\textcolor{red}{$\downarrow$} & 60.69\textcolor{blue}{-}\textcolor{red}{$\downarrow$} & 6.08\textcolor{purple}{+}\textcolor{red}{$\downarrow$} & 61.10\textcolor{purple}{+}\textcolor{darkgreen}{$\uparrow$} & 5.25\textcolor{blue}{-}\textcolor{red}{$\downarrow$} & 29.35\textcolor{blue}{-}\textcolor{red}{$\downarrow$} & 4.22\textcolor{purple}{+}\textcolor{red}{$\downarrow$} \\

FedStream        & 84.25\textcolor{blue}{-}\textcolor{darkgreen}{$\uparrow$} & 14.10\textcolor{blue}{-}\textcolor{darkgreen}{$\uparrow$} & 80.78\textcolor{blue}{-}\textcolor{darkgreen}{$\uparrow$} & 8.83\textcolor{blue}{-}\textcolor{darkgreen}{$\uparrow$} & 99.02\textcolor{purple}{+}\textcolor{darkgreen}{$\uparrow$} & 13.46\textcolor{blue}{-}\textcolor{darkgreen}{$\uparrow$} & 70.57\textcolor{blue}{-}\textcolor{darkgreen}{$\uparrow$} & 5.14\textcolor{purple}{+}\textcolor{red}{$\downarrow$} & 68.10\textcolor{purple}{+}\textcolor{darkgreen}{$\uparrow$} & 4.99\textcolor{blue}{-}\textcolor{red}{$\downarrow$} & 33.55\textcolor{blue}{-}\textcolor{darkgreen}{$\uparrow$} & 3.67\textcolor{purple}{+}\textcolor{darkgreen}{$\uparrow$} \\

\hline \hline

\textbf{FedCoSR}& \textbf{87.57}$^*$\textcolor{blue}{-}\textcolor{darkgreen}{$\uparrow$} & \textbf{12.26}$^{\ddagger}$\textcolor{blue}{-}\textcolor{darkgreen}{$\uparrow$}
& \textbf{84.81}$^*$\textcolor{blue}{-}\textcolor{darkgreen}{$\uparrow$} & \textbf{7.37}$^*$\textcolor{blue}{-}\textcolor{darkgreen}{$\uparrow$} 
& \textbf{99.40}$^*$\textcolor{purple}{+}\textcolor{darkgreen}{$\uparrow$} & \textbf{10.78}\textcolor{blue}{-}\textcolor{darkgreen}{$\uparrow$}
&  \textbf{76.59}$^{\dagger}$\textcolor{blue}{-}\textcolor{darkgreen}{$\uparrow$} & \textbf{4.17}\textcolor{purple}{+}\textcolor{darkgreen}{$\uparrow$}
& \textbf{72.44}$^*$\textcolor{purple}{+}\textcolor{darkgreen}{$\uparrow$} & \textbf{3.56}$^{\dagger}$\textcolor{purple}{+}\textcolor{darkgreen}{$\uparrow$}
& \textbf{42.78}$^*$\textcolor{purple}{+}\textcolor{darkgreen}{$\uparrow$} & \textbf{2.17}$^{\dagger}$\textcolor{purple}{+}\textcolor{darkgreen}{$\uparrow$} \\ \hline

\textbf{Averaged}& 79.62  & 14.54 & 76.58 & 9.91 & 90.12 & 13.69 &  69.55 & 5.12 & 58.09  & 4.82 & 32.20 & 3.82 \\
\hline\hline
\end{tabular}
\end{table*}

\subsubsection{Data Scarcity-Robustness against Model Degradation}\label{sec: robustness scarce}
\begin{figure}[t]
    \centering
\includegraphics[width=0.5\textwidth]{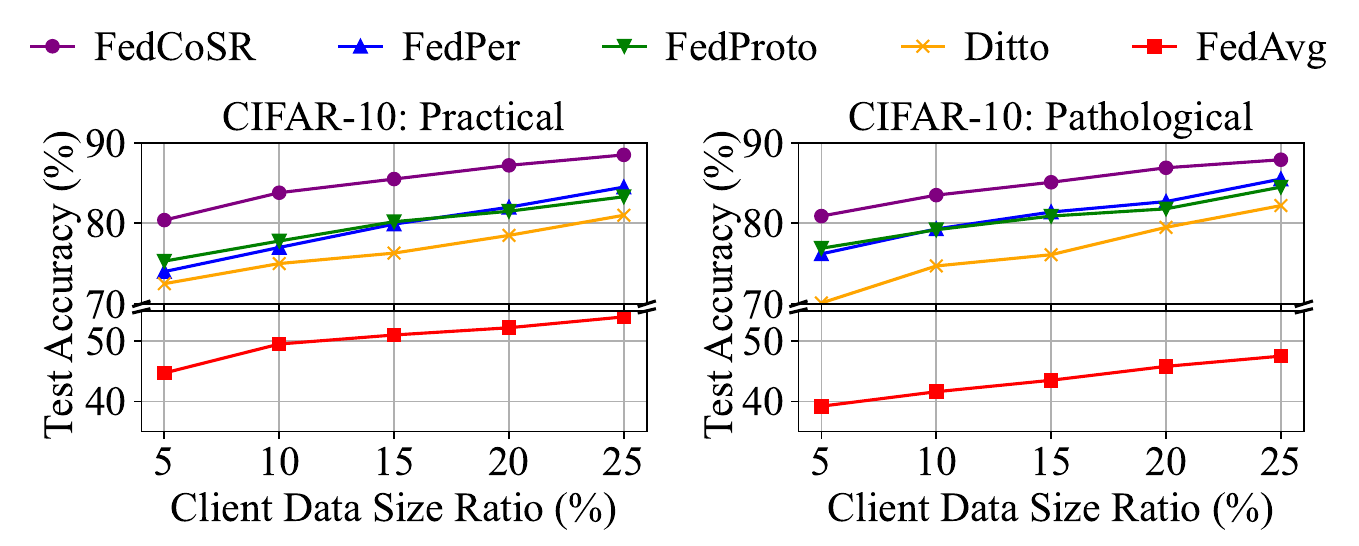}
    \includegraphics[width=0.48\textwidth]{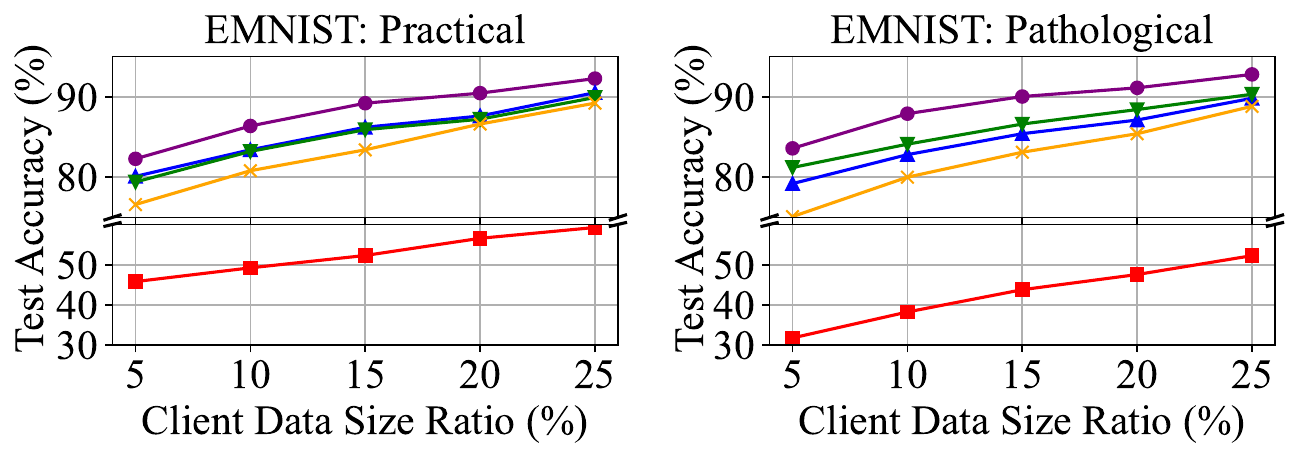}
    \includegraphics[width=0.48\textwidth]{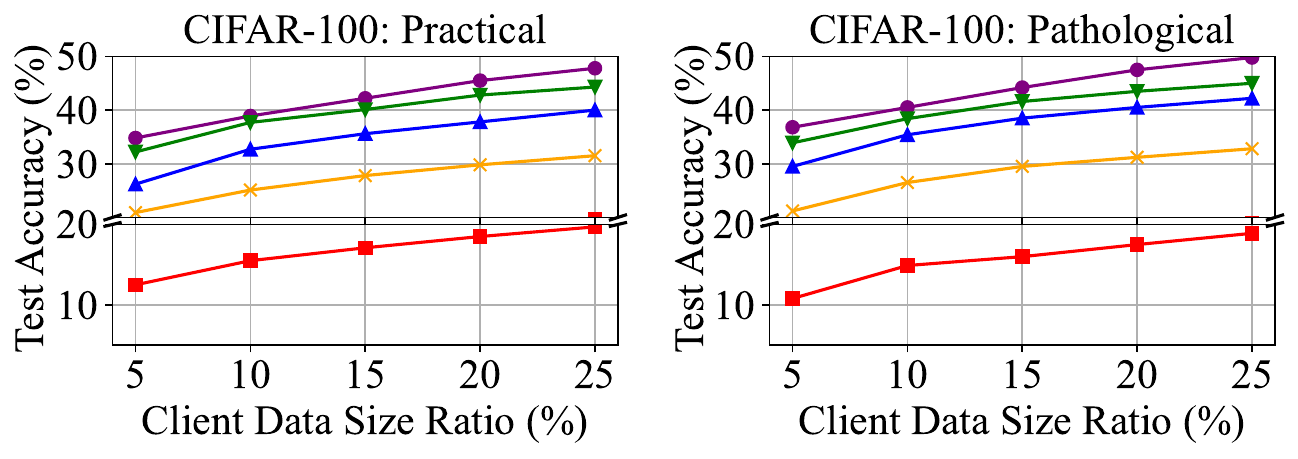}
    \caption{Robustness to the scenario where the data sizes of all clients are small.}
    \label{Fig:robust}
\end{figure}

We test our FedCoSR with varying local dataset sizes on CIFAR-10 under the two scenarios. As shown in Fig.~\ref{Fig:robust}, clients of varying local dataset sizes consistently reap the advantages of engaging in FL, while our method achieves the best performance. Compared with results in Table~\ref{tab:main_tab}, small local datasets indeed cause substantial performance degradation. Despite all methods have decreasing trends when the local dataset size becomes smaller, the degree of decline under FedCoSR is less than that of other methods, demonstrating FedCoSR's strongest robustness to data scarcity. This can be attributed to that the information embedded in the shared representations is effectively enhanced through contrastive learning among clients, compensating for the lack of data volume.

\subsection{Result Analysis and Discussion on Practicality}

\subsubsection{Scalability}
In Table~\ref{tab:scala_and_hete}, we evaluate scalability by increasing the number of clients $N$ to 100.
Compared with the $N=20$ results in Table~\ref{tab:main_tab}, most methods suffer notable drops of 6-20\%, due to more extreme label scarcity and skew in the pathological setting, which hinder personalization.
Overall, model-splitting and representation-sharing approaches show better scalability.
FedCoSR drops only 6\% in the practical setting and still ranks first in both settings, underscoring its scalability and the benefit of applying CRL to shared representations.

\subsubsection{Computation Overhead}

\begin{table}[!t]
\renewcommand{\arraystretch}{0.9}
\setlength{\tabcolsep}{8pt}
\centering
\scriptsize
\caption{The convergence speed (in number of iterations) and computational cost (in seconds) per iteration averaged on CIFAR-10, EMNIST, and CIFAR-100 under the practical and pathological heterogeneous setting. Additionally, the communication costs per iteration of each client are listed, where $\varphi(\cdot)$ is denoted as the parameters of specific modules, and $\theta$, $\phi$, $\pi$, $\boldsymbol{\Omega}$, and $C$ are denoted as the entire model, the base model, the head model, the label-centroid representations, and the class number, respectively.}
\label{tab:eff}
\begin{tabular}{cccc}
\hline\hline
\textbf{Method} & \textbf{Convergence (iter.)} & \textbf{Comp. cost (s)} & \textbf{Comm. cost} \\
\hline
Local      & 24.33 & 34.17 & - \\
FedAvg     & 33.33 & 45.17 & 2$\varphi(\theta)$ \\
FedProx    & 32.00 & 42.67 & 2$\varphi(\theta)$ \\
MOON       & 56.00 & 62.33 & 2$\varphi(\theta)$ \\
FedRCL     & 49.50 & 53.00 & 2$\varphi(\theta)$ \\
pFedMe     & 49.83 & 73.33 & 2$\varphi(\theta)$ \\
Ditto      & 48.50 & 78.50 & 2$\varphi(\theta)$ \\
PerFedAvg  & 42.33 & 46.33 & 2$\varphi(\theta)$ \\
PeFLL      & 48.67 & 74.67 & 2$\varphi(\boldsymbol{\Omega}+\theta)$ \\
FedRep     & 85.67 & 89.33 & 2$\varphi(\phi)$ \\
LG-FedAvg  & 81.00 & 47.00 & 2$\varphi(\theta)$ \\
FedPer     & 68.33 & 46.83 & 2$\varphi(\phi)$ \\
FedAMP     & 49.83 & 52.67 & 2$\varphi(\theta)$ \\
FedALA     & 71.17 & 62.17 & 2$\varphi(\theta)$ \\
FedAS      & 62.00 & 65.67 & 2$\varphi(\boldsymbol{\Omega}+\phi)$ \\
FedProto   & 77.33 & 75.67 & 2$\varphi(\boldsymbol{\Omega})$ \\
FedPAC     & 77.67 & 95.00 & 2$\varphi(\theta)$ \\
FedGH      & 57.50 & 56.50 & 2$\varphi(\boldsymbol{\Omega}+\pi)$ \\
pFedFDA    & 48.83 & 72.83 & 2$\varphi(C^2+\phi)$ \\
FedDBL    & 37.50 & 42.67 & 2$\varphi(\boldsymbol{\Omega})$ \\
FedStream    & 68.50 & 79.33 & 2$\varphi(\boldsymbol{\Omega})$ \\
\textbf{FedCoSR}    & 72.67 & 51.83 & 2$\varphi(\boldsymbol{\Omega}+\phi)$ \\
\hline
Averaged & 56.62 & 61.45 & $\thickapprox 2\varphi(\theta)$ \\
\hline\hline
\end{tabular}
\end{table}

In Table~\ref{tab:eff}, we report the iterations to convergence and per-iteration time cost, averaged over three datasets in two default heterogeneity settings.
Most algorithms converge quickly in practical settings, typically within 100 iterations, and even faster in pathological settings where extreme data distribution increases overfitting risk.
FedCoSR converges slightly slower than average in both settings but remains within an acceptable range.
Its time cost, positively correlated with the number of label classes, is slightly higher than many methods, particularly on datasets with more classes, yet performance also improves with more classes.
Overall, FedCoSR is well-suited for distributed scenarios, as computational costs are naturally spread across clients.

\subsubsection{Communication Overhead}
We evaluate the communication overhead per client in single iteration, recorded in Table~\ref{tab:eff}. We denote $\varphi(\cdot)$ as the number of parameters. Based on Eq. (\!\!~\ref{Eq9-1}) where $\theta$ is concatenated by $\phi$ and $\pi$, most methods incur the same computational overhead as FedAvg, which uploads and downloads only one entire model, denoted as 2$\varphi(\theta)$. FedProto only transmits representations, thus in general (we suppose a representation is much smaller than a model), it has the least communication overhead denoted as 2$\varphi(\bar{\boldsymbol{\Omega}})$. FedGH uploads representations and downloads projection layers, costing 2$\varphi(\bar{\boldsymbol{\Omega}}) + \varphi(\pi)$, but also takes time for global training on the server. Since FedCoSR uploads and downloads parameters of representation layers and averaged representations of each label, its overhead can be regarded as 2$[\varphi(\phi)+\varphi(\bar{\boldsymbol{\Omega}})]$, where the depth of $\phi$ is $|\theta|-1$. This communication overhead is similar to the major overhead 2$\varphi(\theta)$ and thus deemed acceptable.

\subsubsection{Privacy Concerns}
In FL, the risk of data privacy leakage is inevitable due to reverse engineering techniques, such as gradient inversion~\cite{jeon2021gradient}.
However, specific strategies can mitigate these risks without significantly affecting system performance. In our work, we adopt the following approaches to reduce privacy leakage:
\begin{itemize}
    \item \textbf{Selective model parameter sharing:} We upload only a subset of local model parameters without disclosing the model structure, thereby reducing the risk of full data reconstruction and limiting exposure to model-specific inference attacks.
    \item \textbf{Mean representations sharing:} Instead of uploading raw embeddings, we only share mean values of representations. This design ensures that shared vectors are coarse-grained and anonymized, reducing the potential for inversion attacks and limiting semantic leakage.
    \item \textbf{Minimum number of samples participating:} To further mitigate risks in sensitive cases, e.g., when the number of labels is small or certain labels appear only in one or two clients, we can set that only label centroids with support from a minimum number of clients can participate in the FL process. This avoids unintended exposure from rare or unique classes.
\end{itemize} 
Additionally, Differential Privacy (DP) can be employed on both model parameters and representations to further reduce privacy risks~\cite{wei2020federated}. 
DP adds noise to the shared data, but this comes with a trade-off between privacy and model utility, as too much noise may degrade performance.

\subsection{Result Analysis and Discussion on Methodology}

\subsubsection{Ablation Study}

\begin{figure}[!th]
\centering
\subfigure[Batch size $b$]{\label{Fig:ablation b}
    \includegraphics[width=.22\textwidth]{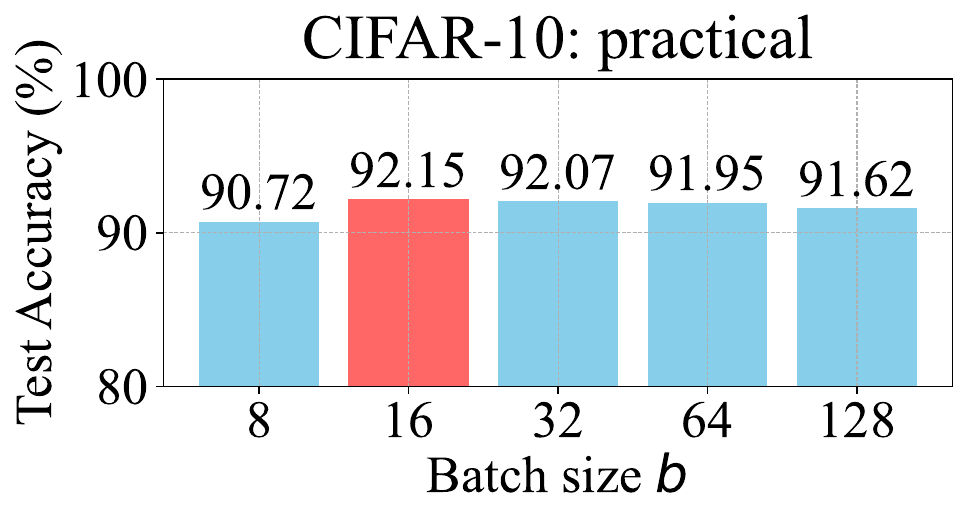}  }
\subfigure[Loss factor $\alpha$]{\label{Fig:ablation alpha}
    \includegraphics[width=.22\textwidth]{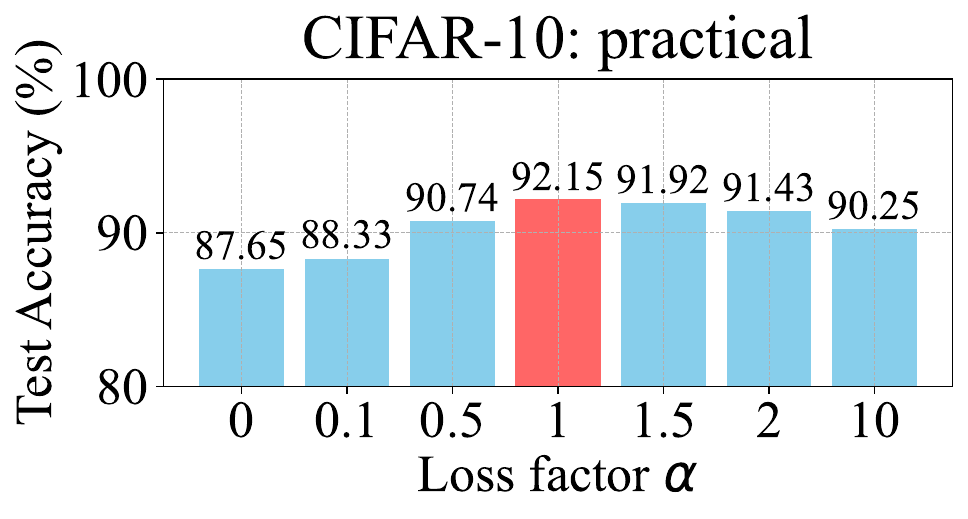}  }
\caption{The accuracy of FedCoSR with varying hyperparameters.}
\label{Fig:ablation hyper}
\end{figure}

First, we study the hyperparameter sensitivity of the proposed FedCoSR by conducting ablation studies on batch size $b$ and loss factor $\alpha$ using CIFAR-10 under practical heterogeneity. The results are depicted in Fig.~\ref{Fig:ablation hyper}.

As shown in Fig.~\ref{Fig:ablation b}, the model achieves peak performance at an optimal batch size, beyond which further increases lead to diminishing returns or even a decline. Notably, varying the batch size has minimal impact on our CRL-based method, as each local representation interacts with all global representations (equal to the total number of classes, $C$). Thus, a batch size of 16 is optimal considering both performance and computational efficiency.

As shown in Fig.~\ref{Fig:ablation alpha}, our method still performs the best when $\alpha=1$.
Moreover, when $\alpha=1.5$ or $\alpha=2$, which assigns more weight to the regularization term (1.5 or 2 times more than the personalization term), the accuracy is significantly higher than that when $\alpha=0.5$.
The results indicate that the regularization contributes more to the performance than the personalization term on the practical heterogeneous CIFAR-10 dataset.

\begin{figure}[t]
    \centering
    \includegraphics[width=0.48\textwidth]{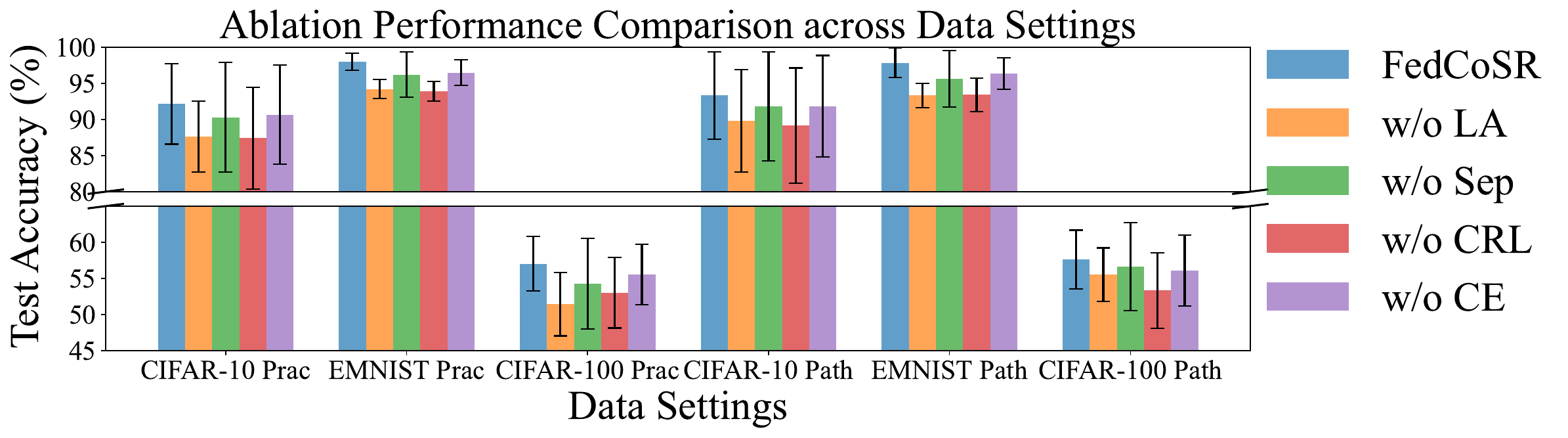}
    \caption{The accuracy of FedCoSR and its ablated variants under the default settings.}
    \label{Fig:abla}
\end{figure}

Next, to validate the effectiveness of main techniques employed in FedCoSR, we remove them and create three variants (``$w/o$'' is short for ``without''): 1) FedCoSR $w/o$ LA: $w/o$ loss-wise local aggregation; 2) FedCoSR $w/o$ Sep: $w/o$ separating the representation layer $f$ and the linear layer $g$; 3) FedCoSR $w/o$ CRL: $w/o$ CRL loss term in local training; 4) FedCoSR $w/o$ CE: $w/o$ CE, the supervised loss term, in local training. The results are shown in Fig.~\ref{Fig:abla}. 

The results in Fig.~\ref{Fig:abla} present the contribution of each ablated parts: 1) Removing LA leads to a significant performance drop, indicating its essential role in aggregating model-level information across clients to enhance accuracy and fairness; 2) When Sep is disabled, the overall accuracy impact is minor, but the variance across clients increases notably, showing that model separation helps stabilize convergence and preserve local knowledge, thus improving fairness; 3) Removing either CE (for personalization) or CRL (for regularization) degrades both accuracy and fairness. CE enables tailoring to client-specific characteristics, while CRL plays a more critical role by integrating label-wise information across clients to regularize local models under data heterogeneity. These results confirm that each component contributes uniquely, with LA and CRL being particularly vital for achieving accurate and fair predictions in federated settings.

\begin{figure}[t]
    \centering   
    \subfigure{\label{Fig:origin}
    \includegraphics[width=.15\textwidth]{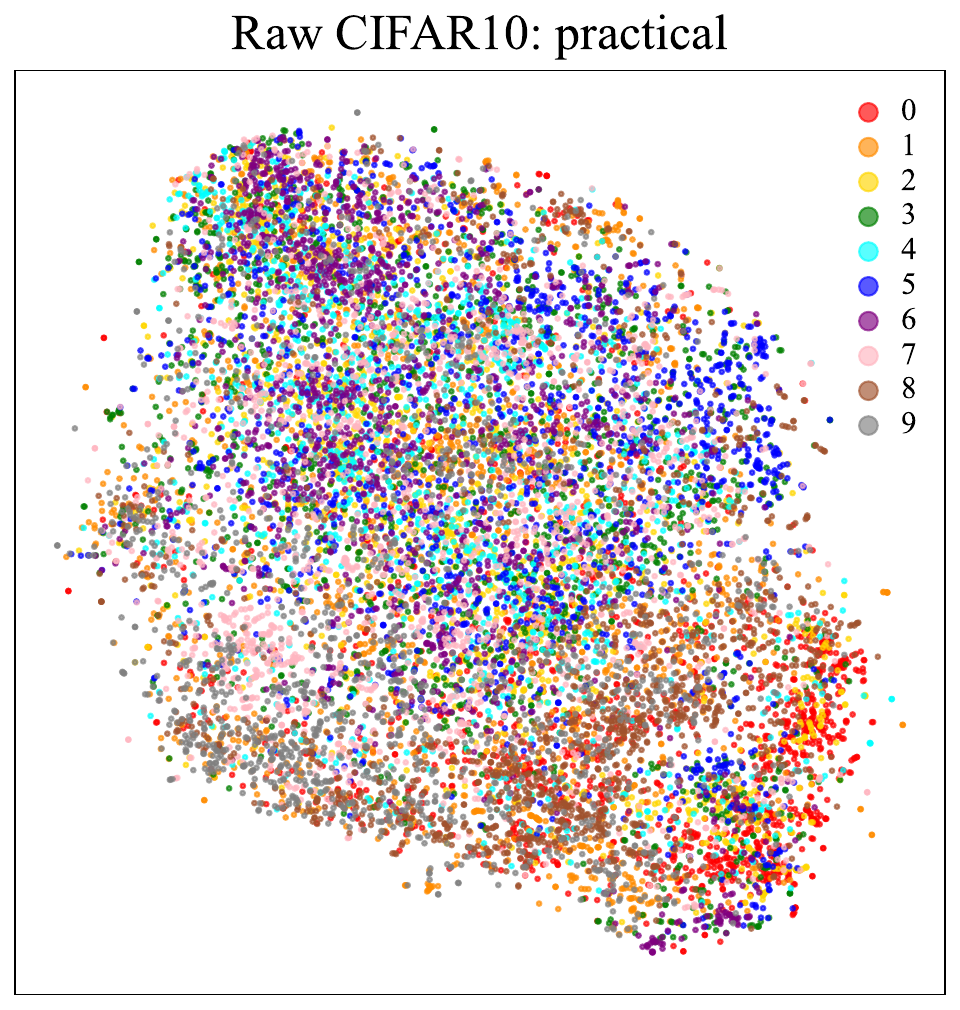}  }
    \subfigure{\label{Fig:rep_avg}
    \hspace{-8pt}\includegraphics[width=.15\textwidth]{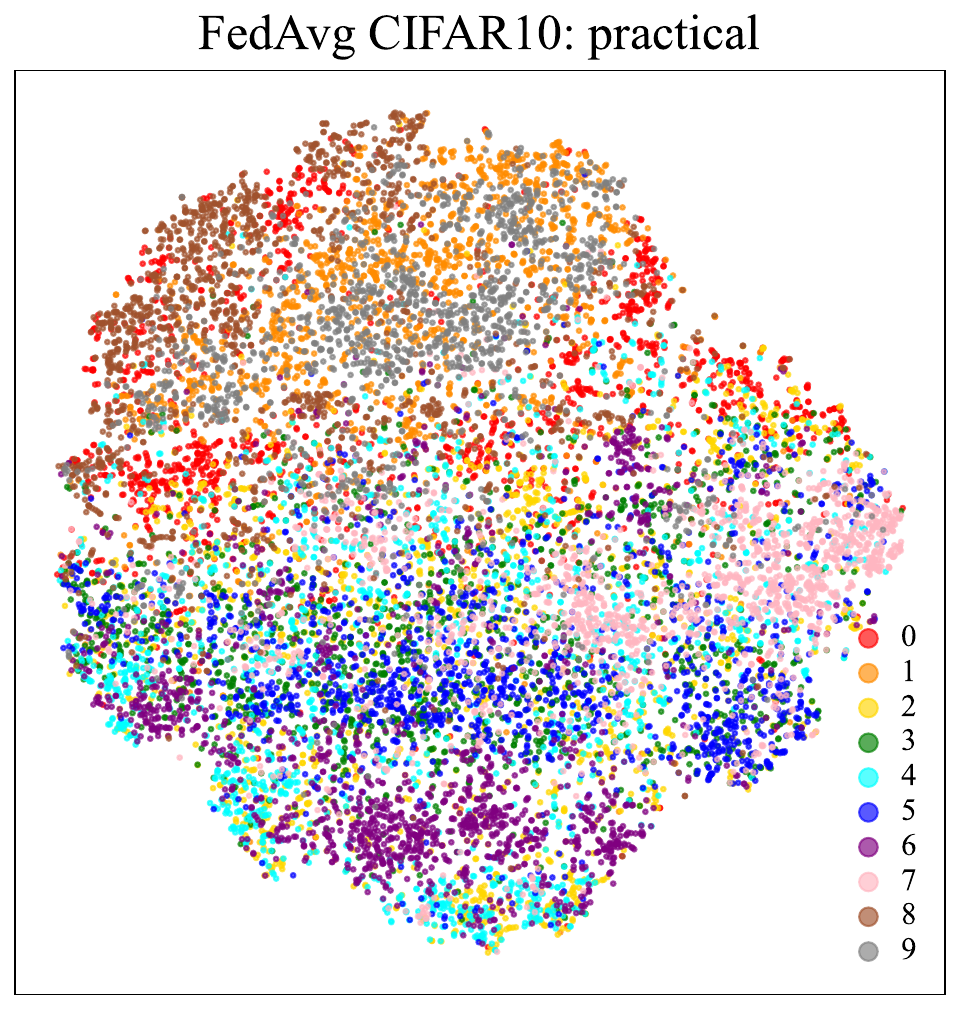}  }
    \subfigure{\label{Fig:rep_lg}
    \hspace{-8pt}\includegraphics[width=.15\textwidth]{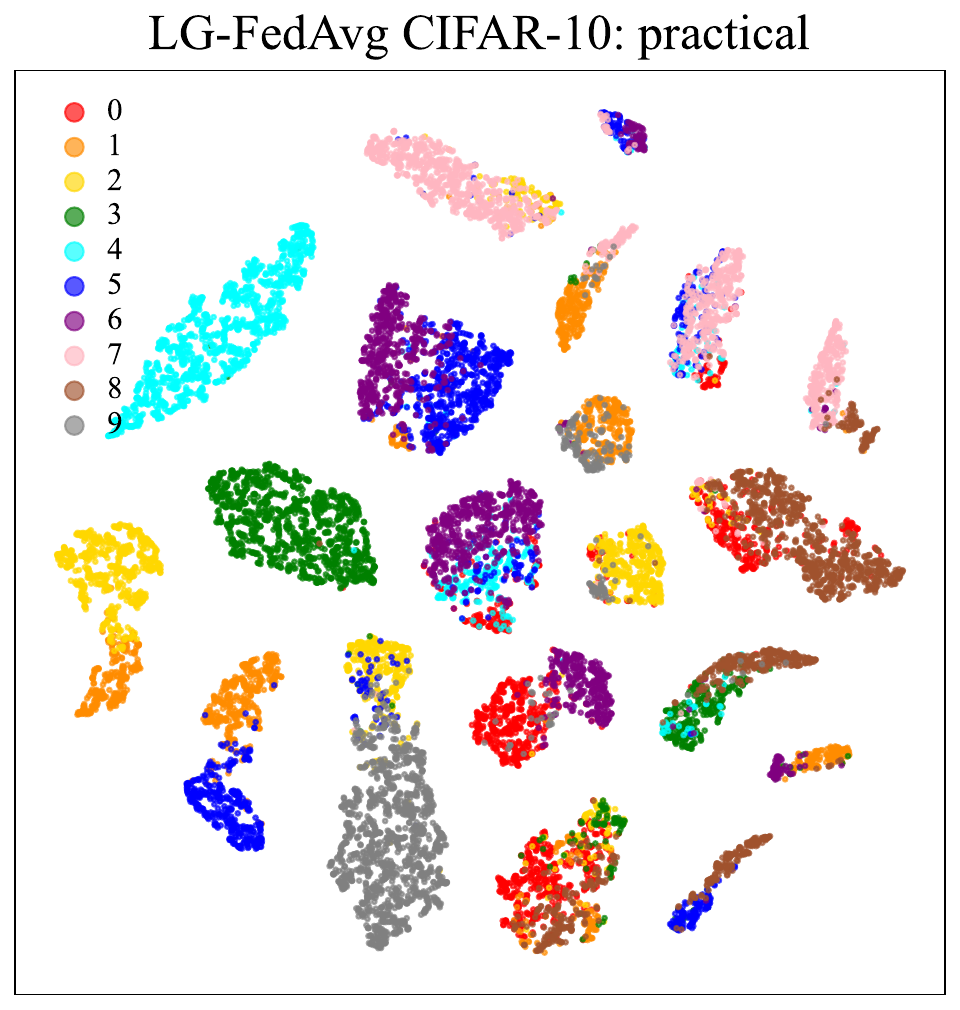}  }
    \subfigure{\label{Fig:rep_GH}
    \includegraphics[width=.15\textwidth]{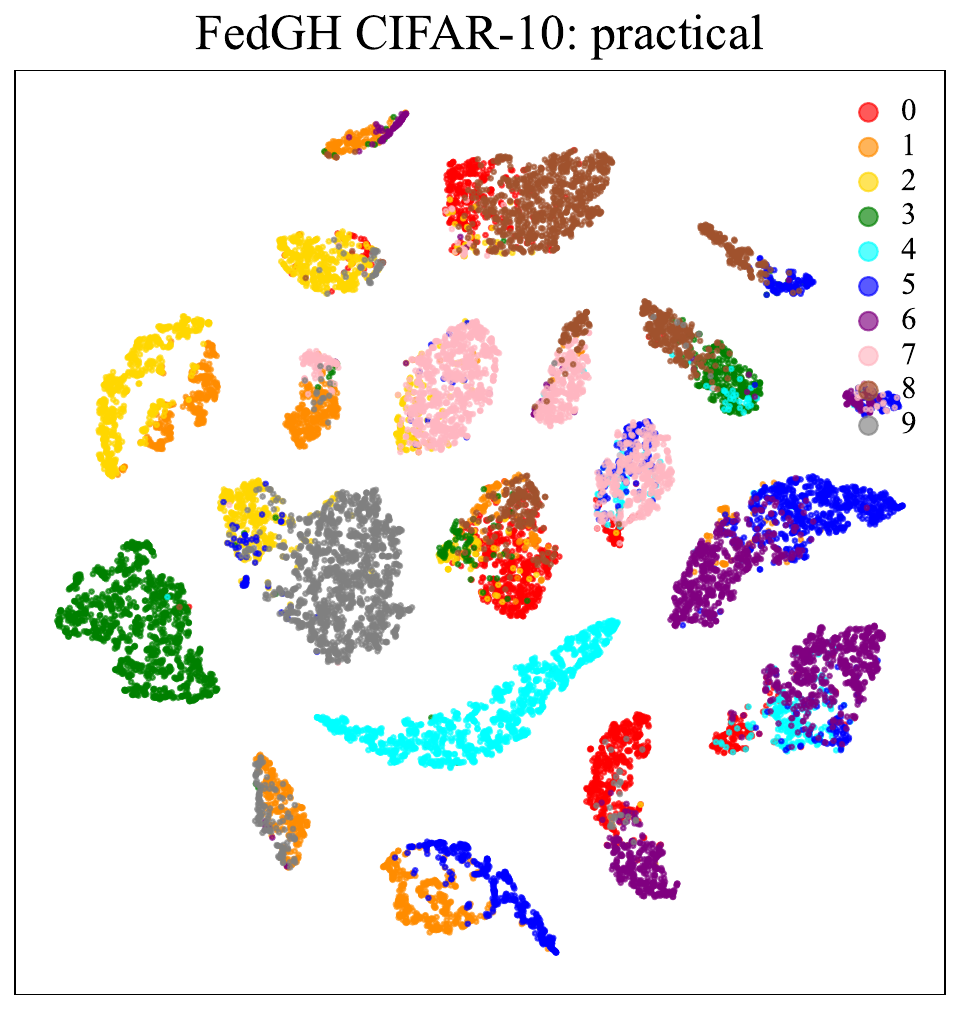}  }
    \subfigure{\label{Fig:rep_proto}
    \hspace{-8pt}\includegraphics[width=.15\textwidth]{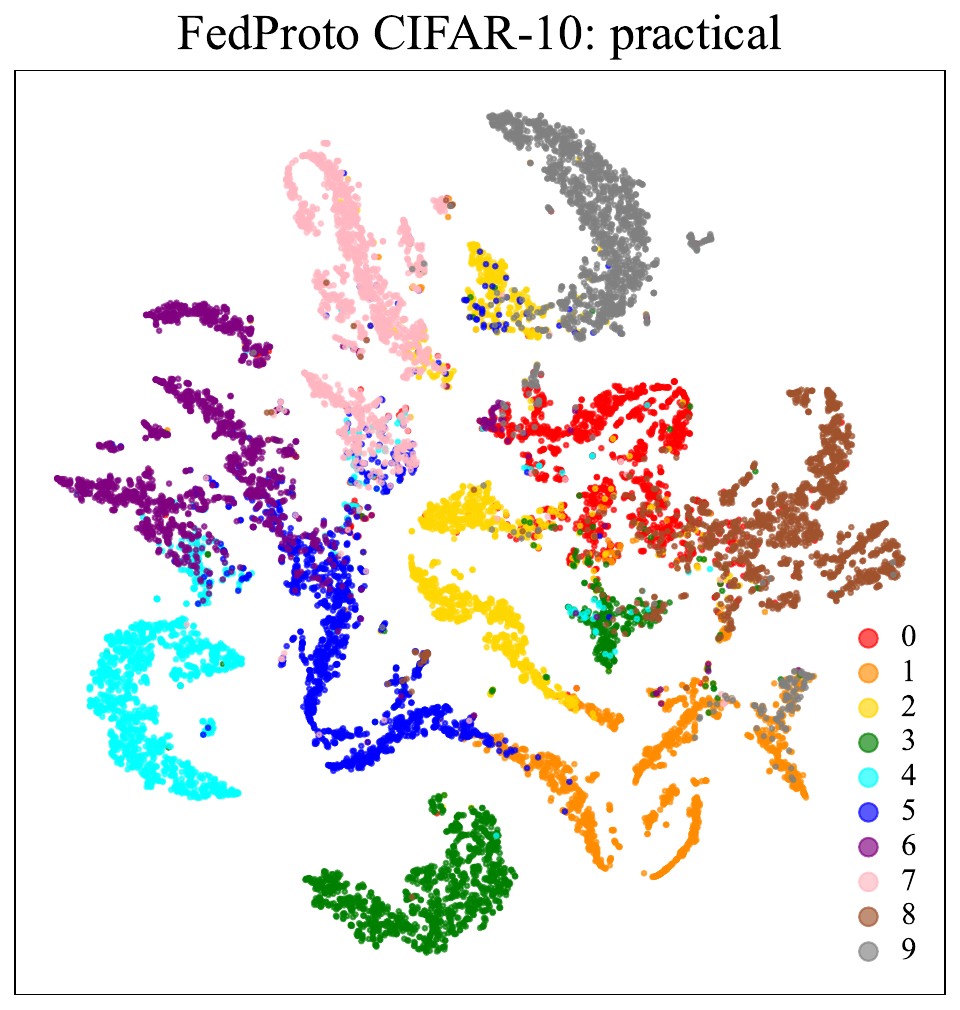}  }
    \subfigure{\label{Fig:FedCoSR}
    \hspace{-8pt}\includegraphics[width=.15\textwidth]{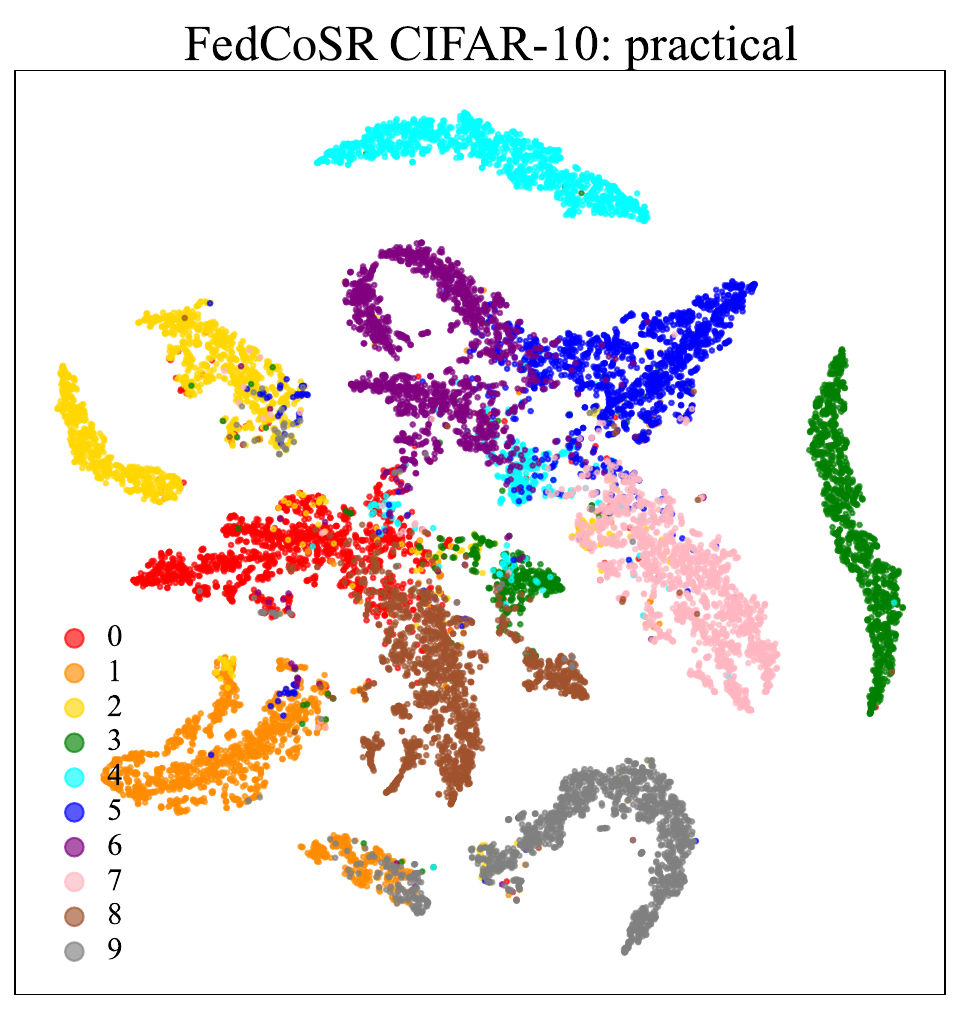}  }
    \captionsetup{width=.99\linewidth}
    \caption{Practical setting: Visualization of the raw data, representations of FedCoSR and other four baselines through t-SNE.}
    \label{Fig:rep_visual_prac}
\end{figure}

\subsubsection{Visualization of Representations}\label{app:visual}

We visualize the CIFAR-10 dataset using t-SNE. Fig.~\ref{Fig:rep_visual_prac} shows that while FedAvg achieves some clustering, PFL demonstrates more distinct clusters based on data patterns. Model splitting-based methods like LG-FedAvg and FedGH mix at least two label classes within clusters due to incomplete aggregation, hindering effective fusion of representation and projection layers, making model-splitting suboptimal. In contrast, FedProto and FedCoSR more clearly separate representations into uniform clusters, with FedProto showing some overlap between different labels, especially in the pathological setting. This occurs because FedProto lacks model-level information and focuses only on local personalization, whereas FedCoSR combines model aggregation with shared representation learning, balancing generalization and personalization.

\section{Conclusion and Future Work}\label{Sec:conclusion}
This paper presents FedCoSR, a PFL framework that applies contrastive learning to shareable representations to deal with label heterogeneity, including label distribution skew and data scarcity. It enhances local model training by leveraging global representations to form sample pairs, thereby enriching the knowledge of clients, especially those with limited data. The proposed loss-wise weighting model aggregation dynamically balances local and global models, ensuring personalized performance. Experiment results demonstrate that FedCoSR outperforms state-of-the-art methods in various heterogeneous settings, showing its effectiveness and fairness with heterogeneous or scarce data. Future work will extend the practicality of FedCoSR by studying its potential for addressing other forms of statistical heterogeneity, including feature condition skew, where clients exhibit similar label distributions but distinct sample distributions.

\maketitle

\small
\bibliographystyle{IEEEtran}
\bibliography{reference}

\begin{thebibliography}{10}
\providecommand{\url}[1]{#1}
\csname url@samestyle\endcsname
\providecommand{\newblock}{\relax}
\providecommand{\bibinfo}[2]{#2}
\providecommand{\BIBentrySTDinterwordspacing}{\spaceskip=0pt\relax}
\providecommand{\BIBentryALTinterwordstretchfactor}{4}
\providecommand{\BIBentryALTinterwordspacing}{\spaceskip=\fontdimen2\font plus
\BIBentryALTinterwordstretchfactor\fontdimen3\font minus \fontdimen4\font\relax}
\providecommand{\BIBforeignlanguage}[2]{{%
\expandafter\ifx\csname l@#1\endcsname\relax
\typeout{** WARNING: IEEEtran.bst: No hyphenation pattern has been}%
\typeout{** loaded for the language `#1'. Using the pattern for}%
\typeout{** the default language instead.}%
\else
\language=\csname l@#1\endcsname
\fi
#2}}
\providecommand{\BIBdecl}{\relax}
\BIBdecl

\bibitem{fortino2020internet}
G.~Fortino, C.~Savaglio, G.~Spezzano, and M.~Zhou, ``Internet of things as system of systems: A review of methodologies, frameworks, platforms, and tools,'' \emph{IEEE Transactions on Systems, Man, and Cybernetics: Systems}, vol.~51, no.~1, pp. 223--236, 2021.

\bibitem{lim2020federated}
W.~Y.~B. Lim, N.~C. Luong, D.~T. Hoang, Y.~Jiao, Y.-C. Liang, Q.~Yang, D.~Niyato, and C.~Miao, ``Federated learning in mobile edge networks: A comprehensive survey,'' \emph{IEEE Communications Surveys \& Tutorials}, vol.~22, no.~3, pp. 2031--2063, 2020.

\bibitem{li2023fedlga}
X.~Li, Z.~Qu, B.~Tang, and Z.~Lu, ``Fedlga: Toward system-heterogeneity of federated learning via local gradient approximation,'' \emph{IEEE Transactions on Cybernetics}, vol.~54, no.~1, pp. 401--414, 2024.

\bibitem{zhang2023federated}
C.~Zhang, Y.~Xie, H.~Bai, X.~Hu, B.~Yu, and Y.~Gao, ``Federated active semi-supervised learning with communication efficiency,'' \emph{IEEE Transactions on Systems, Man, and Cybernetics: Systems}, vol.~53, no.~11, pp. 6744--6756, 2023.

\bibitem{kaheni2024selective}
M.~Kaheni, M.~Lippi, A.~Gasparri, and M.~Franceschelli, ``Selective trimmed average: A resilient federated learning algorithm with deterministic guarantees on the optimality approximation,'' \emph{IEEE Transactions on Cybernetics}, vol.~54, no.~8, pp. 4402--4415, 2024.

\bibitem{zhang2021privacy}
L.~Zhang, W.~Cui, B.~Li, Z.~Chen, M.~Wu, and T.~S. Gee, ``Privacy-preserving cross-environment human activity recognition,'' \emph{IEEE Transactions on Cybernetics}, vol.~53, no.~3, pp. 1765--1775, 2023.

\bibitem{liang2022secure}
S.~Liang, J.~Lam, and H.~Lin, ``Secure estimation with privacy protection,'' \emph{IEEE Transactions on Cybernetics}, vol.~53, no.~8, pp. 4947--4961, 2023.

\bibitem{mcmahan2017communication}
B.~McMahan, E.~Moore, D.~Ramage, S.~Hampson, and B.~A. y~Arcas, ``Communication-efficient learning of deep networks from decentralized data,'' in \emph{Artificial Intelligence and Statistics}.\hskip 1em plus 0.5em minus 0.4em\relax PMLR, 2017, pp. 1273--1282.

\bibitem{kairouz2021advances}
P.~Kairouz, H.~B. McMahan, B.~Avent, A.~Bellet, M.~Bennis, A.~N. Bhagoji, K.~Bonawitz, Z.~Charles, G.~Cormode, R.~Cummings \emph{et~al.}, ``Advances and open problems in federated learning,'' \emph{Foundations and Trends{\textregistered} in Machine Learning}, vol.~14, no. 1--2, pp. 1--210, 2021.

\bibitem{ye2023heterogeneous}
M.~Ye, X.~Fang, B.~Du, P.~C. Yuen, and D.~Tao, ``Heterogeneous federated learning: State-of-the-art and research challenges,'' \emph{ACM Computing Surveys}, vol.~56, no.~3, pp. 1--44, 2023.

\bibitem{imteaj2021survey}
A.~Imteaj, U.~Thakker, S.~Wang, J.~Li, and M.~H. Amini, ``A survey on federated learning for resource-constrained iot devices,'' \emph{IEEE Internet of Things Journal}, vol.~9, no.~1, pp. 1--24, 2022.

\bibitem{tan2022towards}
A.~Z. Tan, H.~Yu, L.~Cui, and Q.~Yang, ``Towards personalized federated learning,'' \emph{IEEE Transactions on Neural Networks and Learning Systems}, vol.~34, no.~12, pp. 9587--9603, 2023.

\bibitem{10909241}
L.~Gao, B.~Liu, P.~Fu, M.~Xu, Y.~Zhang, and Y.~Huang, ``Self-supervised pretraining with multimodality representation enhancement for salient object detection in {RGB-D} images,'' \emph{IEEE Transactions on Instrumentation and Measurement}, vol.~74, pp. 1--14, 2025.

\bibitem{li2020federated}
T.~Li, A.~K. Sahu, M.~Zaheer, M.~Sanjabi, A.~Talwalkar, and V.~Smith, ``Federated optimization in heterogeneous networks,'' in \emph{Proceedings of Machine Learning and Systems}, vol.~2, 2020, pp. 429--450.

\bibitem{t2020personalized}
C.~T~Dinh, N.~Tran, and J.~Nguyen, ``Personalized federated learning with moreau envelopes,'' \emph{Advances in Neural Information Processing Systems}, vol.~33, pp. 21\,394--21\,405, 2020.

\bibitem{li2021ditto}
T.~Li, S.~Hu, A.~Beirami, and V.~Smith, ``Ditto: Fair and robust federated learning through personalization,'' in \emph{International Conference on Machine Learning}.\hskip 1em plus 0.5em minus 0.4em\relax PMLR, 2021, pp. 6357--6368.

\bibitem{arivazhagan2019federated}
M.~G. Arivazhagan, V.~Aggarwal, A.~K. Singh, and S.~Choudhary, ``Federated learning with personalization layers,'' \emph{arXiv preprint arXiv:1912.00818}, 2019.

\bibitem{liang2020think}
P.~P. Liang, T.~Liu, L.~Ziyin, N.~B. Allen, R.~P. Auerbach, D.~Brent, R.~Salakhutdinov, and L.-P. Morency, ``Think locally, act globally: Federated learning with local and global representations,'' \emph{arXiv preprint arXiv:2001.01523}, 2020.

\bibitem{li2021model}
Q.~Li, B.~He, and D.~Song, ``Model-contrastive federated learning,'' in \emph{2021 IEEE/CVF Conference on Computer Vision and Pattern Recognition (CVPR)}, 2021, pp. 10\,708--10\,717.

\bibitem{collins2021exploiting}
L.~Collins, H.~Hassani, A.~Mokhtari, and S.~Shakkottai, ``Exploiting shared representations for personalized federated learning,'' in \emph{International Conference on Machine Learning}.\hskip 1em plus 0.5em minus 0.4em\relax PMLR, 2021, pp. 2089--2099.

\bibitem{11075729}
M.~Xu, Z.~Sun, Y.~Hu, H.~Tang, Y.~Hu, X.~Song, and L.~Nie, ``Superpixel segmentation with edge guided local-global attention network,'' \emph{IEEE Transactions on Circuits and Systems for Video Technology}, pp. 1--13, 2025.

\bibitem{wang2023federated}
S.~Wang, X.~Fu, K.~Ding, C.~Chen, H.~Chen, and J.~Li, ``Federated few-shot learning,'' in \emph{Proceedings of the 29th ACM SIGKDD Conference on Knowledge Discovery and Data Mining}, 2023, pp. 2374--2385.

\bibitem{bengio2013representation}
Y.~Bengio, A.~Courville, and P.~Vincent, ``Representation learning: A review and new perspectives,'' \emph{IEEE Transactions on Pattern Analysis and Machine Intelligence}, vol.~35, no.~8, pp. 1798--1828, 2013.

\bibitem{karg2020efficient}
B.~Karg and S.~Lucia, ``Efficient representation and approximation of model predictive control laws via deep learning,'' \emph{IEEE Transactions on Cybernetics}, vol.~50, no.~9, pp. 3866--3878, 2020.

\bibitem{tan2022fedproto}
Y.~Tan, G.~Long, L.~Liu, T.~Zhou, Q.~Lu, J.~Jiang, and C.~Zhang, ``Fedproto: Federated prototype learning across heterogeneous clients,'' \emph{Proceedings of the AAAI Conference on Artificial Intelligence}, vol.~36, no.~8, pp. 8432--8440, Jun. 2022.

\bibitem{FedGH}
L.~Yi, G.~Wang, X.~Liu, Z.~Shi, and H.~Yu, ``{FedGH}: Heterogeneous federated learning with generalized global header,'' in \emph{Proceedings of the 31st ACM International Conference on Multimedia}, 2023, pp. 8686--8696.

\bibitem{xu2022personalized}
J.~Xu, X.~Tong, and S.-L. Huang, ``Personalized federated learning with feature alignment and classifier collaboration,'' in \emph{The Eleventh International Conference on Learning Representations}, 2023.

\bibitem{chen2020simple}
T.~Chen, S.~Kornblith, M.~Norouzi, and G.~Hinton, ``A simple framework for contrastive learning of visual representations,'' in \emph{International Conference on Machine Learning}.\hskip 1em plus 0.5em minus 0.4em\relax PMLR, 2020, pp. 1597--1607.

\bibitem{fallah2020personalized}
A.~Fallah, A.~Mokhtari, and A.~Ozdaglar, ``Personalized federated learning with theoretical guarantees: A model-agnostic meta-learning approach,'' \emph{Advances in Neural Information Processing Systems}, vol.~33, pp. 3557--3568, 2020.

\bibitem{10856888}
D.~Wang, Y.~Gao, S.~Pang, C.~Zhang, X.~Zhang, and M.~Li, ``{FedMPS}: A robust differential privacy federated learning based on local model partition and sparsification for heterogeneous iiot data,'' \emph{IEEE Internet of Things Journal}, vol.~12, no.~10, pp. 13\,757--13\,768, 2025.

\bibitem{huang2021personalized}
Y.~Huang, L.~Chu, Z.~Zhou, L.~Wang, J.~Liu, J.~Pei, and Y.~Zhang, ``Personalized cross-silo federated learning on non-iid data,'' \emph{Proceedings of the AAAI Conference on Artificial Intelligence}, vol.~35, no.~9, pp. 7865--7873, May 2021.

\bibitem{zhang2023fedala}
J.~Zhang, Y.~Hua, H.~Wang, T.~Song, Z.~Xue, R.~Ma, and H.~Guan, ``Fed{ALA}: Adaptive local aggregation for personalized federated learning,'' \emph{Proceedings of the AAAI Conference on Artificial Intelligence}, vol.~37, no.~9, pp. 11\,237--11\,244, Jun. 2023.

\bibitem{10835748}
P.~Wei, T.~Zhou, W.~Liu, J.~Du, T.~Wang, and G.~Yue, ``{FedPDN}: Personalized federated learning with interclass similarity constraint for medical image classification through parameter decoupling,'' \emph{IEEE Transactions on Instrumentation and Measurement}, vol.~74, pp. 1--13, 2025.

\bibitem{10655727}
X.~Yang, W.~Huang, and M.~Ye, ``Fed{AS}: Bridging inconsistency in personalized federated learning,'' in \emph{2024 IEEE/CVF Conference on Computer Vision and Pattern Recognition (CVPR)}, 2024, pp. 11\,986--11\,995.

\bibitem{scott2024pefll}
J.~Scott, H.~Zakerinia, and C.~H. Lampert, ``Pe{FLL}: Personalized federated learning by learning to learn,'' in \emph{The Twelfth International Conference on Learning Representations}, 2024.

\bibitem{10988632}
T.~Gao, K.~Liu, Y.~Yang, X.~Liu, P.~Zhang, and G.~Wang, ``{FedPC}: An efficient prototype-based clustered federated learning on medical imaging,'' \emph{IEEE Journal of Biomedical and Health Informatics}, pp. 1--13, 2025.

\bibitem{mclaughlin2025personalized}
C.~Mclaughlin and L.~Su, ``Personalized federated learning via feature distribution adaptation,'' \emph{Advances in Neural Information Processing Systems}, vol.~37, pp. 77\,038--77\,059, 2025.

\bibitem{10572001}
T.~Deng, Y.~Huang, G.~Han, Z.~Shi, J.~Lin, Q.~Dou, Z.~Liu, X.-j. Guo, C.~L. Philip~Chen, and C.~Han, ``{FedDBL}: Communication and data efficient federated deep-broad learning for histopathological tissue classification,'' \emph{IEEE Transactions on Cybernetics}, vol.~54, no.~12, pp. 7851--7864, 2024.

\bibitem{10198520}
C.~B. Mawuli, L.~Che, J.~Kumar, S.~U. Din, Z.~Qin, Q.~Yang, and J.~Shao, ``{FedStream}: Prototype-based federated learning on distributed concept-drifting data streams,'' \emph{IEEE Transactions on Systems, Man, and Cybernetics: Systems}, vol.~53, no.~11, pp. 7112--7124, 2023.

\bibitem{10950126}
Y.~Sun, S.~Pan, A.~Sun, Z.~Fu, S.~Long, and Z.~Li, ``{FedLFP}: Communication-efficient personalized federated learning on non-iid data in mobile edge computing environments,'' \emph{IEEE Transactions on Mobile Computing}, vol.~24, no.~9, pp. 8811--8823, 2025.

\bibitem{10855800}
H.~Li, G.~Chen, B.~Wang, Z.~Chen, Y.~Zhu, F.~Hu, J.~Dai, and W.~Wang, ``{pFedKD}: Personalized federated learning via knowledge distillation using unlabeled pseudo data for internet of things,'' \emph{IEEE Internet of Things Journal}, vol.~12, no.~11, pp. 16\,314--16\,324, 2025.

\bibitem{shi2023ffedcl}
X.~Shi, L.~Yi, X.~Liu, and G.~Wang, ``{FFEDCL}: Fair federated learning with contrastive learning,'' in \emph{ICASSP 2023 - 2023 IEEE International Conference on Acoustics, Speech and Signal Processing (ICASSP)}, 2023, pp. 1--5.

\bibitem{zhang2023doubly}
Y.~Zhang, Y.~Xu, S.~Wei, Y.~Wang, Y.~Li, and X.~Shang, ``Doubly contrastive representation learning for federated image recognition,'' \emph{Pattern Recognition}, vol. 139, p. 109507, 2023.

\bibitem{khosla2020supervised}
P.~Khosla, P.~Teterwak, C.~Wang, A.~Sarna, Y.~Tian, P.~Isola, A.~Maschinot, C.~Liu, and D.~Krishnan, ``Supervised contrastive learning,'' \emph{Advances in Neural Information Processing Systems}, vol.~33, pp. 18\,661--18\,673, 2020.

\bibitem{10658073}
S.~Seo, J.~Kim, G.~Kim, and B.~Han, ``Relaxed contrastive learning for federated learning,'' in \emph{2024 IEEE/CVF Conference on Computer Vision and Pattern Recognition (CVPR)}, 2024, pp. 12\,279--12\,288.

\bibitem{tan2022federated}
Y.~Tan, G.~Long, J.~Ma, L.~Liu, T.~Zhou, and J.~Jiang, ``Federated learning from pre-trained models: A contrastive learning approach,'' \emph{Advances in Neural Information Processing Systems}, vol.~35, pp. 19\,332--19\,344, 2022.

\bibitem{mu2023fedproc}
X.~Mu, Y.~Shen, K.~Cheng, X.~Geng, J.~Fu, T.~Zhang, and Z.~Zhang, ``{FedProc}: Prototypical contrastive federated learning on non-iid data,'' \emph{Future Generation Computer Systems}, vol. 143, pp. 93--104, 2023.

\bibitem{miao2023fedseg}
J.~Miao, Z.~Yang, L.~Fan, and Y.~Yang, ``{FedSeg}: Class-heterogeneous federated learning for semantic segmentation,'' in \emph{2023 IEEE/CVF Conference on Computer Vision and Pattern Recognition (CVPR)}, 2023, pp. 8042--8052.

\bibitem{yu2023multimodal}
Q.~Yu, Y.~Liu, Y.~Wang, K.~Xu, and J.~Liu, ``Multimodal federated learning via contrastive representation ensemble,'' in \emph{The Eleventh International Conference on Learning Representations}, 2023.

\bibitem{liu2021self}
X.~Liu, F.~Zhang, Z.~Hou, L.~Mian, Z.~Wang, J.~Zhang, and J.~Tang, ``Self-supervised learning: Generative or contrastive,'' \emph{IEEE Transactions on Knowledge and Data Engineering}, vol.~35, no.~1, pp. 857--876, 2023.

\bibitem{yosinski2014transferable}
J.~Yosinski, J.~Clune, Y.~Bengio, and H.~Lipson, ``How transferable are features in deep neural networks?'' \emph{Advances in Neural Information Processing Systems}, vol.~27, 2014.

\bibitem{ghimire2022recent}
B.~Ghimire and D.~B. Rawat, ``Recent advances on federated learning for cybersecurity and cybersecurity for federated learning for internet of things,'' \emph{IEEE Internet of Things Journal}, vol.~9, no.~11, pp. 8229--8249, 2022.

\bibitem{de2005tutorial}
P.-T. De~Boer, D.~P. Kroese, S.~Mannor, and R.~Y. Rubinstein, ``A tutorial on the cross-entropy method,'' \emph{Annals of Operations Research}, vol. 134, pp. 19--67, 2005.

\bibitem{shamsian2021personalized}
A.~Shamsian, A.~Navon, E.~Fetaya, and G.~Chechik, ``Personalized federated learning using hypernetworks,'' in \emph{International Conference on Machine Learning}.\hskip 1em plus 0.5em minus 0.4em\relax PMLR, 2021, pp. 9489--9502.

\bibitem{huang2024fedcosrapp}
\BIBentryALTinterwordspacing
C.~Huang, X.~Chen, Y.~Zhang, and H.~Wang, ``Supplementary materials: {FedCoSR}: Personalized federated learning with contrastive shareable representations for label heterogeneity in non-iid data,'' 2024. [Online]. Available: \url{https://drive.google.com/file/d/1Q-tgjje26hDm9WIyTkKGqIGkPdlHvJs3/view?usp=sharing}
\BIBentrySTDinterwordspacing

\bibitem{jeon2021gradient}
J.~Jeon, K.~Lee, S.~Oh, J.~Ok \emph{et~al.}, ``Gradient inversion with generative image prior,'' \emph{Advances in Neural Information Processing Systems}, vol.~34, pp. 29\,898--29\,908, 2021.

\bibitem{wei2020federated}
K.~Wei, J.~Li, M.~Ding, C.~Ma, H.~H. Yang, F.~Farokhi, S.~Jin, T.~Q.~S. Quek, and H.~Vincent~Poor, ``Federated learning with differential privacy: Algorithms and performance analysis,'' \emph{IEEE Transactions on Information Forensics and Security}, vol.~15, pp. 3454--3469, 2020.

\end{thebibliography}

\begin{IEEEbiography}
[{\includegraphics[width=1in,height=1.25in,clip,keepaspectratio]{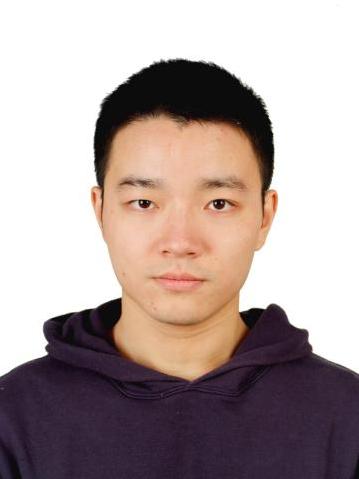}}] 
{Chenghao Huang}
received the B.E. degree in software engineering from University of Electronic Science and Technology of China (UESTC) in 2020, and M.S. degree in computer science and engineering from UESTC in 2023. He is currently pursuing the Ph.D. degree with the Faculty of Information Technology, Monash University. His research interests include deep learning, federated learning, reinforcement learning and smart grid.
\end{IEEEbiography}

\begin{IEEEbiography}[{\includegraphics[width=1in,height=1.25in,clip,keepaspectratio]{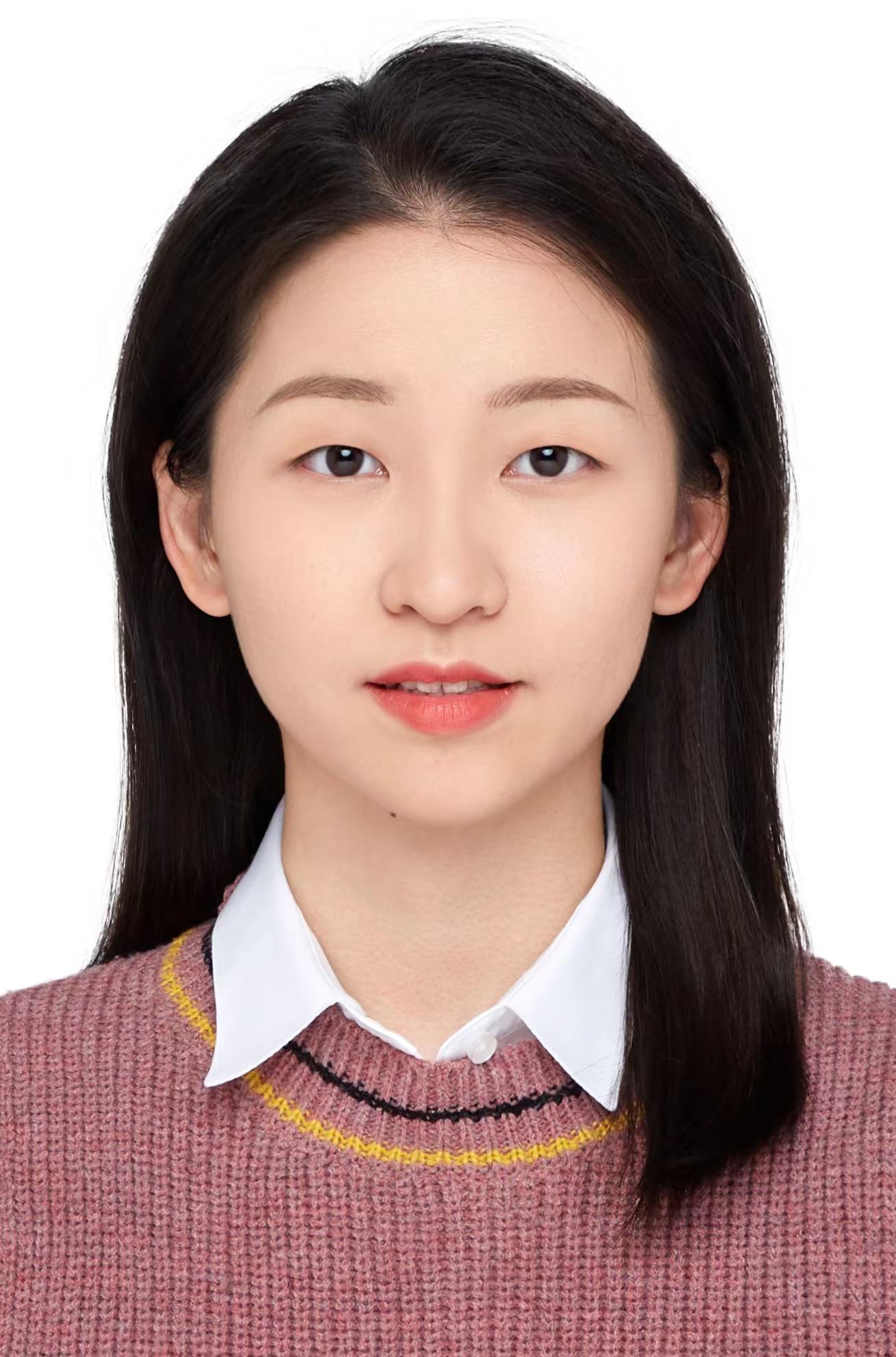}}]{Xiaolu Chen}
received a B.E. degree from the Department of Computer Science and Technology, at the University of Electronic Science and Technology of China, in 2022. She is currently pursuing M.S. degree with the Department of Computer Science and Technology, at the University of Electronic Science and Technology of China. Her main research interests include deep learning, federated learning, and smart grid.
\end{IEEEbiography}

\begin{IEEEbiography}[{\includegraphics[width=1in,height=1.25in,clip,keepaspectratio]{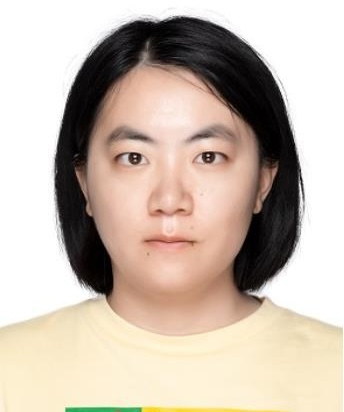}}]{Yanru Zhang}
(S'13-M'16) received the B.S. degree in electronic engineering from the University of Electronic Science and Technology of China (UESTC) in 2012, and the Ph.D. degree from the Department of Electrical and Computer Engineering, University of Houston (UH) in 2016. She worked as a Postdoctoral Fellow at UH and the Chinese University of Hong Kong successively. She is currently a Professor with the Shenzhen Institute for Advanced Study and School of Computer Science, UESTC. Her current research involves game theory, machine learning, and deep learning in network economics, Internet and applications, wireless communications, and networking. She received the Best Paper Award at IEEE HPCC 2022, DependSys 2022, ICCC 2017, and ICCS 2016. 
\end{IEEEbiography}

\begin{IEEEbiography}
[{\includegraphics[width=1in,height=1.25in,clip,keepaspectratio]{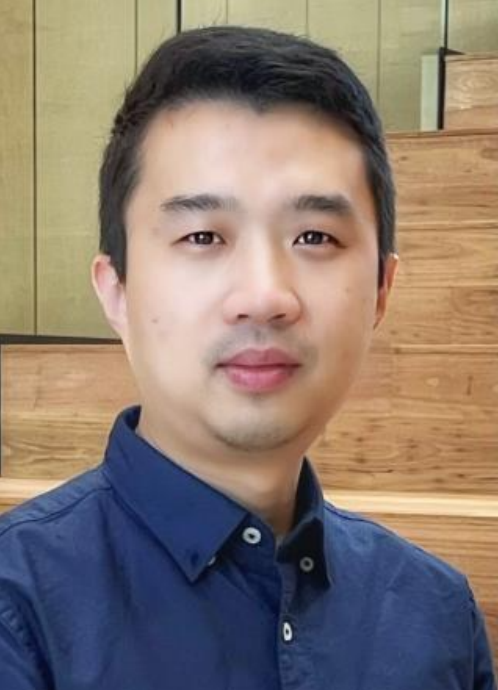}}]{Hao Wang} (M'16) received his Ph.D. in Information Engineering from The Chinese University of Hong Kong, Hong Kong, in 2016. He was a Postdoctoral Research Fellow at Stanford University, Stanford, CA, USA, and a Washington Research Foundation Innovation Fellow at the University of Washington, Seattle, WA, USA. He is currently a Senior Lecturer and ARC DECRA Fellow in the Department of Data Science and AI, Faculty of IT, Monash University, Melbourne, VIC, Australia. His research interests include optimization, machine learning, and data analytics for power and energy systems.
\end{IEEEbiography}

\vfill

\clearpage
\appendices
\section{Analysis and Proof of Local Loss Function}\label{app:loss_proof}

As shown in~\cite{de2005tutorial}, $\mathcal{L}_{\text{Per}}$, the cross entropy loss function, is convex and demonstrates good convergence. Then we make the following assumption and definition for explaining the effectiveness of $\mathcal{L}_{\text{Reg}}$.

\begin{assumption}[Sub-Gaussian Design]\label{app: ass sub gaussian}
Let $\forall \mathbf{x}^i \in \mathbb{R}^d$ be IID with mean $\mathbf{0} \in \mathbb{R}^d$ and covariance $\mathbf{I}_d \in \mathbb{R}^{d\times d}$, and assume they are $\mathbf{I}_d$-sub-Gaussian, i.e., $\mathbb{E}[e^{\mathbf{v}^\top \mathbf{x}^i}] \leq e^{\frac{||\mathbf{v}||_2^2}{2}}, \forall \mathbf{v}\in \mathbb{R}^d$.
\end{assumption}

\begin{definition}[Mutual Information]\label{app:loss-def-1}
Mutual information is an evaluation of the statistical correspondence between 2 stochastic variables:
\begin{align}
    I(X;Y) = \sum_{x\in X,y\in Y} p(x,y) \log \bigg[\frac{p(x,y)}{p(x)p(y)}\bigg].
\end{align}
To construct contrastive learning task, mutual information is introduced between anchor samples $X$ and the similar samples $X^+$, also called positive sample:
\begin{align}
    I(X^+; X) &= \sum_{x^+\in X^+,x \in X} p(x^+,x) \log \bigg[\frac{p(x^+,x)}{p(x^+)p(x)}\bigg] \notag\\ 
    &= \sum_{x^+\in X^+,x \in X} p(x^+,x) \log \bigg[\frac{p(x^+|x)}{p(x^+)}\bigg]. \label{app:loss-def-eq2}
\end{align}

A scoring metric $h(\cdot)$ is used to evaluate the correspondence between $x$ and $x^+$. Referring to~\cite{chen2020simple}, $h(x^+,x)$ proportionally preserves the mutual information between $x^+$ and $x$:
\begin{align}
    h(x^+,x) \varpropto \frac{p(x^+|x)}{p(x^+)}. \label{app:loss-def-eq3}
\end{align}
\end{definition}

Then, we have the following theorem.
\begin{theorem}[InfoNCE Minimization in FedCoSR]\label{app:theo-loss-1}
Minimizing InfoNCE equals to maximizing the mutual information between the anchor representation and its positive sample, and meanwhile minimizing the mutual information between it and its negative samples.
\end{theorem}

Based on Assumption~\ref{app: ass sub gaussian}, Definition~\ref{app:loss-def-1}, Definition~\ref{def:rep_CL}, Theorem~\ref{app:theo-loss-1}, and Eq. (\!\!~\ref{Eq11}) in FedCoSR, optimizing the local regularization objective $\mathcal{L}^i_{\text{Reg}}$ can enlarge the information enhancement for the $i$th client. $\mathcal{L}^i_{\text{Reg}}$ can be optimized by minimizing infoNCE between the anchor representations $\omega^i_{j,t+1}$ and the global representations $\bar{\Omega}^{\text{Global}}_{t+1}$. This guarantees the enhancement of combining $\mathcal{L}_{\text{Per}}$ with $\mathcal{L}_{\text{Reg}}$.

\begin{remark}[Nature of CRL in FedCoSR]\label{app:remark-1}
    Since the labels in the regularization loss of FedCoSR are all confirmed, the CRL here can also be regraded as a variant of supervised contrastive learning~\cite{khosla2020supervised}, whose infoNCE loss is similar to soft-nearest neighbors loss.
\end{remark}

\textbf{The proof of Theorem}~\ref{app:theo-loss-1}:

Given an anchor representation $\omega^i_j=\hat{\theta}^i_{t+1}(\mathbf{x}^i_j)$ whose $y^i_j=c$, the global representations $\bar{\Omega}^{\text{Global}}_t$ are split into the positive sample $\bar{\omega}^{\text{Global}}_{c,t}$ and the negative samples $\{\bar{\omega}^{\text{Global}}_{\hat{c},t}\}_{\hat{c}}^{\mathcal{C}\backslash c}$. Based on Eq. (\!\!~\ref{app:loss-def-eq3}) in Definition~\ref{app:loss-def-1} and Eq. (\!\!~\ref{Eq11}), we set $h(x,y)$ as $e^{[D(x,y)/\tau_{\text{CL}}]}$. Briefly, we abbreviate $\mathbb{E}_{\mathcal{D}^i_{b,t+1}\sim\mathcal{D}^i}\mathbb{E}_{(\mathbf{x}^i_j,\mathbf{y}^i_j) \sim \mathcal{D}^i_{b,t+1}}$ in Eq. (\!\!~\ref{Eq11}) as $\mathbb{E}_{\omega^i_j \sim \Omega^i_{t+1}}$. Then we give the proof:
\begin{equation}
\begin{aligned}
    \mathcal{L}^i_{\text{Reg},t+1} &= -\mathbb{E}_{\omega^i_j \sim \Omega^i_{t+1}} \log \bigg[ \frac{\frac{p(\bar{\omega}^{\text{Global}}_{c,t}|\omega^i_j)}{p(\bar{\omega}^{\text{Global}}_{c,t})}}{\frac{p(\bar{\omega}^{\text{Global}}_{c,t}|\omega^i_j)}{p(\bar{\omega}^{\text{Global}}_{c,t})} + \sum_{\hat{c} \sim \mathcal{C}\backslash c} \frac{p(\bar{\omega}^{\text{Global}}_{\hat{c},t}|\omega^i_j)}{p(\bar{\omega}^{\text{Global}}_{\hat{c},t})}}\bigg]\\
    &= \mathbb{E}_{\omega^i_j \sim \Omega^i_{t+1}} \log \bigg[1+ \frac{p(\bar{\omega}^{\text{Global}}_{c,t})}{p(\bar{\omega}^{\text{Global}}_{c,t}|\omega^i_j)} \sum_{\hat{c} \sim \mathcal{C}\backslash c} \frac{p(\bar{\omega}^{\text{Global}}_{\hat{c},t}|\omega^i_j)}{p(\bar{\omega}^{\text{Global}}_{\hat{c},t})} \bigg] \notag \\
\end{aligned}
\end{equation}
\begin{equation}
\begin{aligned}
    &\thickapprox \mathbb{E}_{\omega^i_j \sim \Omega^i_{t+1}} \log \bigg[ 1+ \frac{p(\bar{\omega}^{\text{Global}}_{c,t})}{p(\bar{\omega}^{\text{Global}}_{c,t}|\omega^i_j)} (C-1) \mathbb{E}_{\hat{c}} \big[ \frac{p(\bar{\omega}^{\text{Global}}_{\hat{c},t}|\omega^i_j)}{p(\bar{\omega}^{\text{Global}}_{\hat{c},t})}\big] \bigg] \\
    &= \mathbb{E}_{\omega^i_j \sim \Omega^i_{t+1}} \log \bigg[ 1+ (C-1)\frac{p(\bar{\omega}^{\text{Global}}_{c,t})}{p(\bar{\omega}^{\text{Global}}_{c,t}|\omega^i_j)} \bigg] \\
    &\geq \mathbb{E}_{\omega^i_j \sim \Omega^i_{t+1}} \log \bigg[ C \frac{p(\bar{\omega}^{\text{Global}}_{c,t})}{p(\bar{\omega}^{\text{Global}}_{c,t}|\omega^i_j)} \bigg] \\
    &\overset{(a)}{=} -I(\bar{\omega}^{\text{Global}}_{c,t}, \omega^i_j) + \log (C),
\end{aligned}
\end{equation}
where $(a)$ follows the Eq. (\!\!~\ref{app:loss-def-eq2}) in Definition~\ref{app:loss-def-1}.

Thus we get $I(\cdot) \geq \log (C) - \mathcal{L}_{\text{Reg}}$. When $C$ becomes larger, the lower bound of similar representations will increase, making the final performance of FedCoSR more accurate.

\section{Convergence Analysis and Proof of FL Communication}\label{app:FL_proof}

Without loss of generality, we denote the total number of local training epochs as $R$, which is set as 1 in this paper. Specifically, we define $tR+r$ as the $r$th epoch at iteration $t$, $tR$ as the end of iteration $t$ (end of the local training), $tR+0$ as the beginning of iteration $t$ (local aggregation), and $tR+\frac{1}{2}$ as the point where global representations are utilized (after local aggregation) at iteration $t$. Then, we give convergence analysis and proof for the $i$th client.

\subsection{Assumptions}
\begin{assumption}[Lipschitz Smoothness]\label{ass:app-1}
    The $i$th local loss function is $L_1$-Lipschitz smooth, implying that the gradient of the local loss function is $L_1$-Lipschitz continuous:
    \begin{align}
        &||\nabla \mathcal{L}^i_{tR+r_1}(\mathbf{x}^i,\mathbf{y}^i;\hat{\theta}^i_{tR+r_2}) - \nabla \mathcal{L}^i_{tR+r_2}(\mathbf{x}^i,\mathbf{y}^i;\hat{\theta}^i_{tR+r_2})||_2 \notag \\
        &\leq L_1||\hat{\theta}^i_{tR+r_1} - \hat{\theta}^i_{tR+r_2}||_2, \\
        &\forall r_1, r_2 \in \{0,\frac{1}{2},1,...,R\}, i\in \{1,...,N\}, \notag\\
        &(\mathbf{x}^i,\mathbf{y}^i)\in\mathcal{D}^i, L_1 > 0, t>0, R>0, \notag
    \end{align}
    which derives the following quadratic bound:
    \begin{align}
        &\mathcal{L}^i_{tR+r_1} - \mathcal{L}^i_{tR+r_2}
        \leq \notag \\ 
        & \langle\nabla\mathcal{L}^i_{tR+r_1},(\hat{\theta}^i_{tR+r_1}-\hat{\theta}^i_{tR+r_2})\rangle + \frac{L_1}{2}||\hat{\theta}^i_{tR+r_1}-\hat{\theta}^i_{tR+r_2}||^2_2. \label{ass1-eq2}
    \end{align}
\end{assumption}

\begin{assumption}[Unbiased Gradient and Bounded Variance]\label{ass:app-3}
The stochastic gradient $\nabla\mathcal{L}^i_{tR}(\hat{\theta}^i_{tR};\mathcal{D}^i_{b,tR})$ is an unbiased estimator of the local gradient, whose expectation is:
\begin{align}
    \mathbb{E}_{\mathcal{D}^i_{b,tR}\in\mathcal{D}^i}[\nabla\mathcal{L}^i_{tR}(\hat{\theta}^i_{tR};\mathcal{D}^i_{b,tR})]=\nabla\mathcal{L}^i_{tR}(\hat{\theta}^i_{tR}), \label{ass3-eq1}
\end{align}
and the variance of $\nabla\mathcal{L}^i_{tR}(\hat{\theta}^i_{tR};\mathcal{D}^i_{b,tR})$ is bounded by $\sigma$:
\begin{align}
    & \text{Var}[\nabla \mathcal{L}^i_{tR}(\hat{\theta}^i_{tR};\mathcal{D}^i_{b,tR})]= \notag \\
    &\mathbb{E}_{\mathcal{D}^i_{b,tR}\in\mathcal{D}^i}\bigg[||\nabla\mathcal{L}^i_{tR}(\hat{\theta}^i_{tR};\mathcal{D}^i_{b,tR})-\nabla\mathcal{L}^i_{tR}(\hat{\theta}^i_{tR})||^2_2\bigg]\leq\sigma^2. \label{ass3-eq2}
\end{align}
\end{assumption}

\begin{assumption}[Bounded Variance of Representation Layers]\label{ass:app-4}
The variance between $f^i_{tR}$ and $f^{\text{Global}}_{tR}$ is bounded, whose parameter bound is:
\begin{align}
    \mathbb{E}[||f^i_{tR}-f^{\text{Global}}_{tR}||^2_2] \leq \varepsilon^2.
\end{align}
\end{assumption}

\begin{assumption}[Linear Invariance]\label{ass:app-7}
The linear combination of $f^i_{tR}$ and $f^{\text{Global}}_{tR}$ holds the above assumptions.
\end{assumption}

\subsection{Lemmas}

\begin{lemma}[Bounded Multi-Step Training Loss]\label{app:fl-lem1}
Based on Assumption~\ref{ass:app-1} and~\ref{ass:app-3}, for the whole iteration $t$, the local loss function of the $i$th client during the local training can be bounded as:
\begin{align}
    & \mathbb{E}[\mathcal{L}^i_{(t+1)R}] \leq \notag \\
    & \mathbb{E}[\mathcal{L}^i_{tR+\frac{1}{2}}] - \big(\eta-\frac{L_1 \eta^2}{2}\big) \sum_{r=\frac{1}{2}}^R||\nabla\mathcal{L}^i_{tR+r}||^2_2 + \frac{RL_1\eta^2\sigma^2}{2},
\end{align}
where $\eta$ is the learning rate.
\end{lemma}

\textbf{The proof of Lemma~\ref{app:fl-lem1}}:

In general, we establish the following proposition.
\begin{proposition}[]\label{prop:app-1}
For multi-step local training in the same iteration, the model update equals to the sum of gradients:
\begin{align}
    \hat{\theta}^i_{tR+r_1} - \hat{\theta}^i_{tR+r_2} &= \eta \sum_{r=r_1}^{r_2} \nabla \mathcal{L}^i_{tR+r}(\hat{\theta}^i_{tR+r};\mathcal{D}^i_{b,tR+r}), \\
    &\frac{1}{2}\leq r_1<r_2\leq R. \notag
\end{align}
\end{proposition}

Firstly, we train 1 round for the $i$th client and get the loss at iteration $tR+1$, which is bounded by:
\begin{align}
    &\mathcal{L}^i_{tR+1} \overset{(a)}{\leq}  \notag \\
    &\mathcal{L}^i_{tR+\frac{1}{2}} + \langle\nabla\mathcal{L}^i_{tR+\frac{1}{2}}, (\hat{\theta}^i_{tR+1} - \hat{\theta}^i_{tR+\frac{1}{2}})\rangle + \frac{L_1}{2} || \hat{\theta}^i_{tR+1}-\hat{\theta}^i_{tR+\frac{1}{2}}||_2^2 \notag \\
    & \overset{(b)}{=} \mathcal{L}^i_{tR+\frac{1}{2}} - \eta \langle \nabla \mathcal{L}^i_{tR+\frac{1}{2}}, \nabla \mathcal{L}^i_{tR+\frac{1}{2}}(\hat{\theta}^i_{tR+\frac{1}{2}};\mathcal{D}^i_{b,tR+\frac{1}{2}}) \rangle + \notag \\
    & \frac{L_1}{2}||\eta \nabla \mathcal{L}^i_{tR+\frac{1}{2}}(\hat{\theta}^i_{tR+\frac{1}{2}};\mathcal{D}^i_{b,tR+\frac{1}{2}})||_2^2,
\end{align}

where $(a)$ follows Eq. (\!\!~\ref{ass1-eq2}) in Assumption~\ref{ass:app-1}, and $(b)$ follows Proposition~\ref{prop:app-1}. Then we take expectation of both sides of the
above inequality on the random variable of iteration $tR+0$:
\begin{align}
\mathbb{E}[\mathcal{L}^i_{tR+1}] &\leq \mathbb{E}[\mathcal{L}^i_{tR+\frac{1}{2}}] - \eta \mathbb{E}\big[\langle\nabla \mathcal{L}^i_{tR+\frac{1}{2}},\nabla \mathcal{L}^i_{tR+\frac{1}{2}}(\hat{\theta}^i_{tR+\frac{1}{2}};\mathcal{D}^i_{b,tR+\frac{1}{2}})\rangle\big] + \notag \\
&\frac{L_1\eta^2}{2}\mathbb{E}\big[||\nabla \mathcal{L}^i_{tR+\frac{1}{2}}(\hat{\theta}^i_{tR+\frac{1}{2}};\mathcal{D}^i_{b,tR+\frac{1}{2}}||^2_2\big] \notag \\
&\overset{(c)}{=} \mathbb{E}[\mathcal{L}^i_{tR+\frac{1}{2}}] - \eta ||\nabla \mathcal{L}^i_{tR+\frac{1}{2}}||^2_2 + \notag \\
&\frac{L_1\eta^2}{2}\mathbb{E}\big[||\nabla \mathcal{L}^i_{tR+\frac{1}{2}}(\hat{\theta}^i_{tR+\frac{1}{2}};\mathcal{D}^i_{b,tR+\frac{1}{2}}||^2_2\big] \notag \\
& \overset{(d)}{\leq} \mathbb{E}[\mathcal{L}^i_{tR+\frac{1}{2}}] - \eta ||\nabla \mathcal{L}^i_{tR+\frac{1}{2}}||^2_2 + \frac{L_1\eta^2}{2} \big[||\nabla \mathcal{L}^i_{tR+\frac{1}{2}}||_2^2 + \notag \\
&\text{Var}[\nabla \mathcal{L}^i_{tR+\frac{1}{2}}(\hat{\theta}^i_{tR+\frac{1}{2}};\mathcal{D}^i_{b,tR+\frac{1}{2}}]\big] \notag \\
&= \mathbb{E}[\mathcal{L}^i_{tR+\frac{1}{2}}] -(\eta-\frac{L_1\eta^2}{2})||\nabla \mathcal{L}^i_{tR+\frac{1}{2}}||_2^2 + \notag \\ 
&\frac{L_1\eta^2}{2} \text{Var}[\nabla \mathcal{L}^i_{tR+\frac{1}{2}}(\hat{\theta}^i_{tR+\frac{1}{2}};\mathcal{D}^i_{b,tR+\frac{1}{2}})] \notag \\
&\overset{(e)}{\leq} \mathbb{E}[\mathcal{L}^i_{tR+\frac{1}{2}}] -(\eta-\frac{L_1\eta^2}{2})||\nabla \mathcal{L}^i_{tR+\frac{1}{2}}||_2^2 + \frac{L_1\eta^2}{2} \sigma^2.
\end{align}

$(c)$ follows the Eq. (\!\!~\ref{ass3-eq1}) in Assumption~\ref{ass:app-3}.

$(d)$ follows $\text{Var}(x)=\mathbb{E}[x^2]-(\mathbb{E}[x])^2$.

$(e)$ follows the Eq. (\!\!~\ref{ass3-eq2}) in Assumption~\ref{ass:app-3}.

Then, we get Lemma~\ref{app:fl-lem1} by finishing one-iteration local training, which trains $R$ epochs.

\begin{lemma}[Bounded Loss by Local Aggregation]\label{app:fl-lem2}
Based on Assumption~\ref{ass:app-4}, the loss of the local aggregated representation layers of the $i$th client $\hat{f}^i_t= \tau_t f^i_{t-1} + (1-\tau_t) f^{\text{Global}}_t$ is bounded by:
\begin{align}
    \mathbb{E}[\mathcal{L}^i_{tR+0}] &\leq \mathbb{E}[\mathcal{L}^i_{tR}] + \frac{L_1 (\varepsilon^2+\varepsilon)}{2}.
\end{align}
\end{lemma}

\textbf{The proof of Lemma~\ref{app:fl-lem2}}:

Specifically, we reformulate $\hat{f}^i_t= \tau_t f^i_{t-1} + (1-\tau_t) f^{\text{Global}}_{t-1}$ into $\hat{f}^i_{tR+0}= \tau_{tR+0} f^i_{tR} + (1-\tau_{tR+0}) f^{\text{Global}}_{tR}$. Because the aggregated global representations is independent to local model aggregation, this proof omits the changing global representation, assuming that the global representation remains the same.

\begin{equation}
\begin{aligned}
    \mathcal{L}^i_{tR+0} &= \mathcal{L}^i_{tR} + \mathcal{L}^i_{tR+0} - \mathcal{L}^i_{tR} \\
    & \overset{(a)}{=} \mathcal{L}^i_{tR} + \mathcal{L}^i\big(\mathbf{x}^i,\mathbf{y}^i;[\hat{f}^i_{tR+0};g^i_{tR}]\big) - \\
    &\mathcal{L}^i\big(\mathbf{x}^i,\mathbf{y}^i;[f^i_{tR};g^i_{tR}]\big)\\
    & \overset{(b)}{\leq} \mathcal{L}^i_{tR} + \langle \nabla \mathcal{L}^i\big([f^i_{tR};g^i_{tR}]\big), [\hat{f}^i_{tR+0};g^i_{tR}] - [f^i_{tR};g^i_{tR}] \rangle + \\
    & \frac{L_1}{2}||[\hat{f}^i_{tR+0};g^i_{tR}]- [f^i_{tR};g^i_{tR}]||^2_2 \\
    & \overset{(c)}{\leq} \mathcal{L}^i_{tR} + \frac{L_1}{2}||[\hat{f}^i_{tR+0};g^i_{tR}]- [f^i_{tR};g^i_{tR}]||^2_2 \\
    & \overset{(d)}{\leq} \mathcal{L}^i_{tR} + \frac{L_1}{2}||\hat{f}^i_{tR+0}- f^i_{tR}||^2_2 \\
    & \overset{(e)}{=} \mathcal{L}^i_{tR} + \frac{L_1}{2}||(1-\tau_{tR+0})(f^{\text{Global}}_{tR}-f^i_{tR})||_2^2 \\
    & \overset{(f)}{\leq} \mathcal{L}^i_{tR} + \frac{L_1}{2}||1-\tau_{tR+0}||_2^2 ||f^{\text{Global}}_{tR}-f^i_{tR}||_2^2. \\
\end{aligned}
\end{equation}
Then we take the expectation of both sides of the above inequality:
\begin{equation}
\begin{aligned}
    \mathbb{E}[\mathcal{L}^i_{tR+0}] & \leq \mathbb{E}[\mathcal{L}^i_{tR}] + \frac{L_1}{2} \mathbb{E}\big[||1-\tau_{tR+0}||_2^2 ||f^{\text{Global}}_{tR}-f^i_{tR}||_2^2\big] \\
    & \overset{(g)}{=} \mathbb{E}[\mathcal{L}^i_{tR}] + \frac{L_1}{2}\bigg[ \mathbb{E}\big[||1-\tau_{tR+0}||_2^2\big] \mathbb{E}\big[||f^{\text{Global}}_{tR}-f^i_{tR}||_2^2\big]\\ &+\text{Cov}\big(||1-\tau_{tR+0}||_2^2, ||f^{\text{Global}}_{tR}-
    f^i_{tR}||_2^2\big) \bigg] \\
    & \overset{(h)}{\leq}  \mathbb{E}[\mathcal{L}^i_{tR}] + \frac{L_1 \varepsilon^2}{2} \mathbb{E}\big[||1-\tau_{tR+0}||_2^2\big] + \\
    &\frac{L_1}{2}\text{Cov}\big(||1-\tau_{tR+0}||_2^2, ||f^{\text{Global}}_{tR}-f^i_{tR}||_2^2\big)\\
    & \overset{(i)}{\leq} \mathbb{E}[\mathcal{L}^i_{tR}] + \frac{L_1 \varepsilon^2}{2} \mathbb{E}\big[||1-\tau_{tR+0}||_2^2\big] + \\
    &\frac{L_1}{2} \sqrt{\text{Var}(||1-\tau_{tR+0}||_2^2) \text{Var}(||f^{\text{Global}}_{tR}-f^i_{tR}||_2^2)}\\
    & \overset{(j)}{\leq} \mathbb{E}[\mathcal{L}^i_{tR}] + \frac{L_1 \varepsilon^2}{2} \mathbb{E}\big[1-\tau_{tR+0}\big] + \\
    &\frac{L_1}{2} \sqrt{\text{Var}(1-\tau_{tR+0}) \text{Var}(||f^{\text{Global}}_{tR}-f^i_{tR}||_2)}\\
    & \overset{(k)}{\leq} \mathbb{E}[\mathcal{L}^i_{tR}] + \frac{L_1 \varepsilon^2}{2} \mathbb{E}\big[1-\tau_{tR+0}\big] + \\
    &\frac{L_1\varepsilon}{2} \sqrt{\text{Var}(1-\tau_{tR+0})}.\\
    & \overset{(l)}{=} \mathbb{E}[\mathcal{L}^i_{tR}] + \frac{L_1 \varepsilon^2}{2} \mathbb{E}\big[1-e^{-\gamma \mathcal{L}^i_{tR+0}}\big] +\\
    &\frac{L_1\varepsilon}{2} \sqrt{\text{Var}(1-e^{-\gamma \mathcal{L}^i_{tR+0}})}.\\
\end{aligned}
\end{equation}

Suppose $1-e^{-\gamma \mathcal{L}^i_{tR+0}}$ is usually big enough, meaning the loss is not close to 0, then we can not use Taylor series on $1-e^{-\gamma \mathcal{L}^i_{tR+0}}$. Thus we directly give a rough upper bound that $1-e^{-\gamma \mathcal{L}^i_{tR+0}} \leq 1$:
\begin{equation}
\begin{aligned}
    \mathbb{E}[\mathcal{L}^i_{tR+0}] &\leq \mathbb{E}[\mathcal{L}^i_{tR}] + \frac{L_1 (\varepsilon^2+\varepsilon)}{2}.
\end{aligned}
\end{equation}

$(a)$ means extending the loss function with datasets and models.

$(b)$ follows the Eq. (\!\!~\ref{ass1-eq2}) in Assumption~\ref{ass:app-1}.

$(c)$ follows the case: generally, the local gradient direction is reverse to the global update direction. Suppose $\theta^{\text{Global}}$ is the center of $\{\theta^i\}_{i=1}^N$, and $\{\mathcal{D}^i\}_{i=1}^N$ are fairly heterogeneous. Local personalized training moves $\theta^i$ far away from the center, the global aggregation drag $\theta^i$ to approach $\theta^{\text{Global}}$. Thus, $\langle \nabla \mathcal{L}^i\big([f^i_{tR};g^i_{tR}]\big), [\hat{f}^i_{tR+0};g^i_{tR}] - [f^i_{tR};g^i_{tR}] \rangle$ is usually negative.

$(d)$ follows the case: By regarding $[\hat{f}^i_{tR+0};g^i_{tR}]$ and $[f^i_{tR};g^i_{tR}]$
as matrices, we can denote $[\hat{f}^i_{tR+0};g^i_{tR}] - [f^i_{tR};g^i_{tR}]$ as $[\hat{f}^i_{tR+0} - f^i_{tR}; \mathbf{0}]$. Thus, we drop $\mathbf{0}$.

$(e)$ follows the Eq. (\!\!~\ref{Eq9}) that $\hat{f}^i_{tR+0} = \tau_{tR+0} f^i_{tR} + (1-\tau_{tR+0}) f^{\text{Global}}_{tR}$.

$(f)$ follows Cauchy-Schwarz inequality that $||\mathbf{a}\mathbf{b}||_2 \leq ||\mathbf{a}||_2||\mathbf{b}||_2$.

$(g)$ follows $\mathbb{E}[ab] = \mathbb{E}[a]\mathbb{E}[b]+\text{Cov}(a,b)$ due to the case: without loss of generality, we suppose $||1-\tau_{tR+0}||^2_2$ is not independent to $||f^{\text{Global}}_{tR}-f^i_{tR}||_2^2$.

$(h)$ follows Assumption~\ref{ass:app-4}.

$(i)$ follows Cauchy-Schwarz inequality that $\text{Cov}(a,b) \leq \sqrt{\text{Var}(a)\text{Var}(b)}$.

$(j)$ follows the case that $1-\tau_{tR+0} \in [0,1]$ and $||f^{\text{Global}}_{tR}-f^i_{tR}||_2 \in [0,1]$.

$(k)$ follows Assumption~\ref{ass:app-4}.

$(l)$ expands $\tau_{tR+0}$ following the Eq. (\!\!~\ref{Eq8}).

\begin{lemma}[Bounded Loss by Global Aggregated Representations]\label{app:fl-lem3}
The loss of utilizing global representations are different, e.g., $tR+0$ and $tR+\frac{1}{2}$, and bounded by:
\begin{align}
    \mathbb{E}[\mathcal{L}^i_{tR+\frac{1}{2}}] &\leq \mathbb{E}[\mathcal{L}^i_{tR+0}] + \frac{2\alpha}{\tau_{\text{CL}}}.
\end{align}

\end{lemma}

\textbf{The proof of Lemma~\ref{app:fl-lem3}}:

Referring to Theorem~\ref{app:theo-loss-1}, we have $\mathbb{E}_{\omega^i_j\sim \Omega^i_{tR+0}}$. To be brief, we denote $\frac{e^{\big[D(p^+_{j,t+1})/\tau_{\text{CL}}\big]}}{e^{\big[D(p^+_{j,t+1})/\tau_{\text{CL}}\big]} + \sum_{\hat{c} \in \mathcal{C}\backslash c}e^{\big[D(p^-_{j,t+1})/\tau_{\text{CL}}\big]}}$ in Eq. (\!\!~\ref{Eq11}) as $\tilde{h}(\bar{\omega}^{\text{Global}}_{t+1}, \omega^i_j)$, $D(p^+_{j,t+1})/\tau_{\text{CL}}$ as $D(\bar{\omega}^{\text{Global}}_{t+1}, \omega^i_j)$, and $\mathbb{E}_{\omega^i_j\sim \Omega^i_{tR+0}}$ as $\mathbb{E}$. Similar to Lemma~\ref{app:fl-lem2}, this proof omits the change brought by local aggregation, assuming that the local model remains the same.

\begin{equation}
\begin{aligned}
    \mathcal{L}^i_{tR+\frac{1}{2}} &= \mathcal{L}^i_{tR+0} + \mathcal{L}^i_{tR+\frac{1}{2}} - \mathcal{L}^i_{tR+0} \\
    & \overset{(a)}{=} \mathcal{L}^i_{tR+0} + \alpha \mathbb{E} \big[-\log \tilde{h}(\bar{\omega}^{\text{Global}}_{c,tR+\frac{1}{2}},\omega^i_j) + \\
    &\log \tilde{h}(\bar{\omega}^{\text{Global}}_{c,tR+0},\omega^i_j) \big] \\
    & = \mathcal{L}^i_{tR+0} + \alpha \mathbb{E} \bigg[ \log \frac{\tilde{h}(\bar{\omega}^{\text{Global}}_{c,tR+0},\omega^i_j)}{\tilde{h}(\bar{\omega}^{\text{Global}}_{c,tR+\frac{1}{2}},\omega^i_j)}\bigg]\\
    & \overset{(b)}{=} \mathcal{L}^i_{tR+0} + \alpha \mathbb{E} \bigg[ \log \frac{e^{[D(\bar{\omega}^{\text{Global}}_{tR+\frac{1}{2}}, \omega^i_j)/\tau_{\text{CL}}]}}{e^{[D(\bar{\omega}^{\text{Global}}_{tR+0}, \omega^i_j)/\tau_{\text{CL}}]}} \bigg] \\
    & = \mathcal{L}^i_{tR+0} + \frac{\alpha}{\tau_{\text{CL}}}\mathbb{E} \big[D(\bar{\omega}^{\text{Global}}_{tR+\frac{1}{2}}, \omega^i_j) - D(\bar{\omega}^{\text{Global}}_{tR+0}, \omega^i_j)\big] \\
    & \overset{(c)}{\leq} \mathcal{L}^i_{tR+0} + \frac{2\alpha}{\tau_{\text{CL}}}
\end{aligned}
\end{equation}

Then we take the expectation of both sides of the above inequality:
\begin{align}
    \mathbb{E}[\mathcal{L}^i_{tR+\frac{1}{2}}] \leq \mathbb{E} [\mathcal{L}^i_{tR+0}] + \frac{2\alpha}{\tau_{\text{CL}}}.
\end{align}

$(a)$ follows the Eq. (\!\!~\ref{Eq12}) and drops $\mathcal{L}^i_{\text{Per}}$ since $\omega^i_j$ and $\hat{\theta}$ are both unchanged. We combine the 2 expectation terms because the sampling space they operate on is the same, since the local model is unchanged at iteration $tR+0$ and $tR+\frac{1}{2}$.

$(b)$ expands $\tilde{h}(\bar{\omega}^{\text{Global}}_{c,tR+\frac{1}{2}},\omega^i_j)$ and $\tilde{h}(\bar{\omega}^{\text{Global}}_{c,tR+0},\omega^i_j)$. Similar to $(a)$, we omits the denominator.

$(c)$ follows the range that $D(\cdot) \in [-1,1]$ in Eq. (\!\!~\ref{Eq12}).

\subsection{Theorems}

\begin{theorem}[One-Iteration Deviation]\label{app:FL-theo1}
Based on Lemma~\ref{app:fl-lem1},~\ref{app:fl-lem2}, and~\ref{app:fl-lem3}, for the $i$th client, after the iteration $t$, we have:
\begin{align}
    \mathbb{E}[\mathcal{L}^i_{(t+1)R+\frac{1}{2}}] &\leq \mathbb{E}[\mathcal{L}^i_{tR+\frac{1}{2}}] - \big(\eta-\frac{L_1\eta^2}{2}\big)\sum_{r=\frac{1}{2}}^R||\nabla\mathcal{L}^i_{tR+r}||_2^2 \notag \\
    &+ \frac{R L_1\eta^2 \sigma^2}{2} + \frac{L_1(\varepsilon^2 + \varepsilon)}{2}+\frac{2\alpha}{\tau_{\text{CL}}}. \label{eq: one iteration dev}
\end{align}    
\end{theorem}

\begin{corollary}[Non-Convex FedCoSR Convergence]\label{app:FL-colo1}
The loss function of the $i$th client monotonously decreases between iteration $t$ and iteration $t+1$ when:
\begin{align}
     \eta_{r'} < \frac{\mathbb{S} + \sqrt{\big[\mathbb{S} - \frac{(L_1 \mathbb{S} + RL_1\sigma^2)(L_1\varepsilon^2\tau_{\text{CL}} + L_1 \varepsilon\tau_{\text{CL}}+4\alpha)}{\tau_{\text{CL}}}} }{L_1 \mathbb{S} + RL_1\sigma^2},
\end{align}
where $r'=\frac{1}{2}, 1, ..., R$, and briefly, $\mathbb{S}=\sum^{r'}_{r=\frac{1}{2}}||\nabla\mathcal{L}^i_{tR+r}||^2_2$. Thus FedCoSR converges.
\end{corollary}

\textbf{The proof of Corollary~\ref{app:FL-colo1}}:

Briefly, we denote $\sum_{r=\frac{1}{2}}^R||\nabla\mathcal{L}^i_{tR+r}||_2^2$ as $\bar{\mathbb{S}}$. To make sure $- \big(\eta-\frac{L_1\eta^2}{2}\big) \bar{\mathbb{S}} + \frac{R L_1\eta^2 \sigma^2}{2} + \frac{L_1(\varepsilon^2 + \varepsilon)}{2}+\frac{2\alpha}{\tau_{\text{CL}}} \leq 0$, we get:
\begin{align}
    \eta &< \frac{\bar{\mathbb{S}} + \sqrt{\big[\bar{\mathbb{S}} - \frac{(L_1 \bar{\mathbb{S}} + RL_1\sigma^2)(L_1\varepsilon^2\tau_{\text{CL}} + L_1 \varepsilon\tau_{\text{CL}}+4\alpha)}{\tau_{\text{CL}}}} }{L_1 \bar{\mathbb{S}} + RL_1\sigma^2}, \\
    \alpha &< \frac{\tau_{\text{CL}}}{4} \bigg[ 
\frac{\big[\bar{\mathbb{S}}\big]^2}{L_1\bar{\mathbb{S}}+ RL_1\sigma^2} - L_1\varepsilon^2 - L_1\varepsilon \bigg].
\end{align}
For generality, we choose an arbitrary round $r'$ to replace the total rounds $R$, where $r'=\frac{1}{2}, 1, ..., R$. Besides, there is no learning rate decay in FedCoSR, thus the learning rate $\alpha$ at iteration $t$ is fixed. Briefly, we denote $\sum^{r'}_{r=\frac{1}{2}} ||\nabla \mathcal{L}^i_{tR+r}||^2_2$ as $\mathbb{S}'$. Then we reformulate them into:
\begin{align}
    \eta_{r'} &< \frac{\mathbb{S}' + \sqrt{\big[\mathbb{S}' \big]^2 - \frac{(L_1 \mathbb{S}' + RL_1\sigma^2)(L_1\varepsilon^2\tau_{\text{CL}} + L_1 \varepsilon\tau_{\text{CL}}+4\alpha)}{\tau_{\text{CL}}}} }{L_1 \mathbb{S}' + RL_1\sigma^2}, \\
    \alpha_t &< \frac{\tau_{\text{CL}}}{4} \bigg[ 
\frac{[\mathbb{S}']^2}{L_1 \mathbb{S}' + RL_1\sigma^2} - L_1\varepsilon^2 - L_1\varepsilon \bigg].
\end{align}

\begin{theorem}[Non-Convex Convergence Rate of FedCoSR]\label{app:FL-theo2}
Given any $\epsilon > 0$. After $T$ iterations, the $i$th client converges with:
\begin{align}
    \frac{1}{TR} \sum^T_{t=1}\sum^R_{r=\frac{1}{2}} \mathbb{E}[||\nabla \mathcal{L}^i_{tR+r}||^2_2]< \epsilon,
\end{align}
when
\begin{align}
    T &> \frac{2\tau_{\text{CL}} (\mathcal{L}^i_\frac{1}{2} - \mathcal{L}^{i,*}) }{(2\eta-L_1\eta^2)\epsilon\tau_{\text{CL}}R - L_1\eta^2\sigma^2\tau_{\text{CL}}R - \tau_{\text{CL}}L_1(\varepsilon^2-\varepsilon) - 4\alpha}, \\
    \eta &< \frac{\epsilon}{L_1(\epsilon+\sigma^2)} + \notag\\ &\frac{\sqrt{4\epsilon^2\tau_{\text{CL}}^2R^2 - 4 L_1\tau_{\text{CL}}R(\epsilon+\sigma^2)[\tau_{\text{CL}}L_1(\varepsilon^2+\varepsilon)+4\alpha]}}{2L_1\tau_{\text{CL}}R(\epsilon+\sigma^2)}, \\
    \alpha &< \frac{\epsilon^2}{4L_1(\epsilon+\sigma^2)} - \frac{\tau_{\text{CL}}L_1}{4}(\varepsilon^2+\varepsilon).
\end{align}
\end{theorem}

\textbf{The proof of Theorem~\ref{app:FL-theo2}}:

Take expectation on both sides of Eq. (\!\!~\ref{eq: one iteration dev}) in the view of $T$ and $R$, then we have:
\begin{figure*}[hbpt]
\begin{align}
    \frac{1}{TR}\sum^T_{t=1}\sum^R_{r=\frac{1}{2}} \mathbb{E}[||\nabla \mathcal{L}^i_{tR+r}||^2_2] \leq \frac{\frac{1}{TR}\sum^T_{t=1} \bigg[\mathbb{E}\big[\mathcal{L}^i_{tR+\frac{1}{2}}\big] - \mathbb{E}\big[\mathcal{L}^i_{(t+1)R+\frac{1}{2}}\big]\bigg] +\frac{L_1\eta^2\sigma^2}{2} + \frac{L_1(\varepsilon^2+\varepsilon)}{2R} + \frac{2\alpha}{\tau_{\text{CL}}R} }{\eta-\frac{L_1\eta^2}{2}}.
\end{align}
\end{figure*}

Given any $\epsilon>0$, let

\begin{figure*}[hbpt]
\begin{align}
    \frac{\frac{1}{TR}\sum^T_{t=1} \bigg[\mathbb{E}\big[\mathcal{L}^i_{tR+\frac{1}{2}}\big] - \mathbb{E}\big[\mathcal{L}^i_{(t+1)R+\frac{1}{2}}\big]\bigg] +\frac{L_1\eta^2\sigma^2}{2} + \frac{L_1(\varepsilon^2+\varepsilon)}{2R} + \frac{2\alpha}{\tau_{\text{CL}}R} }{\eta-\frac{L_1\eta^2}{2}} < \epsilon,
\end{align}
\end{figure*}

Suppose the final loss is optimal, denoted as $\mathcal{L}^{i,*}$. Besides, we denote the initial loss whose $t=1$ and $r=\frac{1}{2}$ as $\mathcal{L}^i_\frac{1}{2}$. Since $\sum^T_{t=1} \bigg[\mathbb{E}\big[\mathcal{L}^i_{tR+\frac{1}{2}}\big] - \mathbb{E}\big[\mathcal{L}^i_{(t+1)R+\frac{1}{2}}\big]\bigg] \leq \mathcal{L}^i_\frac{1}{2} - \mathcal{L}^{i,*}$, we suppose the above inequality still holds when we replace $\sum^T_{t=1} \bigg[\mathbb{E}\big[\mathcal{L}^i_{tR+\frac{1}{2}}\big] - \mathbb{E}\big[\mathcal{L}^i_{(t+1)R+\frac{1}{2}}\big]\bigg]$ with $\mathcal{L}^i_\frac{1}{2} - \mathcal{L}^{i,*}$. Then we have:
\begin{align}
    \frac{1}{TR} \sum^T_{t=1}\sum^R_{r=\frac{1}{2}} \mathbb{E}[||\nabla \mathcal{L}^i_{tR+r}||^2_2]< \epsilon,
\end{align}
when
\begin{align}
    T &> \frac{2\tau_{\text{CL}} (\mathcal{L}^i_\frac{1}{2} - \mathcal{L}^{i,*}) }{(2\eta-L_1\eta^2)\epsilon\tau_{\text{CL}}R - L_1\eta^2\sigma^2\tau_{\text{CL}}R - \tau_{\text{CL}}L_1(\varepsilon^2-\varepsilon) - 4\alpha}, \\
    \eta &< \frac{\epsilon}{L_1(\epsilon+\sigma^2)} + \notag \\
    &\frac{\sqrt{4\epsilon^2\tau_{\text{CL}}^2R^2 - 4 L_1\tau_{\text{CL}}R(\epsilon+\sigma^2)[\tau_{\text{CL}}L_1(\varepsilon^2+\varepsilon)+4\alpha]}}{2L_1\tau_{\text{CL}}R(\epsilon+\sigma^2)}, \\
    \alpha &< \frac{\epsilon^2}{4L_1(\epsilon+\sigma^2)} - \frac{\tau_{\text{CL}}L_1}{4}(\varepsilon^2+\varepsilon).
\end{align}

\section{Optimal Hyperparameters}
Table~\ref{tab:ourhyper} shows the details of hyperparameters used in FedCoSR. By default, for local clients, we set $\beta=0.1$ and $N=20$. In local training, we use ADAM as the optimizer. 

\begin{table}[htb]
\centering
\footnotesize
\setlength{\tabcolsep}{4pt}
\caption{Hyperparameters of FedCoSR.}
\label{tab:ourhyper}
\vspace{-5pt}
\begin{tabular}{ccc}
\hline\hline
\textbf{Hyperparameter} & \textbf{Value}      & \textbf{Note}                                             \\ \hline
$\eta$               & 0.003                 & Local learning rate                                       \\
$b$                     & 16                   & Local batch size             \\
$r$                     & 1                   &  Local training epoch             \\
$k$    &   128    &   Dimension of shared representations \\
$\beta_{\text{joint}}$                     & 1                   &  Client joint ratio             \\
$\beta_{\text{dropout}}$                     & 0.3                   &  Dropout rate             \\
$\alpha$                   & 1                   &  Trade-off factor of loss            \\
$\tau_{\text{CL}}$   & 0.1                   &  Temperature of CRL            \\
$\gamma$   & 0.8                   &  Scaling of loss-wise weighting            \\
\hline\hline
\end{tabular}
\vspace{-5pt}
\end{table}

Table~\ref{tab:otherhyper} shows key hyperparameters of the compared methods. By default, if not mentioned in Table~\ref{tab:otherhyper}, then the local learning rate $\eta$ is set to 0.01, local batch size $b$ is set to 16, local learning epoch $r$ is set to $1$, client joint ratio $\beta_{\text{joint}}$ is set to 1 for 20-client task and 0.5 for 100-client task.

\begin{table*}[hbpt]
\centering
\footnotesize
\setlength{\tabcolsep}{11pt}
\caption{Key hyperparameters of compared methods.}
\label{tab:otherhyper}
\begin{tabular}{cccl}
\hline\hline
\textbf{Methods} & \textbf{Hyperparameter} & \textbf{Value}      & \textbf{Note}                                             \\ \hline
Local         & $\eta$    & 0.003   & Local learning rate     \\ \hline

FedAvg         & -                  & -                 & -   \\ \hline

FedProx        & $\mu$                     & 0.01                  &  Regularization factor \\ \hline

\multicolumn{1}{c}{\multirow{2}{*}{pFedMe}}  & $\lambda$                     & 10                  &  Regularization factor \\
\multicolumn{1}{l}{}          & $r$                     & 5                  &  Local training epoch \\ \hline

\multicolumn{1}{c}{\multirow{2}{*}{Ditto}}  & $\lambda$                     & 1               &  Regularization factor \\
\multicolumn{1}{l}{}          & $r$                     & 2                  &  Local training epoch \\ \hline

\multicolumn{1}{c}{\multirow{3}{*}{PerFedAvg}} &   $\alpha$     &  0.2, 0.05, 0.01    &    Steps size of meta learning for CIFAR-10, EMNIST, CIFAR-100   \\ 
&   $r_{\text{Meta}}$     &  5, 10, 20    &    Meta learning steps for CIFAR-10, EMNIST, CIFAR-100 \\
&   $b$     &  32    &  Meta learning batch size  \\

\multicolumn{1}{c}{\multirow{4}{*}{PeFLL}} & $\lambda_h$ & 0.001 & Regularization parameter for the embedding network \\
& $\lambda_v$ & 0.001 & Regularization parameter for the hyper-network \\
& $\lambda_{\theta}$ & 0 & Output regularization \\
& $b$ & 32 & Batch size \\
\hline

\multicolumn{1}{c}{\multirow{2}{*}{FedRep}}  & $\lambda$                     & 1                  &  Regularization factor \\
\multicolumn{1}{l}{}          & $r$                     & 5                  &  Local training epochs \\ \hline     

\multicolumn{1}{c}{\multirow{2}{*}{LG-FedAvg}}   & $\beta_{\text{joint}}$                     & 0.1                  &  Client joint ratio \\
\multicolumn{1}{l}{}          & $r$                     & 5                  &  Local training epochs \\ \hline     

FedPer         & -   & -    & -   \\ \hline

\multicolumn{1}{c}{\multirow{2}{*}{MOON}}  & $\tau_{\text{CL}}$     & 1   & Temperature of contrastive learning  \\     
\multicolumn{1}{l}{}           & $\mu$               & 0.01     & Regularization factor    \\

\multicolumn{1}{c}{\multirow{4}{*}{FedRCL}}  & $\eta$     & 0.1   & Learning rate  \\     
       & $\lambda$               & 0.7     & Threshold for the close set   \\ 
      & $\beta$               & 1     & Weight of the divergence term    \\ 
      & $\tau$               & 0.05     & Contrastive learning temperature \\ 

\hline

\multicolumn{1}{c}{\multirow{3}{*}{FedAMP}}           & $\alpha_K$        & 10000   & Discount of learning rate, decrease by 0.1 per 30 rounds  \\
\multicolumn{1}{l}{}           & $\sigma$       & 1     & Scaling hyperparameter of the attention-inducing function     \\
\multicolumn{1}{l}{}           & $\lambda$               & 1        & Regularization factor     \\ \hline

\multicolumn{1}{c}{\multirow{3}{*}{FedALA}}         & $\eta$     & 0.1    & Local learning rate   \\
\multicolumn{1}{l}{}           & $\lambda$               & 1       & Regularization factor             \\
\multicolumn{1}{l}{}           & $p$     & 1       & The last $p$ layers participating in ALA                  \\

\multicolumn{1}{c}{\multirow{2}{*}{FedAS}}         & $\eta$     & 5e-3    & Local learning rate   \\
& $E_l$     & 5    & Local training epochs   \\

\hline

FedProto    & $\lambda$    & 0.1      & Regularization factor   \\  \hline

FedPAC    & $\lambda$    & 1       & Regularization factor            \\  \hline

FedGH  & $\eta_{\theta}$    & 0.01       & Learning rate for the global projection layer            \\ 

\multicolumn{1}{c}{\multirow{2}{*}{pFedFDA}}         & $k$     & 2    & cross-validation folds   \\
& $E$     & 5    & Local training epochs   \\

\hline\hline
\end{tabular}
\end{table*}

\section{Network Architecture}
The network structure used by all methods experimented in this paper is shown in Table~\ref{tab: app_network}. It consists of 2 convolutional layers followed by a fully connected layer. Each convolutional layer includes a convolution operation, a ReLU activation, and a max-pooling step. The first convolutional layer applies 32 filters of size $5\times5$ with a stride of 1, followed by ReLU activation and a $2\times2$ max pooling with a stride of 2. The second convolutional layer similarly applies 64 filters of size $5\times5$ with a stride of 1, followed by ReLU activation and $2\times2$ max pooling with a stride of 2. The output of the second convolutional layer is flattened and passed through a fully connected layer with 1024 input features and $k$ output features, followed by a ReLU activation, where $k$ is the dimension of shared representations. Finally, the output is passed through another fully connected layer that maps the $k$ input features to $C$ output features, where $C$ is the total number of label classes of the specific dataset.

In FedCoSR, Layer 1-3 consists of the representation layers $f$, and the last FC layer is the projection layer $g$.

\begin{table*}[hbpt]
\centering
\footnotesize
\setlength{\tabcolsep}{5pt}
\caption{The model architecture for each client in all FL methods. FC refers to fully connected layer, Conv2D refers to the 2D convolutional layer, ReLU refers to the activation function, and MaxPool2D refers to the 2D max pooling layer. $k$ refers to the dimension of shared representations, and $C$ is the total amount of label classes of the specific dataset.}
\label{tab: app_network}
\begin{tabular}{cc}
\hline\hline
\textbf{Layer} & \textbf{Details} \\ \hline
1    &  Conv2d(Input channels: 1, kernel: $5\times5$, stride: $1\times1$, output: 32) + ReLU + MaxPool2D(Kernel: $2\times2$, stride: $2\times2$)  \\
2    & Conv2d(Input channels: 32, kernel: $5\times5$, stride: $1\times1$, output: 64) + ReLU + MaxPool2D(Kernel: $2\times2$, stride: $2\times2$)  \\
3    & FC(1024, $k$) + ReLU \\
4    & FC($k$, $C$) \\
\hline\hline
\end{tabular}
\end{table*}

\section{Visualization of Data Distributions}

\begin{figure*}[hbpt]
    \centering   
    \subfigure{\label{Fig:dist_scala_prac_cifar10}
    \includegraphics[width=.7\textwidth]{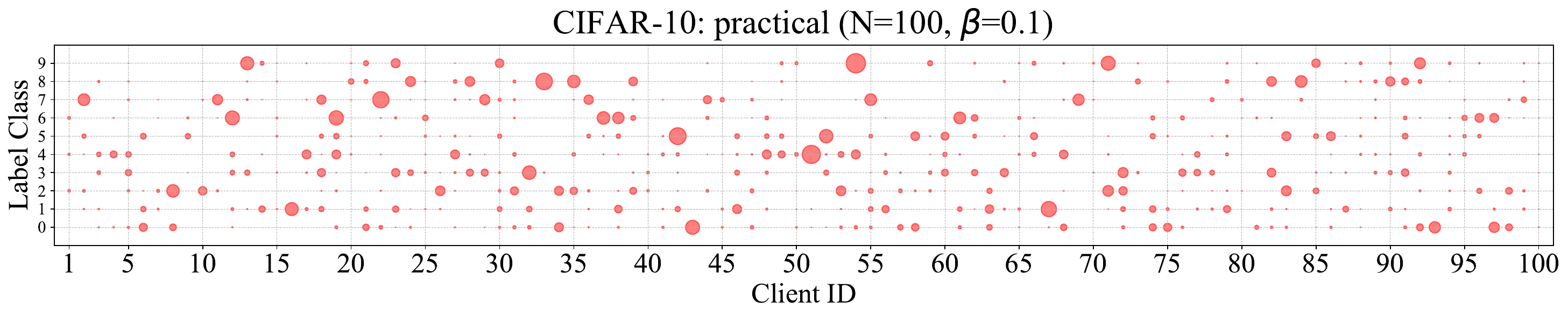}  }
    \subfigure{\label{Fig:dist_scala_path_cifar10}
    \includegraphics[width=.7\textwidth]{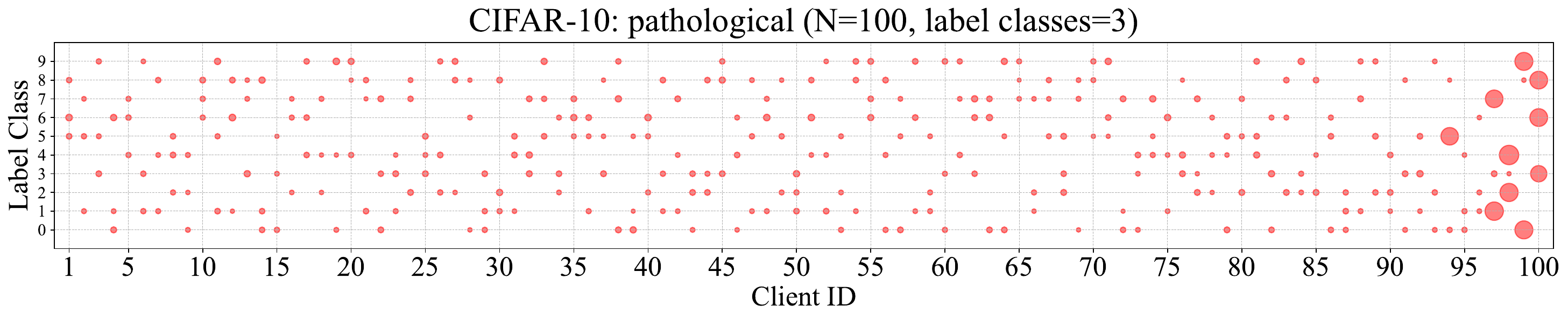}  }
    \captionsetup{width=.99\linewidth}
    \caption{Data distribution of CIFAR-10 for scalability evaluation.}
    \label{Fig:dist_cifar10_scala}
\end{figure*}

Fig.~\ref{Fig:dist_cifar10_scala} is the data distribution of CIFAR-10 for the scalability experiment.

\begin{figure*}[!t]
    \centering   
    \subfigure{\label{Fig:dist_hete001_prac_cifar10}
    \includegraphics[width=.35\textwidth]{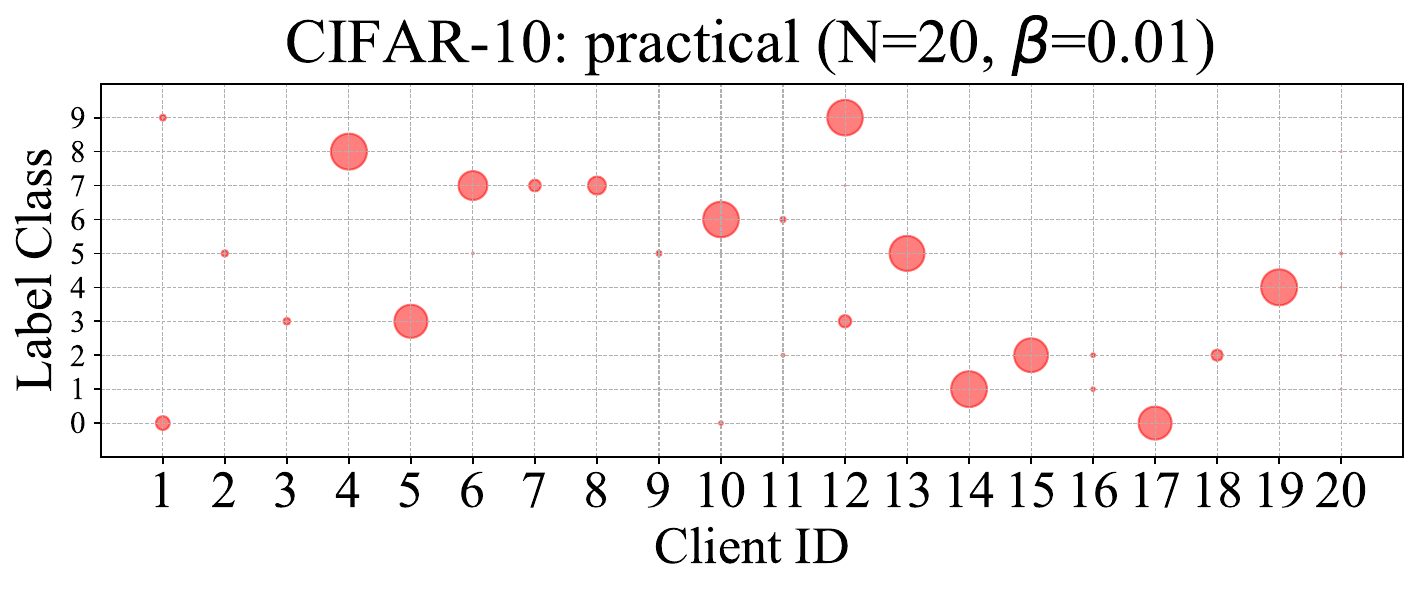}  }
    \subfigure{\label{Fig:dist_hehe1_prac_cifar10}
    \includegraphics[width=.35\textwidth]{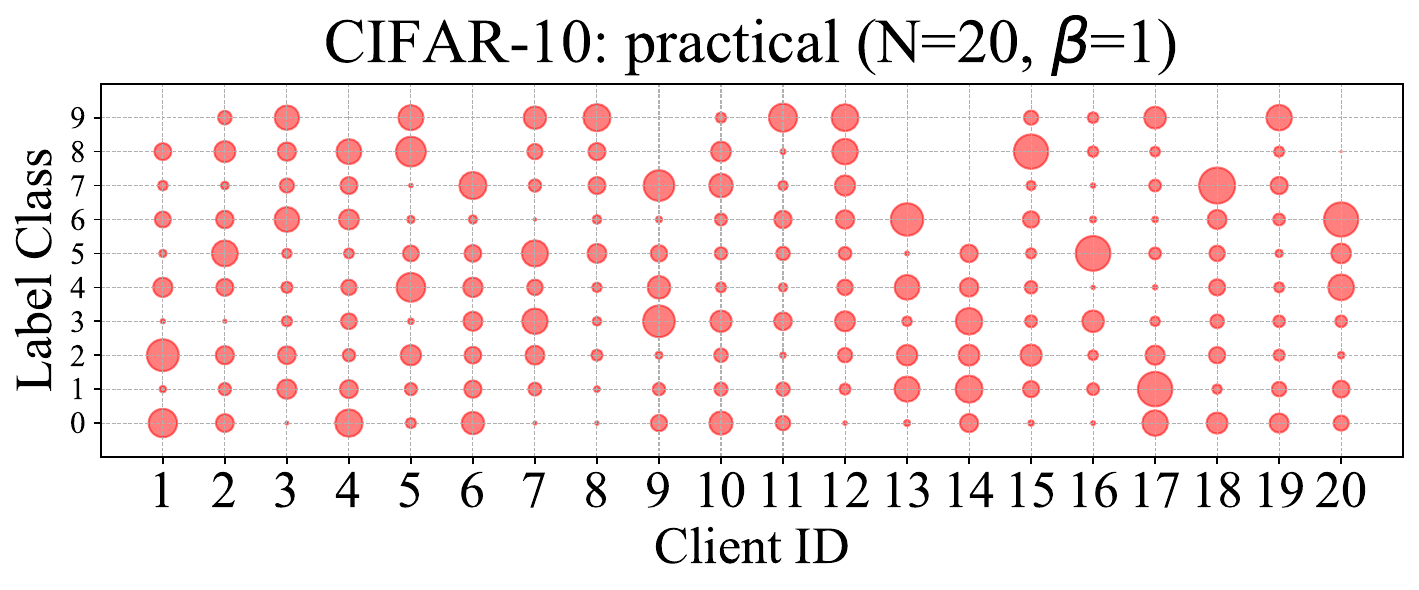}  }
    \captionsetup{width=.99\linewidth}
    \caption{Data distribution of CIFAR-10 for practical heterogeneity evaluation.}
    \label{Fig:dist_cifar10_hete}
\end{figure*}

Fig.~\ref{Fig:dist_cifar10_hete} is the data distribution of CIFAR-10 for the varying practical heterogeneity experiment.

\begin{figure*}[!t]
    \centering   
    \subfigure{\label{Fig:dist_hete10_path_cifar100}
    \includegraphics[width=.35\textwidth]{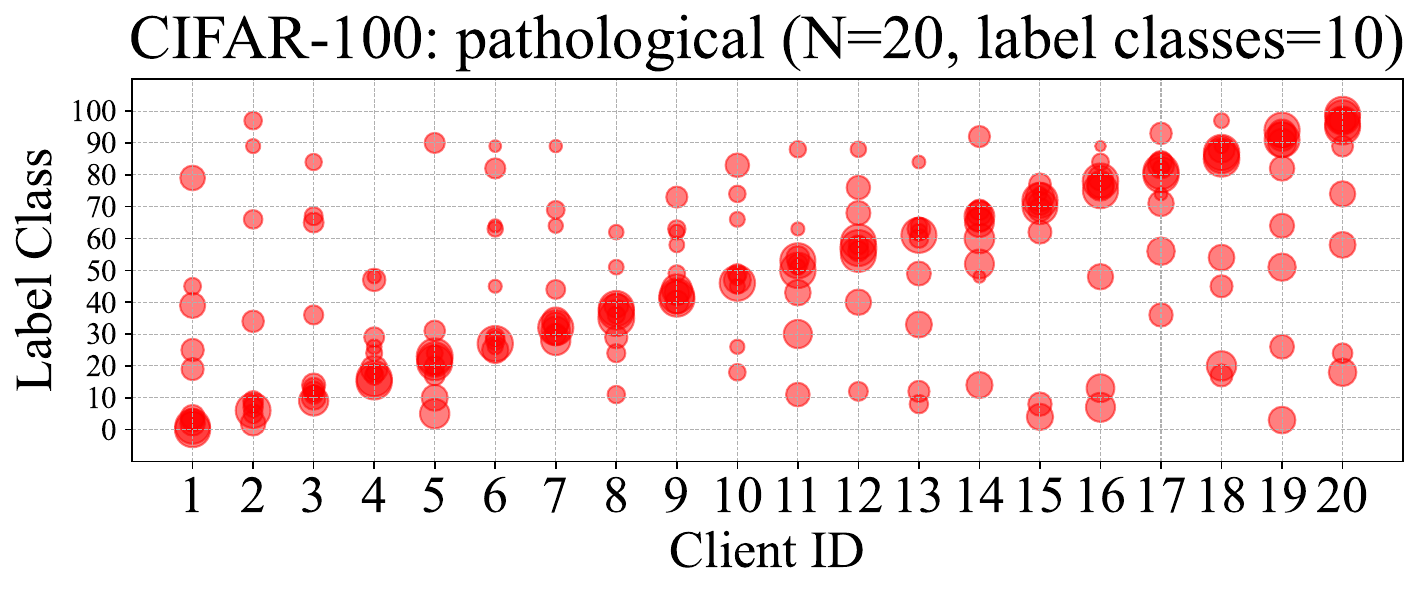}  }
    \subfigure{\label{Fig:dist_hehe50_path_cifar100}
    \includegraphics[width=.35\textwidth]{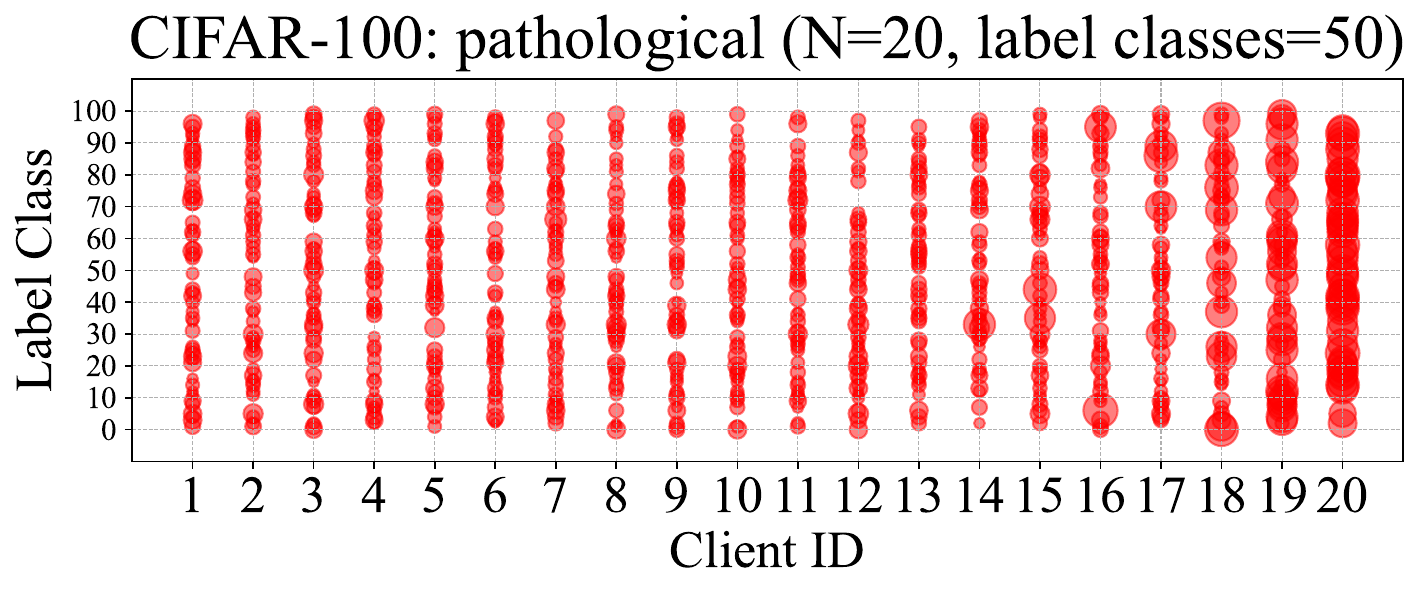}  }
    \captionsetup{width=.99\linewidth}
    \caption{Data distribution of CIFAR-100 for pathological heterogeneity evaluation.}
    \label{Fig:dist_cifar100_hete}
\end{figure*}

Fig.~\ref{Fig:dist_cifar100_hete} is the data distribution of CIFAR-100 for the varying pathological heterogeneity experiment.

\section{Hyperparameter Sensitivity}\label{app:abla}

\begin{figure}[!t]
    \centering   
    \subfigure{
    \includegraphics[width=.22\textwidth]{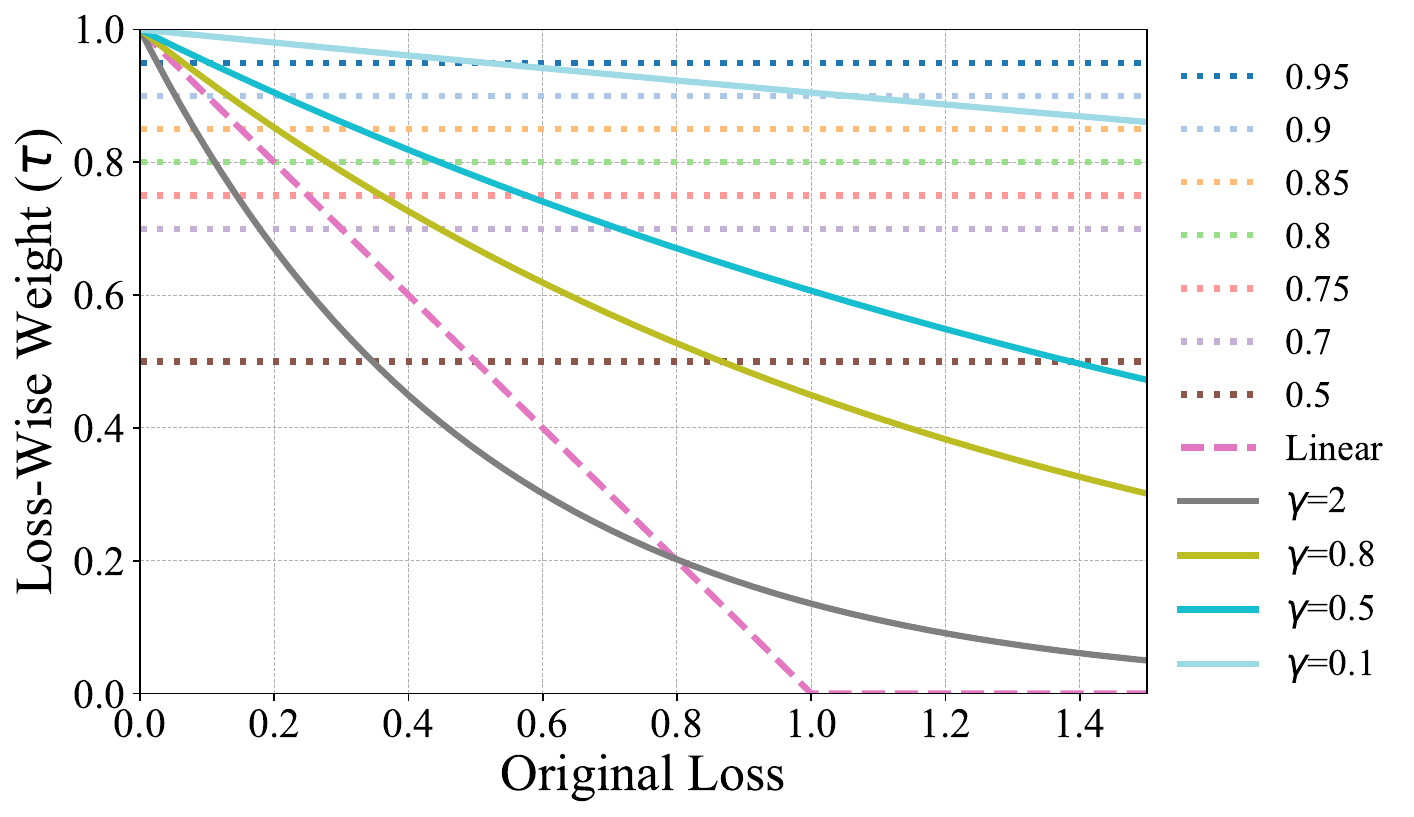}  }
    \subfigure{
    \includegraphics[width=.22\textwidth]{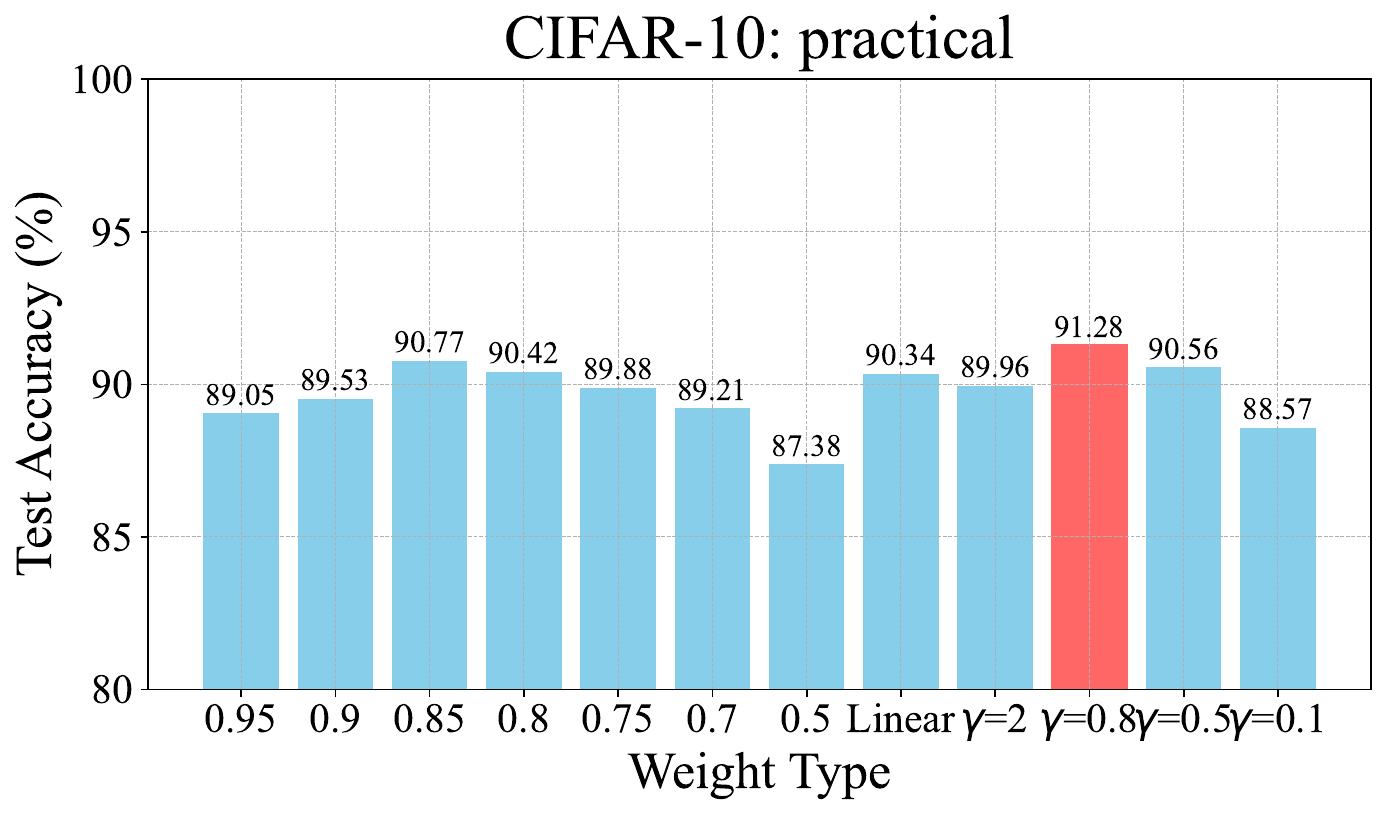}  }
    \captionsetup{width=.99\linewidth}
    \caption{The accuracy of FedCoSR with varying $\tau$, where $\gamma$ is the scaling factor of $\mathcal{L}_{\text{Reg}}$ in Eq. (\!\!~\ref{Eq8}).}
    \vspace{-10pt}
    \label{Fig:abla_weight}
\end{figure}

We study the effect of $\tau$ in the local aggregation by setting $\tau$ as fixed values or linear mathematical form which is $\tau=\max(0, 1-\mathcal{L})$. Also, we adjust $\gamma$ in Eq. (\!\!~\ref{Eq8}). Fig.~\ref{Fig:abla_weight} shows the trends of $\tau$ under varying loss values, along with the corresponding results. We can see that FedCoSR demonstrates superiority of the calculation of $\tau$ among fixed values. Since FL communication is a dynamic process, dynamic calculation of $\tau$ for each iteration is more effective. Besides, compared with ``linear'' and $\gamma=2$, $\gamma=0.8$ demonstrates better performance, intuitively indicating that in FedCoSR, the loss-wise weight should approach 1 faster to be quickly independent to the global model. 
Consequently, loss-wise weighting is not always the most effective approach, but it generally outperforms fixed values and doesn't heavily depend on tuning the hyperparameter $\gamma$.

\begin{figure}[th]
    \centering   
    \subfigure{\label{Fig:origin_}
    \includegraphics[width=.15\textwidth]{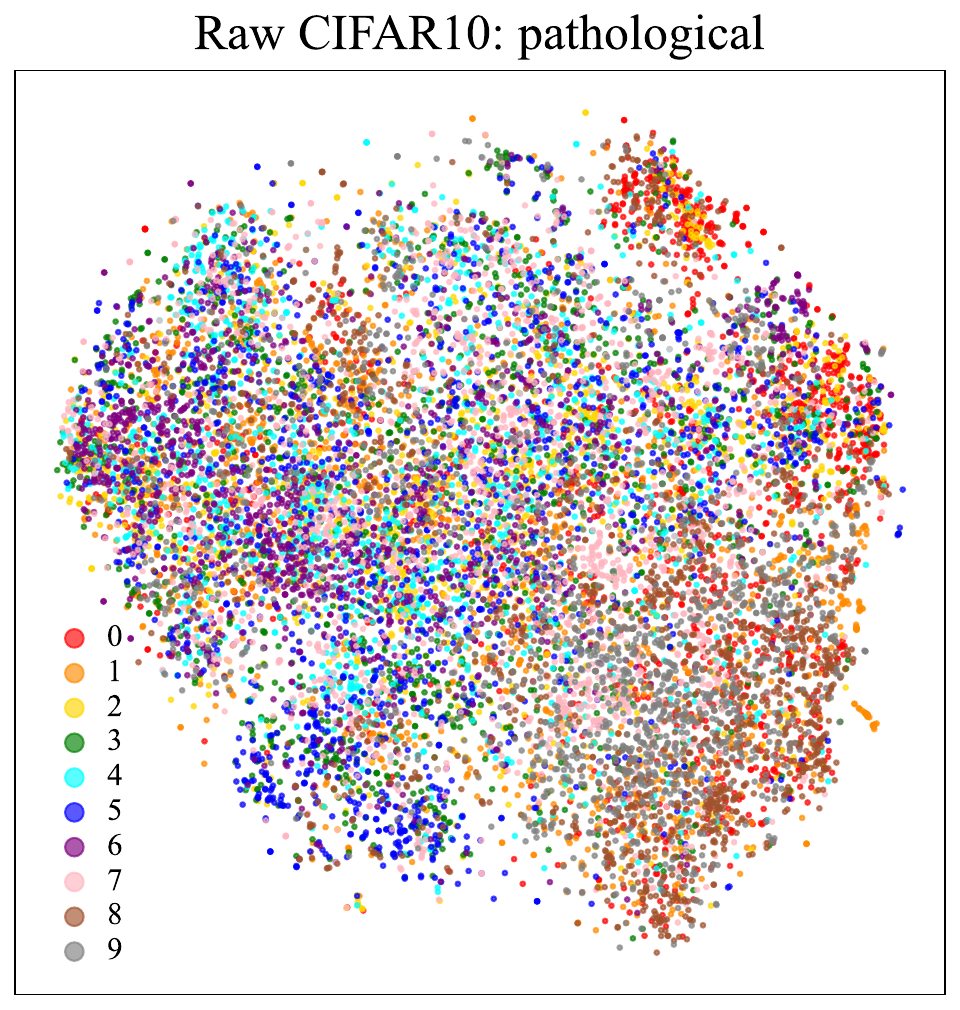}  }
    \subfigure{\label{Fig:rep_avg_}
    \hspace{-8pt}\includegraphics[width=.15\textwidth]{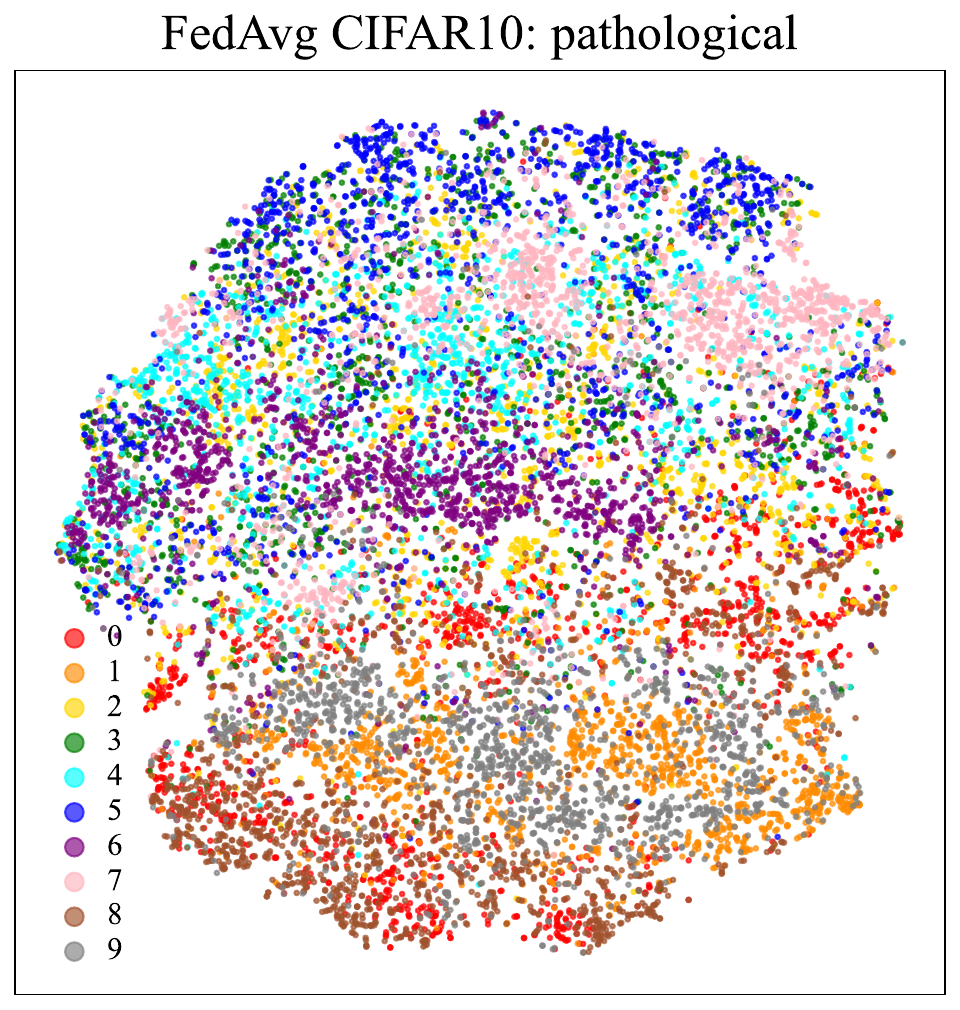}  }
    \subfigure{\label{Fig:rep_lg_}
    \hspace{-8pt}\includegraphics[width=.15\textwidth]{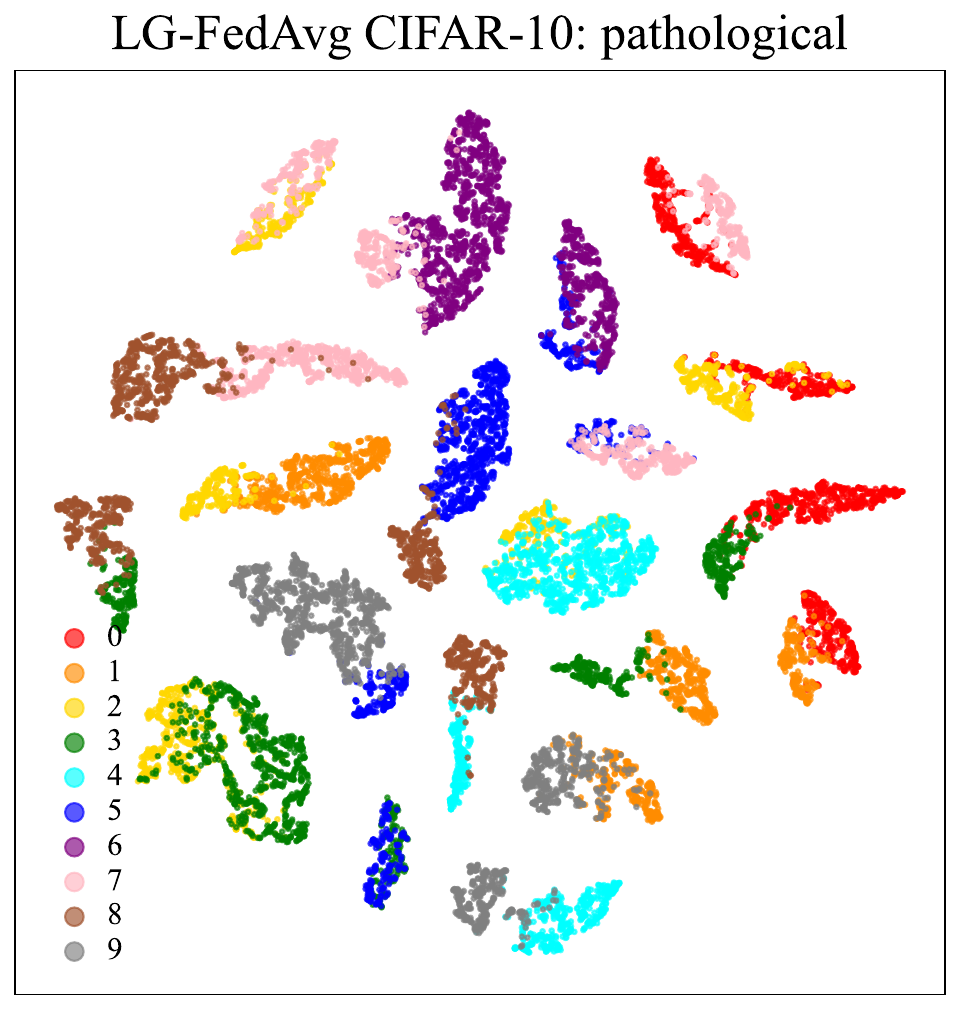}  }
    \subfigure{\label{Fig:rep_GH_}
    \includegraphics[width=.15\textwidth]{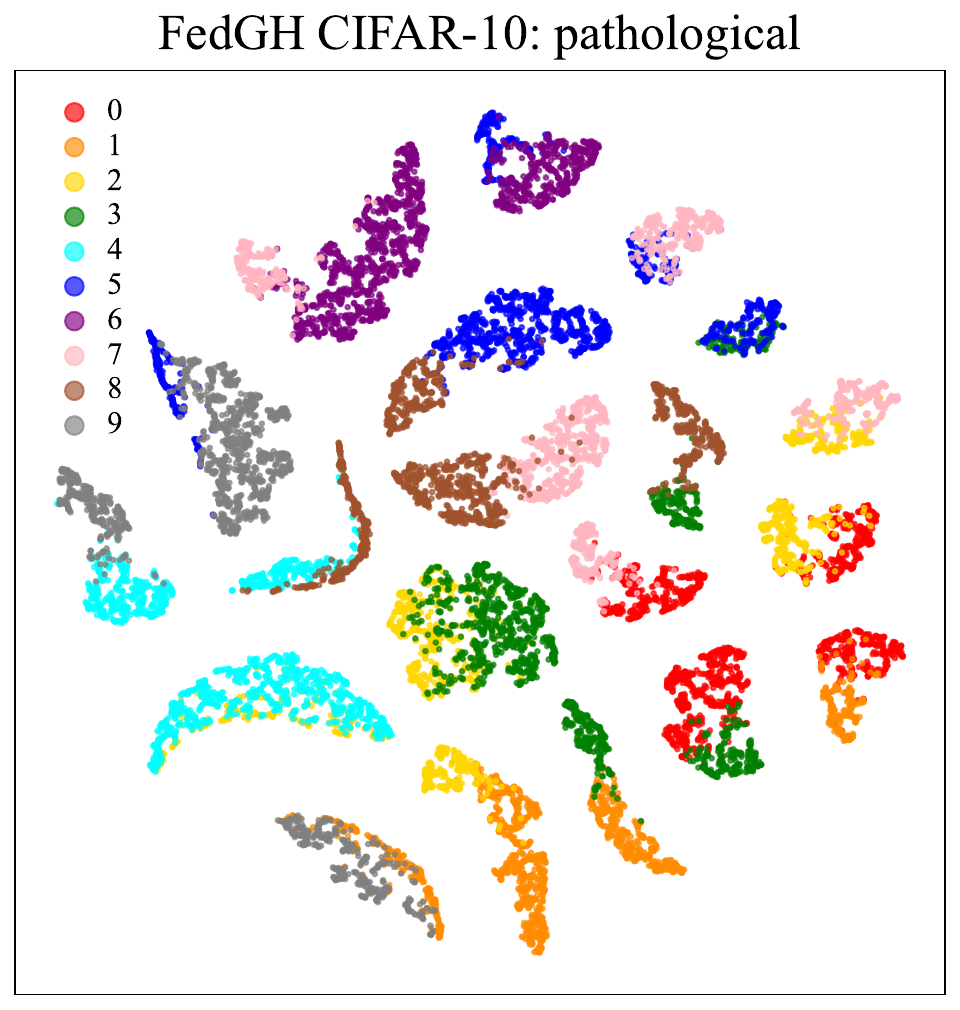}  }
    \subfigure{\label{Fig:rep_proto_}
    \hspace{-8pt}\includegraphics[width=.15\textwidth]{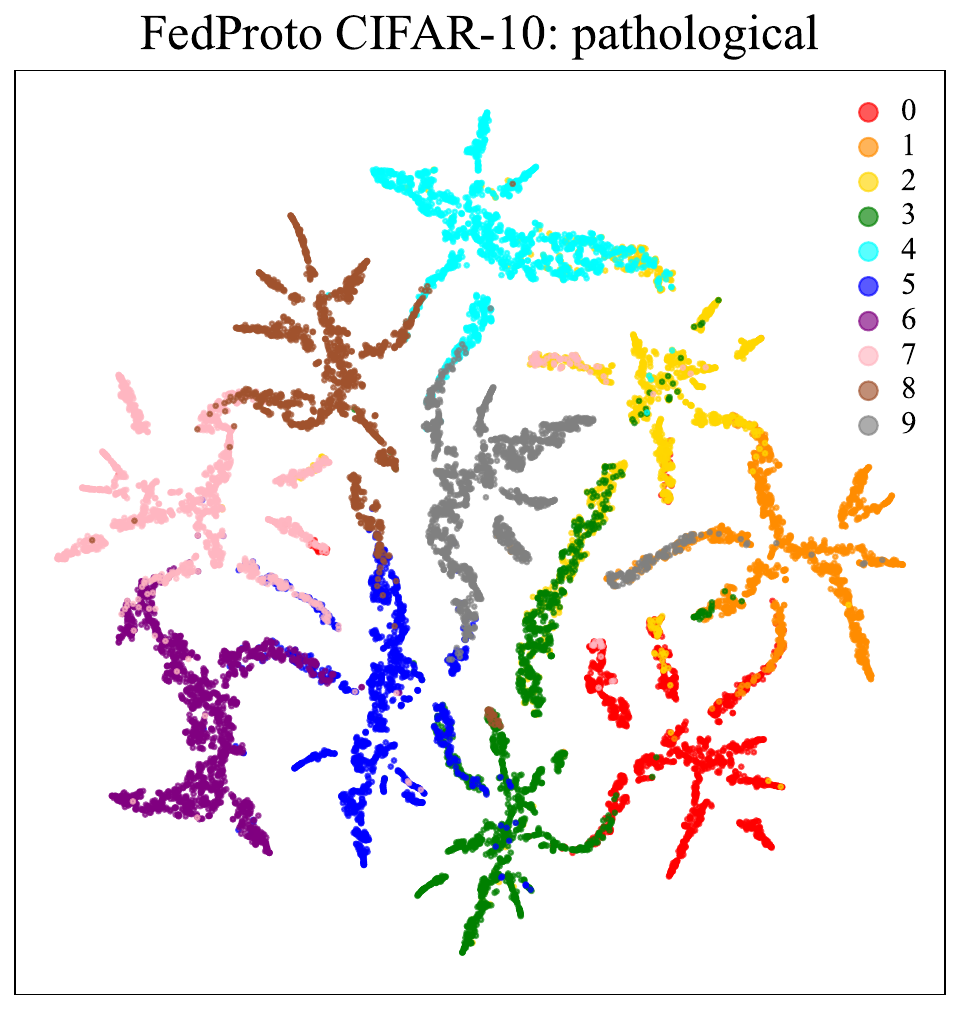}  }
    \subfigure{\label{Fig:FedCoSR_}
    \hspace{-8pt}\includegraphics[width=.15\textwidth]{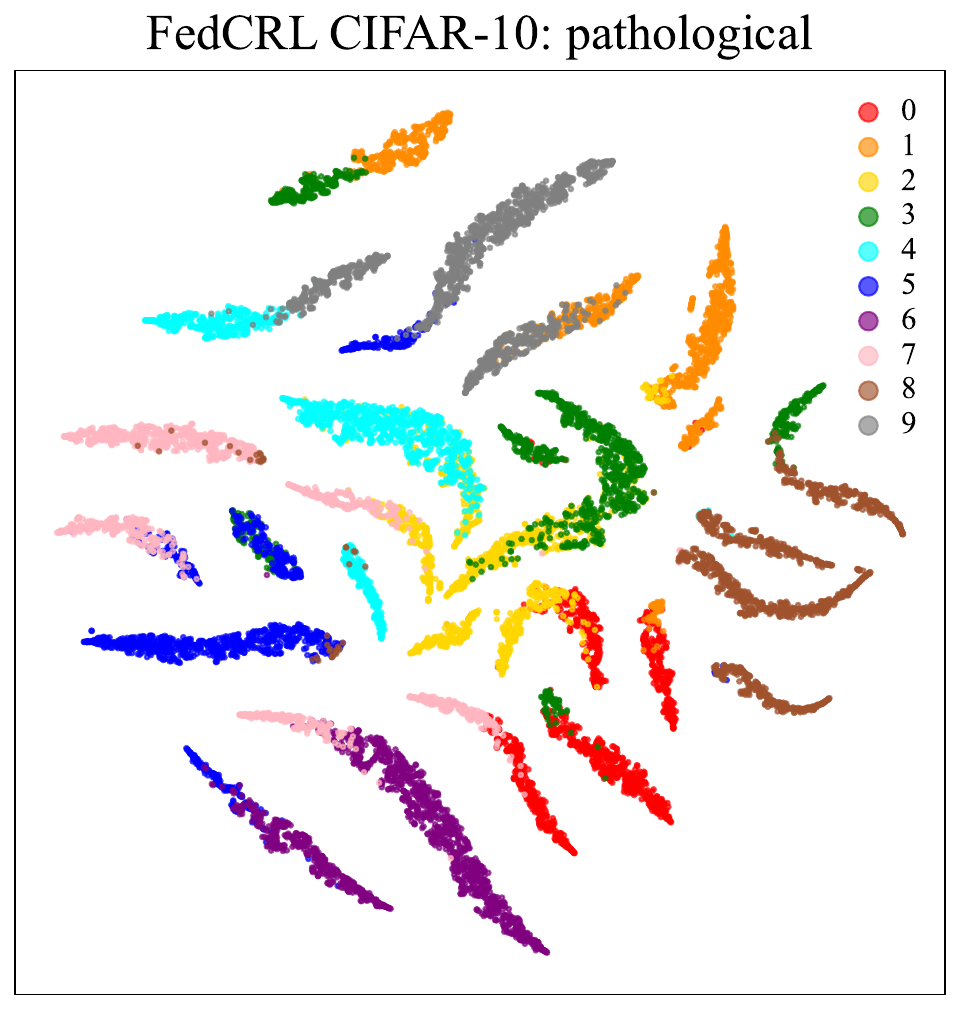}  }
    \captionsetup{width=.99\linewidth}
    \caption{Pathological setting: Visualization of the raw data, representations of FedCoSR and other four baselines through t-SNE.}
    \label{Fig:rep_visual_path}
\end{figure}

\section{Visualization of Representations}

We visualize the samples in CIFAR-10 dataset using t-SNE. Fig.~\ref{Fig:rep_visual_path} show the visualization of the pathological setting. The colored points symbolize the representations of samples from distinct classes. We can see that FedAvg achieves a certain degree of clustering, allowing its representations to be distinguishable from the raw data. However, PFL exhibits more distinctive learning, causing the representations to cluster into different shapes of small clusters according to certain data patterns. It can be observed that the model splitting-based methods like LG-FedAvg and FedGH seem unable to effectively differentiate labels, as their clusters appear to be formed by at least 2 types of labels mixed together. This phenomenon can be attributed to the incomplete aggregation of the model, preventing a good fusion between the parameters of the representation layers and the projection layers of the local model. Therefore, aggregation or training based solely on model splitting is suboptimal.

In contrast, FedProto and our method, FedCoSR, can more clearly distinguish representations into clusters of uniform colors, indicating that methods based on shared representations can better aid the global model in learning the characteristics of the data. Nevertheless, FedProto exhibits a stickiness in clustering, meaning there are connections between the representations corresponding to different labels, causing the representations near the connecting regions to not be well distinguished, especially in the pathological setting. This occurs because FedProto only utilizes shared representations for purely local personalization while ignoring model-level information, resulting in the lack of an effective global model, since its performance based on personalized evaluation is still weaker than FedCoSR. Our method combines model aggregation and the learning of shared representations, striking a balance between generalizability and personalization. Consequently, the global model aggregated from the optimally personalized models of FedCoSR even surpasses strong PFL benchmarks like FedProto in terms of performance.

\end{document}